\documentclass[journal]{IEEEtran}
\usepackage{epsfig}
\usepackage{graphicx}
\usepackage{mathtools,amssymb}
\usepackage{multirow}
\usepackage{dcolumn}
\usepackage{color}

\ifCLASSOPTIONcompsoc
  \usepackage[nocompress]{cite}
\else
  \usepackage{cite}
\fi
\usepackage{algorithm,algorithmic}
\ifCLASSOPTIONcompsoc
  \usepackage[caption=false,font=footnotesize,labelfont=sf,textfont=sf]{subfig}
\else
  \usepackage[caption=false,font=footnotesize]{subfig}
\fi
\usepackage{stfloats}
\usepackage{url}

\begin{document}
%
\title{Multispectral Palmprint Encoding and Recognition}
%
%
%

\author{Zohaib~Khan,
        Faisal~Shafait,
        Yiqun~Hu,
        and~Ajmal~Mian
\thanks{Z.~Khan, F.~Shafait, Y.~Hu and A.~Mian are with the School of Computer Science and Software Engineering, The University of Western Australia.\protect\\
E-mail: zohaib@csse.uwa.edu.au}
\thanks{\textcolor{red}{MATLAB source codes:} \textcolor{blue}{www.sites.google.com/site/zohaibnet/Home/codes}}}

\maketitle

\begin{abstract}
Palmprints are emerging as a new entity in multi-modal biometrics for human identification and verification. Multispectral palmprint images captured in the visible and infrared spectrum not only contain the wrinkles and ridge structure of a palm, but also the underlying pattern of veins; making them a highly discriminating biometric identifier.
In this paper, we propose a feature encoding scheme for robust and highly accurate representation and matching of multispectral palmprints. To facilitate compact storage of the feature, we design a binary hash table structure that allows for efficient matching in large databases. Comprehensive experiments for both identification and verification scenarios are performed on two public datasets -- one captured with a contact-based sensor (PolyU dataset), and the other with a contact-free sensor (CASIA dataset). Recognition results in various experimental setups show that the proposed method consistently outperforms existing state-of-the-art methods. Error rates achieved by our method (0.003\% on PolyU and 0.2\% on CASIA) are the lowest reported in literature on both dataset and clearly indicate the viability of palmprint as a reliable and promising biometric. All source codes are publicly available.


\end{abstract}



%
\IEEEpeerreviewmaketitle

%
\ifCLASSOPTIONcompsoc
  \noindent\raisebox{2\baselineskip}[0pt][0pt]%
  {\parbox{\columnwidth}{\section{Introduction}\label{sec:introduction}%
  \global\everypar=\everypar}}%
  \vspace{-1\baselineskip}\vspace{-\parskip}\par
\else
  \section{Introduction}\label{sec:introduction}\par
\fi
%

%
%
%
%

\IEEEPARstart{T}{he} information present in a human palm has an immense amount of potential for biometric recognition. Information visible to the naked eye includes the principal lines, the wrinkles and the fine ridges which form a unique pattern for every individual \cite{zhang2012comparative}. These superficial features can be captured using standard imaging devices. High resolution scanners capture the fine ridge pattern of a palm which is generally employed for latent palmprint identification in forensics~\cite{jain2009latent}. The principal lines and wrinkles acquired with low resolution sensors are suitable for security applications like user identification or authentication~\cite{zhang2003online}.

\begin{figure}[t]
\centering
\begin{minipage}[b]{0.32\linewidth}
\subfloat[\emph{Lines} ($460$nm)]{\includegraphics[trim = 130pt 10pt 30pt 0pt, clip, width=1\linewidth]{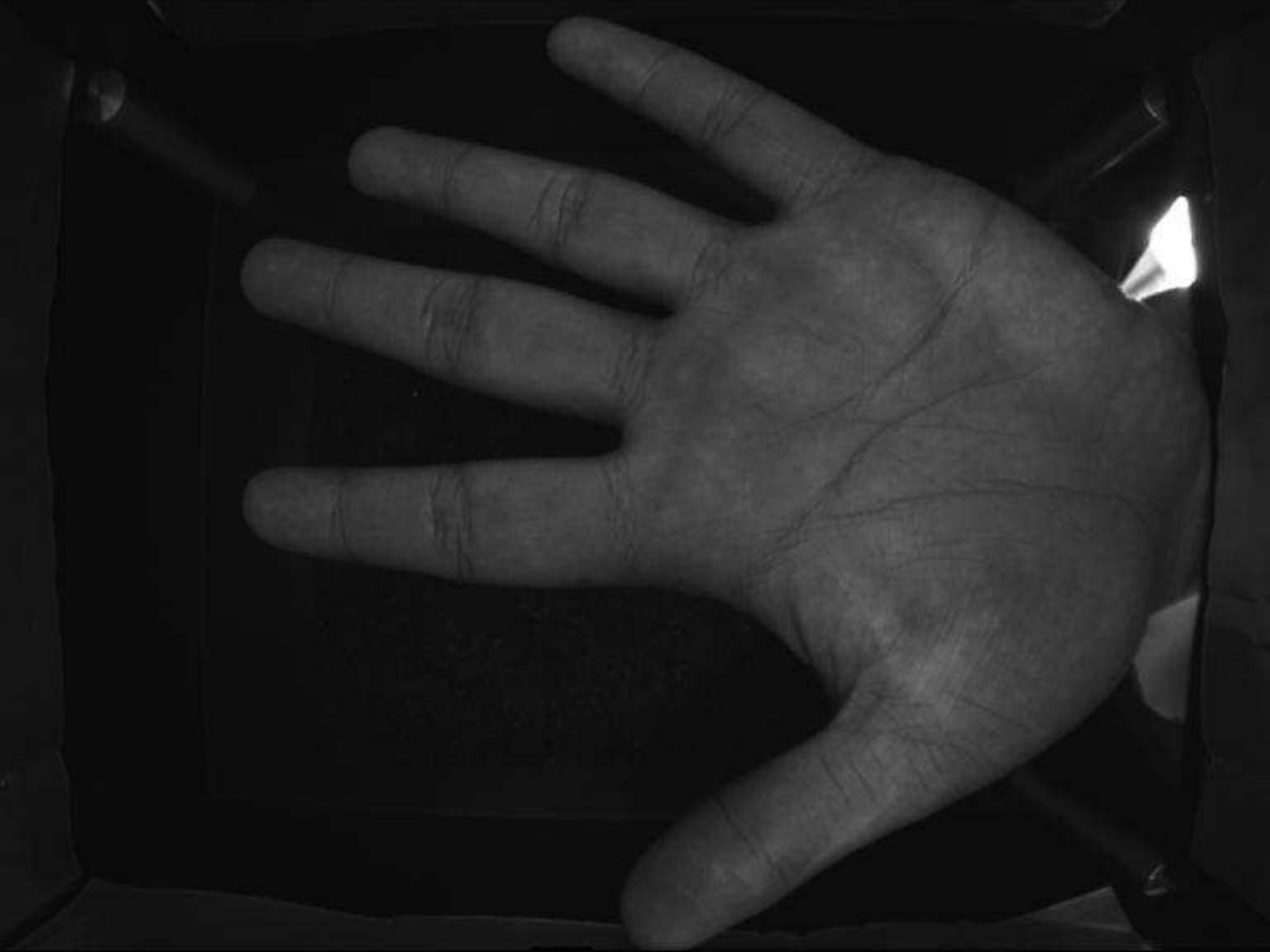}}\\
\subfloat[\emph{Lines} ($470$nm)]{\includegraphics[width=1\linewidth]{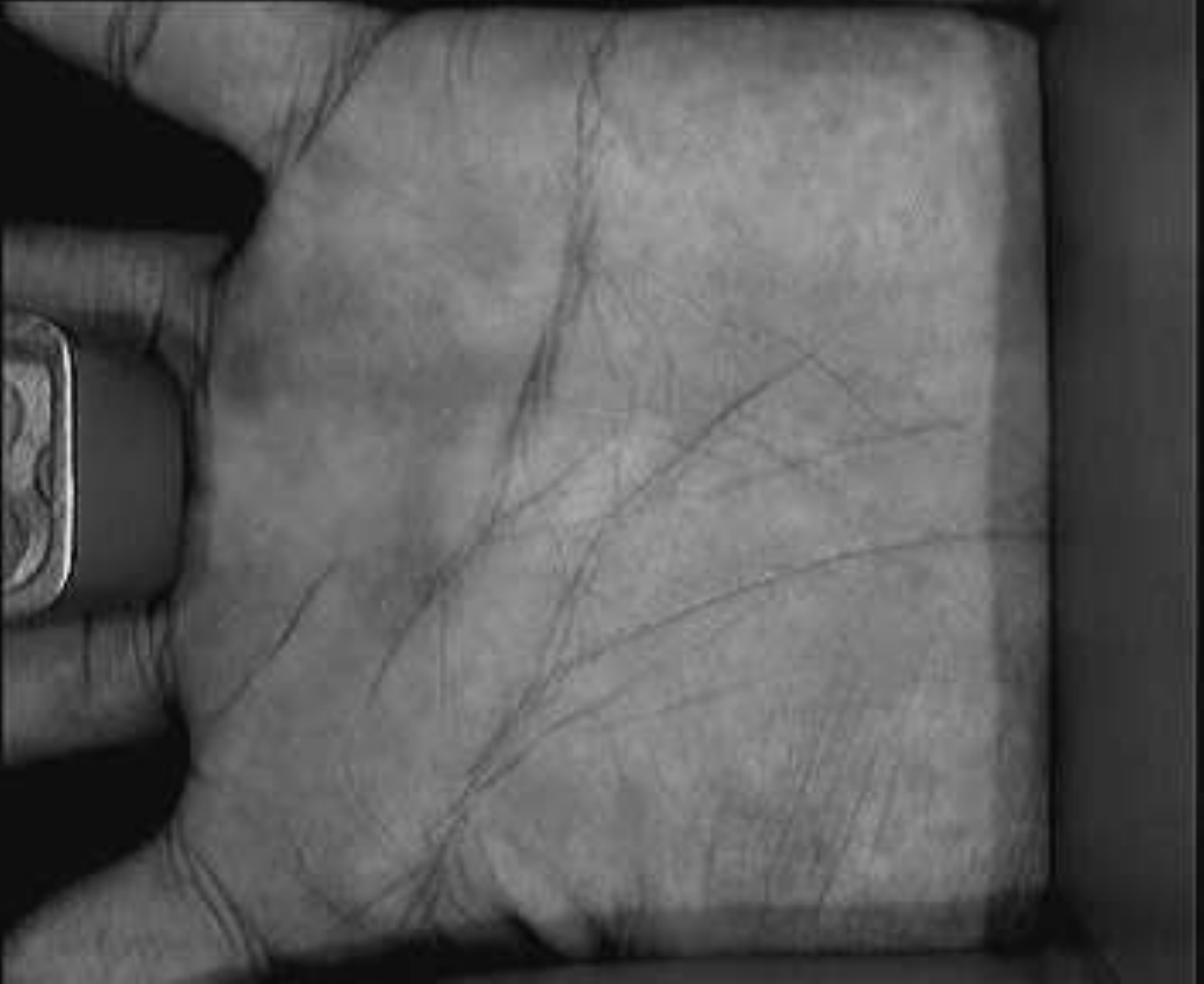}}
\end{minipage}\hspace{0.5pt}
\begin{minipage}[b]{0.32\linewidth}
\subfloat[\emph{Veins} ($940$nm)]{\includegraphics[trim = 130pt 10pt 30pt 0pt, clip, width=1\linewidth]{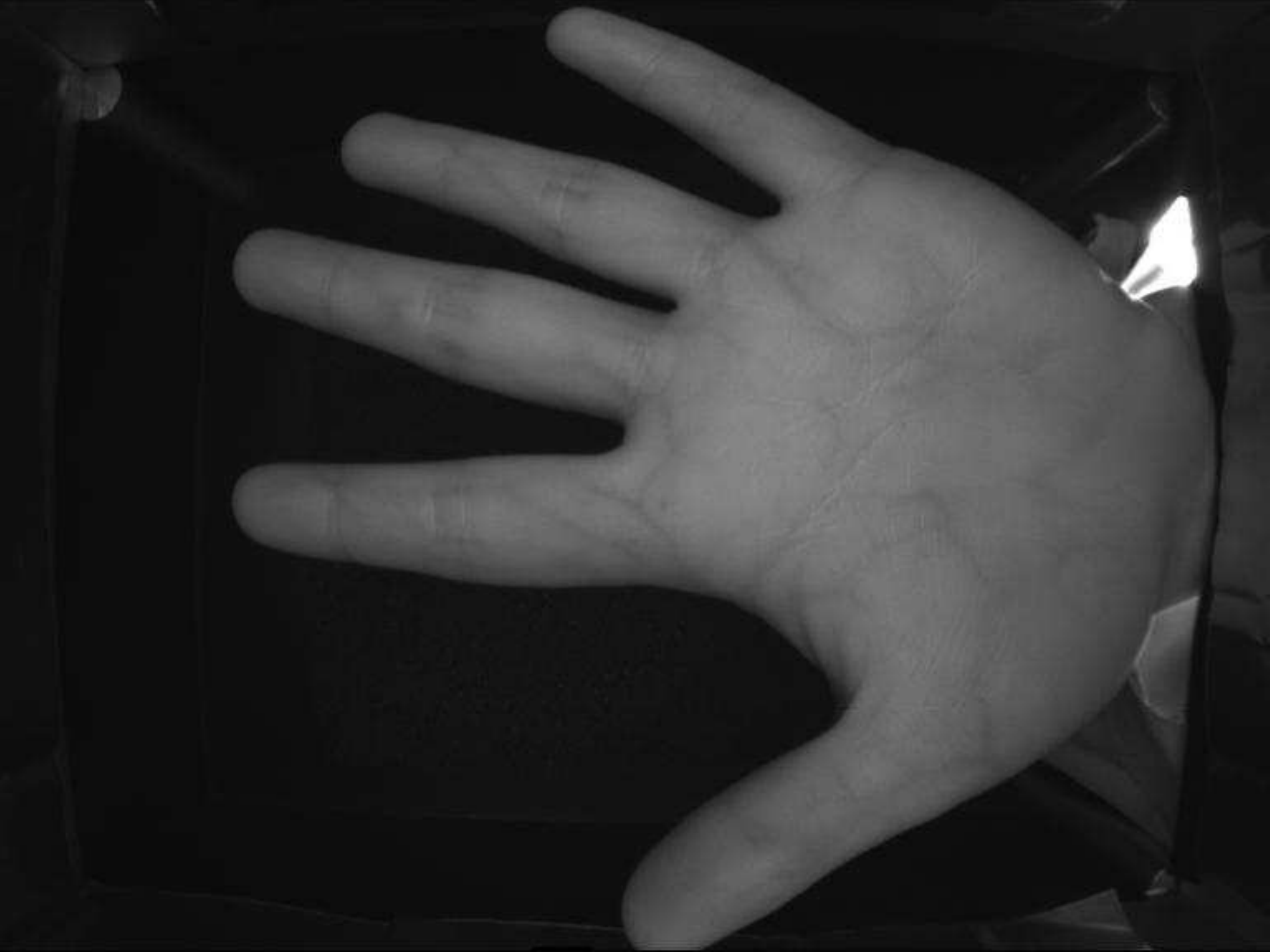}}\\
\subfloat[\emph{Veins} ($880$nm)]{\includegraphics[width=1\linewidth]{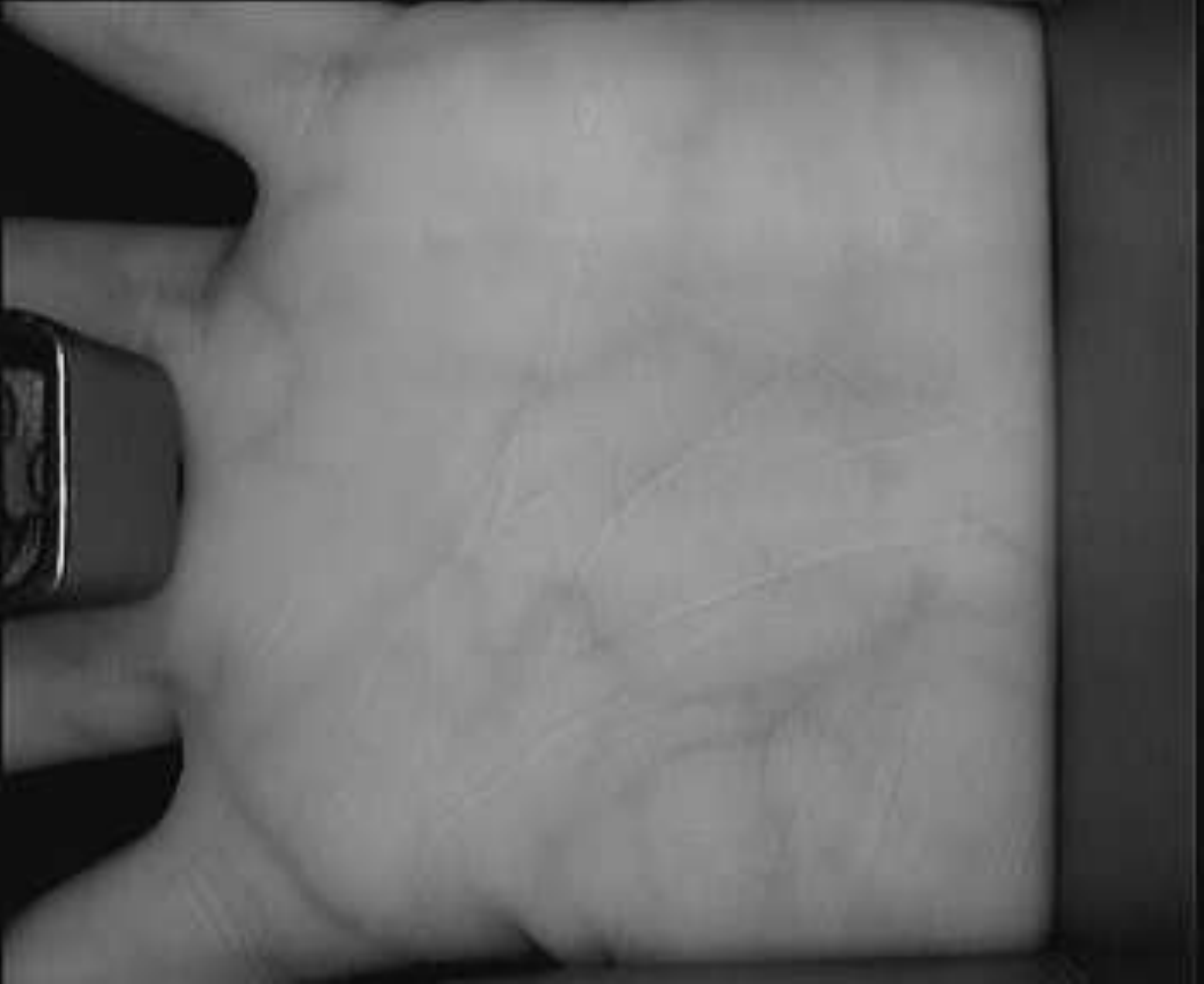}}
\end{minipage}\hspace{0.5pt}
\begin{minipage}[b]{0.32\linewidth}
\subfloat[\emph{Both} ($630$nm)]{\includegraphics[trim = 130pt 10pt 30pt 0pt, clip, width=1\linewidth]{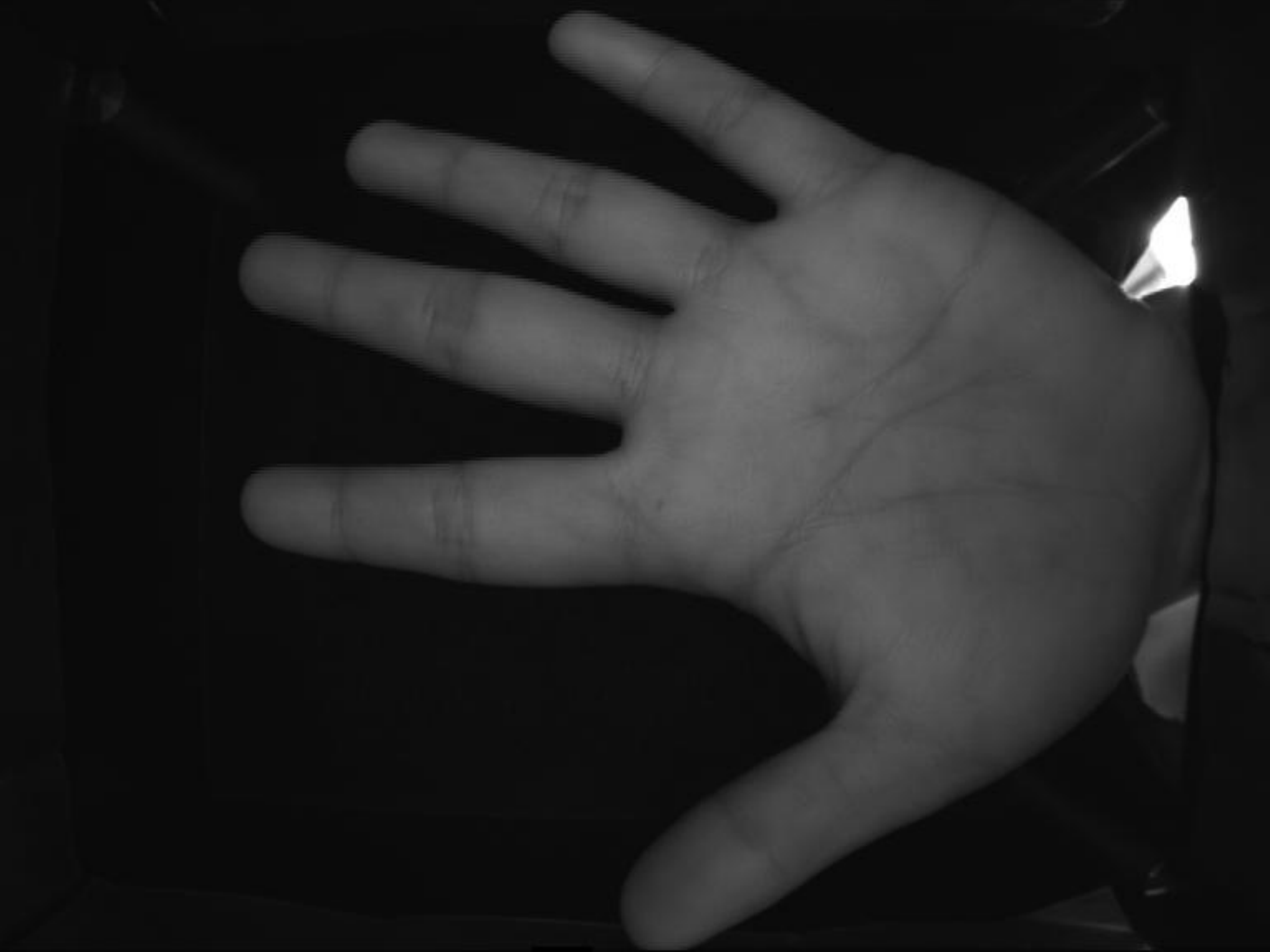}}\\
\subfloat[\emph{Both} ($660$nm)]{\includegraphics[width=1\linewidth]{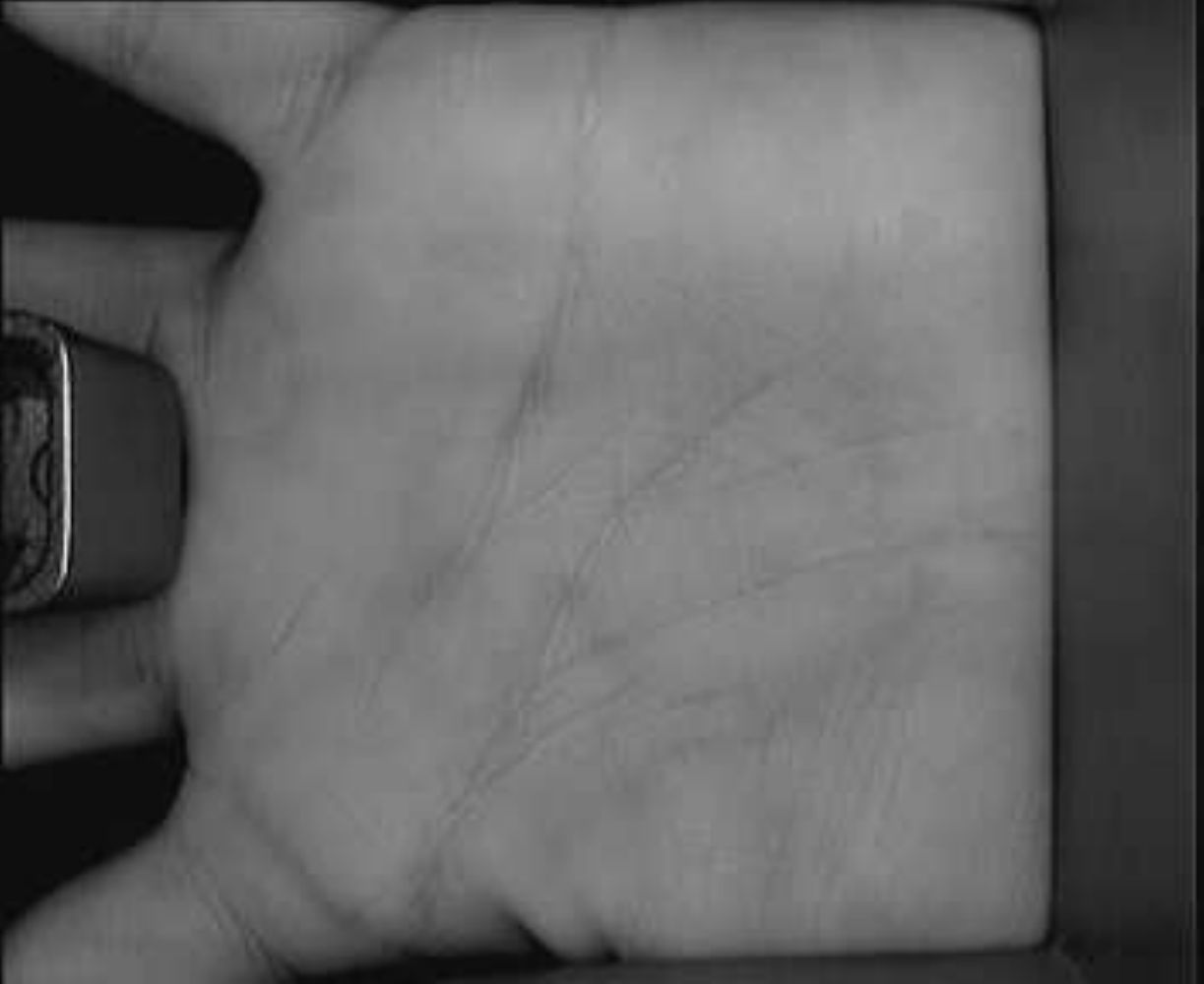}}
\end{minipage}
\caption{Examples of palmprint features in multiple bands. Three bands of a multispectral image captured using a non-contact (top row) and a contact based sensor (bottom row). (a-b) Palm lines captured in the visible range. (c-d) Palm veins captured in the near infrared range. (e-f) A combination of the line and vein features at intermediate wavelengths.}
\label{fig:palms}
\end{figure}

Additional information present in the human palm is the subsurface vein pattern which is indifferent to the palm lines. Such features cannot be easily acquired by a standard imaging sensor. Infrared imaging can capture subsurface features due to its capability to penetrate the human skin. The superficial and subsurface features of a palm have been collectively investigated under the subject of \emph{`multispectral palmprint recognition'}. Using \emph{Multispectral Imaging (MSI)}, it is possible to capture images of an object at multiple wavelengths of light, in the visible spectrum and beyond. Fig.~\ref{fig:palms} shows palm images captured at three different wavelengths. The availability of such complementary features (palm lines and veins) allows for increased discrimination between individuals. These useful characteristics of palmprint make it a suitable choice for recognition where user cooperation is affordable, e.g, at workplace entry, secure access gates and identity records.

A monochromatic camera under spectrally varying illuminations can acquire multispectral palm images~\cite{han2008multispectral,hao2007comparative}. Contact devices, restrict the hand movement but raise user acceptability issues due to hygiene. On the other hand, biometrics that are acquired with non-contact sensors are user friendly and socially more acceptable~\cite{jain2004introduction}, but introduce challenge of rotation, scale and translation (RST) variation. The misalignments casued by movement of the palm can be recovered by reliable, repeatable landmark detection and subsequent ROI extraction. Moreover, the multi-modal nature of palm requires robust feature extraction scheme that accurately detects line-like features. In addition, the large amount of features resulting from multi-modal feature extraction pose slow matching time and cost large storage resources. These challenges can be addressed by efficient feature encoding and matching scheme. We propose an end-to-end framework for multispectral palmprint recognition and attempt to solve all realizable challenges.

The key contributions of this work are
\begin{itemize}
  \item A reliable technique for ROI extraction from non-contact palmprint images.
  \item A robust multidirectional feature encoding for multispectral palmprints.
  \item An efficient hash table scheme for compact storage and matching of multi-directional features.
  \item A thorough analysis of palmprint verification and identification experiments and comparison with the state-of-the art techniques
\end{itemize}



\section{Related Work}
\label{sec:work}

In the past decade, biometrics such as the iris~\cite{boyce2006multispectral}, face~\cite{di2010studies,pan2003face} and fingerprint~\cite{rowe2007multispectral} have been investigated using multispectral images for improved accuracy. Recently, there has been an increased interest in multispectral palmprint recognition~\cite{meraoumia2013efficient,mittal2012rank,luo2012multispectral,akbari2011multispectral,kekre2011palmprint,tahmasebi2011novel,zhang2011online,kisku2010multispectral,guo2010feature,zhang2010online,xu2010multispectral,han2008multispectral,hao2008multispectral,hao2007comparative}. Palmprint recognition approaches can be categorized into line-like feature detectors, subspace learning methods and texture based coding techniques~\cite{kong2009survey}. These three categories are not mutually exclusive and their combinations are also possible. Line detection based approaches commonly extract palm lines using edge detectors. Huang et al.~\cite{huang2008palmprint} proposed a palmprint verification technique based on principal lines. The principal palm lines were extracted using a modified finite Radon transform and a binary edge map was used for representation. However, recognition based solely on palm lines proved insufficient due to their sparse nature and the possibility of different individuals to have highly similar palm lines~\cite{zhang2011online}. Although, line detection can extract palm lines effectively, it may not be equally useful for the extraction of palm veins due to their low contrast and broad structure.

A subspace projection captures the global characteristics of a palm by projecting to the most varying (in case of PCA) or the most discriminative (in case of LDA) dimensions. Subspace projection methods include eigenpalm~\cite{lu2003palmprint}, which globally projects palm images to a PCA space, or fisherpalm \cite {wu2003fisherpalms} which projects to an LDA space. However, the finer local details are not well preserved and modeled by such subspace projections. Wang et al.~\cite{wang2007fusion} fused palmprint and palmvein images and proposed the \emph{Laplacianpalm} representation. Unlike the eigenpalm~\cite{lu2003palmprint} or the fisherpalm \cite {wu2003fisherpalms}, the \emph{Laplacianpalm} representation attempts to preserve the local characteristics as well while projecting onto a subspace. Xu et al.~\cite{xu2010multispectral} represented multispectral palmprint images as quaternion and applied quaternion PCA to extract features. A nearest neighbor classifier was used for recognition using quaternion vectors. The quaternion model did not prove useful for representing multispectral palm images and demonstrated low recognition accuracy compared to the state-of-the-art techniques. The main reason is that subspaces learned from misaligned palms are unlikely to generate accurate representation of each identity.

Orientation codes extract and encode the orientation of lines and have shown state-of-the-art performance in palmprint recognition~\cite{zhang2012comparative}. Examples of orientation codes include the Competitive Code (CompCode)~\cite{kong2004competitive}, the Ordinal Code (OrdCode)~\cite{sun2005ordinal} and the Derivative of Gaussian Code (DoGCode)~\cite{wu2006palmprint}. In the generic form of orientation coding, the response of a palm to a bank of directional filters is computed such that the resulting directional subbands correspond to specific orientations of line. Then, the dominant orientation index from the directional subbands is extracted at each point to form the orientation code. CompCode~\cite{kong2004competitive} employs a directional bank of Gabor filters to extract the orientation of palm lines. The orientation is encoded into a binary code and matched directly using the Hamming distance. The OrdCode~\cite{sun2005ordinal} emphasizes the ordinal relationship of lines by comparing mutually orthogonal filter pairs to extract the feature orientation at a point. The DoGCode~\cite{wu2006palmprint} is a compact representation which only uses vertical and horizontal gaussian derivative filters to extract feature orientation. Orientation codes can be binarized for efficient storage and fast matching unlike other representations which require floating point data storage and computations. Another important aspect of multispectral palmprints is how to combine the different spectral bands. The multispectral palmprints have been investigated with data, feature, score and rank-level fusion.


Multispectral palmprints are generally fused at image level using multi-resolution transforms, such as wavelet and curvelet. Han et al.~\cite{han2008multispectral} used a three level wavelet fusion strategy for combining multispectral palmprint images. After fusion, CompCode was used for feature extraction and matching. Their results showed that the wavelet fusion of multispectral palm images is only useful for blurred source images. Hao et al.~\cite{hao2008multispectral} used various image fusion techniques and the OLOF representation for multispectral palmprint recognition. The best recognition performance was achieved when the curvelet transform was used for band fusion. Kisku et al.~\cite{kisku2010multispectral} proposed wavelet based band fusion and gabor wavelet feature representation for multispectral palm images. To reduce the dimensionality, feature selection was performed using the Ant Colony Optimization (ACO) algorithm~\cite{dorigo1997ant} and classified by normalized correlation and SVM. However, the gabor wavelet based band fusion could not improve palmprint recognition performance compared to the curvelet fusion with OLOF~\cite{hao2008multispectral}. Kekre et al.~\cite{kekre2011palmprint} proposed a hybrid transform by kronecker product of DCT and Walsh transforms, which better describes the energy in the local regions of a multispectral palm. A subset of the regions was selected by comparison with the mean energy map and stored as features for matching. It is observed that fusion of multispectral palmprints is cumbersome due to multimodal nature of palm. A single fused palm image is a compromise of the wealth of complementary information present in different bands, which results in below par recognition performance.


Fusion of spectral bands has been demonstrated at feature level. Luo et al.~\cite{luo2012multispectral} used feature level band fusion for multispectral palmprints. Specifically, a modification of CompCode was combined with the original CompCode and features from the pair of less correlated bands were fused. The results indicated an improvement over image level fusion and were comparable to match-score level fusion. Zhou and Kumar~\cite{zhou2010contactless} encoded palm vein features by enhancement of vascular patterns and using the Hessian phase information. They showed that a combination of various feature representations can be used for achieving improved performance based on palmvein images. Mittal et al.~\cite{mittal2012rank} investigated fuzzy and sigmoid features for multispectral palmprints and a rank-level fusion of scores using various strategies. It was observed that a nonlinear fusion function at rank-level was effective for improved recognition performance. Tahmasebi et al.~\cite{tahmasebi2011novel} used gabor kernels for feature extraction from multispectral palmprints and a rank-level fusion scheme for fusing the outputs from individual band comparisons. One drawback of rank-level fusion is that it assigns fixed weights to the rank outputs of spectral bands, which results in sub-optimal performance.


Zhang et al.~\cite{zhang2010online} compared palmprint matching using individual bands and reported that the red band performed better than the near infrared, blue and green bands. A score level fusion of these bands achieved superior performance compared to any single band. Another joint palmline and palmvein approach for multispectral palmprint recognition was proposed by Zhang et al.~\cite{zhang2011online}. They designed separate feature extraction methodologies for palm line and palm vein and later used score level fusion for computing the final match. The approach yielded promising results, albeit at the cost of increased complexity. A comparison of different fusion strategies indicates that a score level fusion of multispectral bands is promising and most effective compared to a data, feature or rank-level fusion.

It is worth mentioning that a simple extension of the existing palmprint representations to multispectral palmprints may not fully preserve the features that appear in different bands. For example, a single representation may not be able to extract equally useful features from both lines and veins~\cite{zhang2011online}. Comparative studies show that local orientation features are the recommended choice for palmprint feature extraction~\cite{zhang2012comparative}.

In this paper, we propose a novel orientation and binary hash table based encoding for robust and efficient multispectral palmprint recognition. The representation is derived from the coefficients of the Nonsubsampled Contourlet Transform (NSCT) which has the advantage of robust directional frequency localization. Unlike existing orientation codes, which apply a directional filter bank directly to a palm image, we propose a two stage filtering approach to extract only the robust directional features. We develop a unified methodology for the extraction of multispectral (line and vein) features. The feature binary encoded into an efficient hash table structure that only requires indexing and summation operations for simultaneous one-to-many matching with an embedded score level fusion of multiple bands. This paper is an extension of our earlier work~\cite{khan2011contour}. Here we give more detailed descriptions and introduce two variants of the proposed matching technique to improve robustness to inter-band misalignments. We perform a more thorough analysis of the pyramidal directional filter pair combination and the effect of varying other parameters. We also implement three existing state-of-the art orientation features~\cite{kong2004competitive}\cite{sun2005ordinal}\cite{wu2006palmprint} and compare their performances to the proposed feature in various experimental settings including a palmprint identification scenario.

The rest of this paper is organized as follows. In Section~\ref{sec:roi_ext}, we present a reliable technique for ROI extraction from multispectral images acquired via non-contact sensors. The Contour Code extraction, hash table encoding and matching processes are detailed in Section~\ref{sec:cnt_code}. In Section~\ref{sec:exp}, we present a detailed experimental analysis to evaluate the accuracy of the proposed ROI extraction. We also analyze the effect of various parameters and pyramidal-directional filter pair combinations. We perform verification and identification experiments in various experimental settings using two standard multispectral palmprint databases and compare the performance of the proposed Contour Code to existing state-of-the-art. Conclusions are given in Section~\ref{sec:conc}.

\section{Region of Interest Extraction}
\label{sec:roi_ext}

The extraction of a reliable ROI from such images is a challenge which is addressed in this paper. To extract ROI from an image, it is necessary to define some reference \emph{landmarks} from within the image which can be used to normalize its relative movement. The landmark detection must be accurate and repeatable to ensure that the exact same ROI is extracted from every palm image. Among the features commonly used in hand geometry recognition, the valleys between the fingers are a suitable choice for landmarks due to their invariance to hand movement~\cite{fang2009making}.

\subsection{Preprocessing}

The input hand image is first thresholded to segment it from the background (see  Fig.~\ref{fig:hand_gs} and~\ref{fig:hand_tr}). Smaller objects, not connected to the hand, that appear due to noise are removed through binary pixel connectivity. Morphological closing is carried out with a square structuring element to normalize the contour of the hand. Finally, any holes within the hand pixels are filled using binary hole filling. These operations ensure accurate and successful landmarks localization. The resulting preprocessed segmented hand image is shown in Fig.~\ref{fig:hand_bw}.

\subsection{Localization of Landmarks}

\begin{figure}[t]
\setlength\fboxsep{0pt}
\setlength\fboxrule{0.1pt}
\begin{center}
\subfloat[]{\label{fig:hand_gs}\includegraphics[trim = 130pt 10pt 30pt 0pt, clip, width=0.315\linewidth]{001_l_460_01}}\hspace{0.05pt}
\subfloat[]{\label{fig:hand_tr}\includegraphics[trim = 130pt 10pt 30pt 0pt, clip, width=0.315\linewidth]{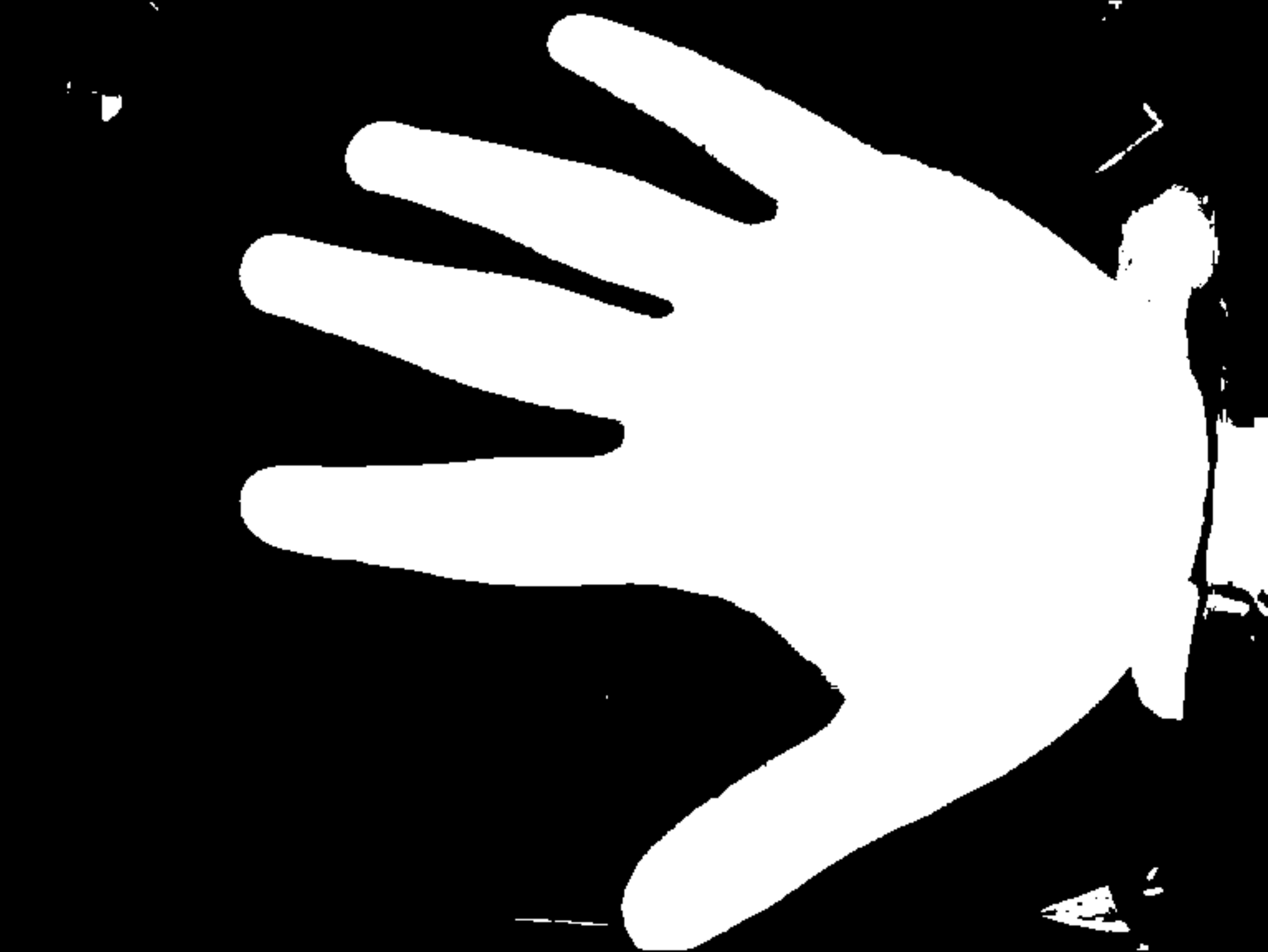}}\hspace{0.05pt}
\subfloat[]{\label{fig:hand_bw}\includegraphics[trim = 130pt 10pt 30pt 0pt, clip, width=0.315\linewidth]{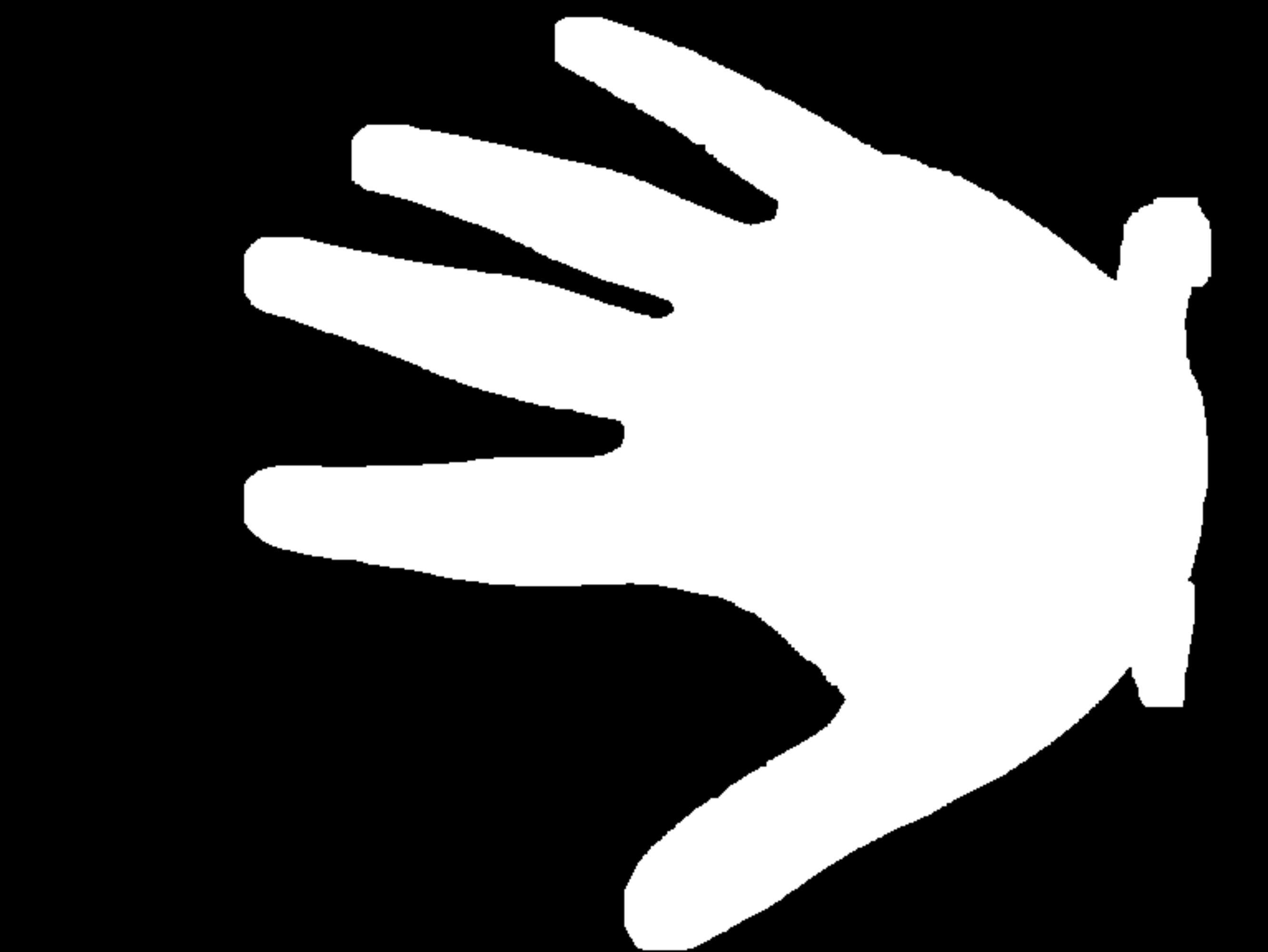}}\\
\subfloat[]{\label{fig:hand_mp}\includegraphics[trim = 80pt 10pt 20pt 0pt, clip, width=0.315\linewidth]{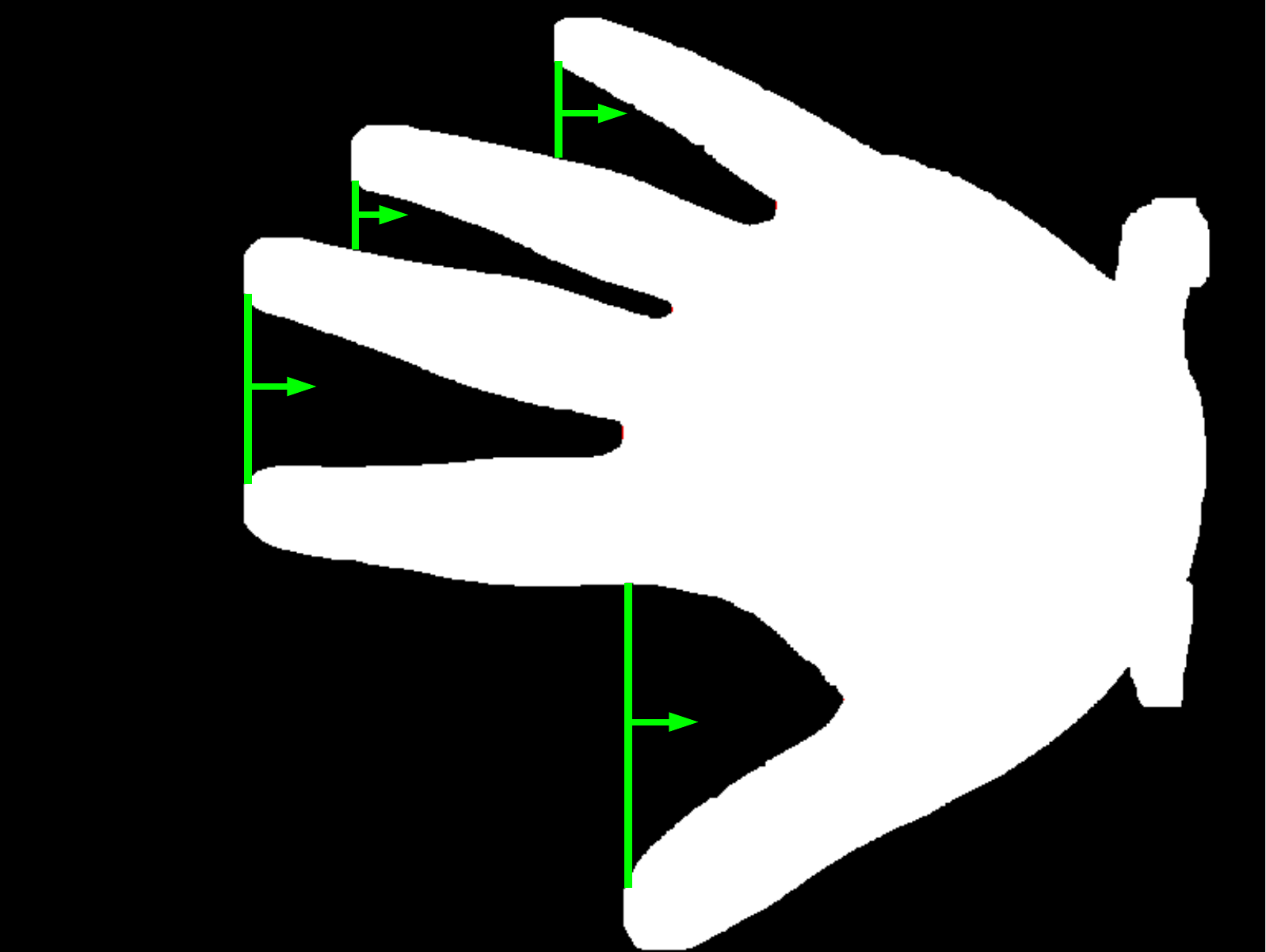}}
\subfloat[]{\label{fig:hand_kpc}
\begin{minipage}[b]{0.163\linewidth}
\fbox{\includegraphics[trim = 264pt 264pt 177pt 70pt, clip, width=1\linewidth]{hand_kpc}}\\
\fbox{\includegraphics[trim = 222pt 230pt 215pt 106pt, clip, width=1\linewidth]{hand_kpc}}\\
\fbox{\includegraphics[trim = 209pt 183pt 233pt 151pt, clip, width=1\linewidth]{hand_kpc}}
\end{minipage}}\hspace{0.05pt}
\subfloat[]{\label{fig:hand_mid}\fbox{\includegraphics[trim = 130pt 215pt 250pt 0pt, clip, width=0.315\linewidth]{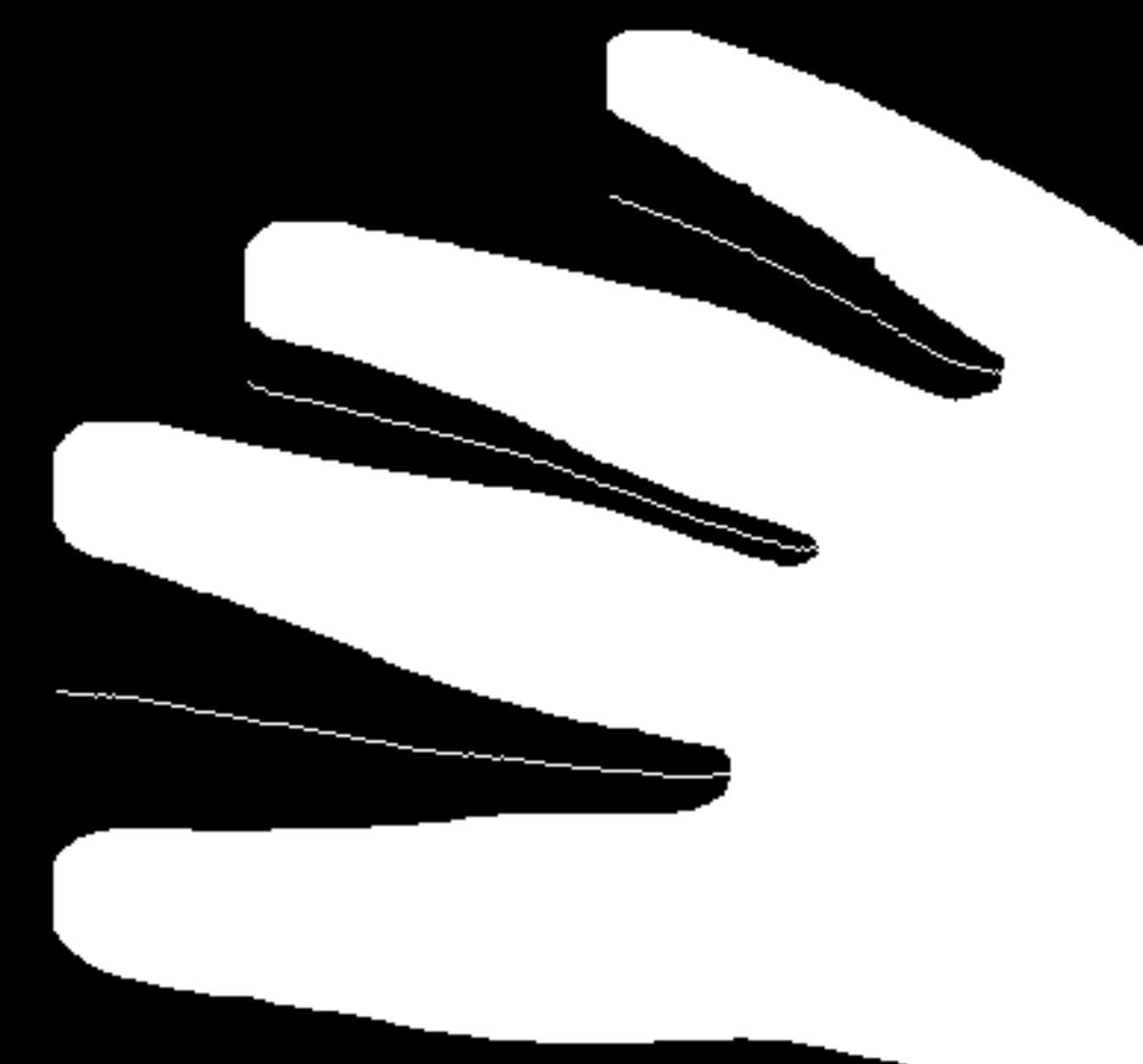}}}
\subfloat[]{\label{fig:hand_kpf}
\begin{minipage}[b]{0.163\linewidth}
\fbox{\includegraphics[trim = 264pt 264pt 177pt 70pt, clip, width=1\linewidth]{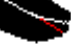}}\\
\fbox{\includegraphics[trim = 222pt 230pt 215pt 106pt, clip, width=1\linewidth]{hand_kpf}}\\
\fbox{\includegraphics[trim = 209pt 183pt 233pt 151pt, clip, width=1\linewidth]{hand_kpf}}
\end{minipage}}\hspace{0.05pt}
\end{center}
\caption{Hand image preprocessing. (a-c) input, thresholded and clean segmented image. (d-e) Landmarks localization in a hand image acquired with a non-contact sensor. (d) Initialization of search for mid points (green). (e) Search termination (red). (f) Located mid points. (g) Polynomial extrapolation of selected mid points (red) to find the landmarks.}
\label{fig:hand_lm}
\end{figure}

Given the binary image, in which the foreground pixels correspond to the hand, the localization proceeds as follows. In a column wise search (Fig.~\ref{fig:hand_mp}), the binary discontinuities, i.e. the edges of the fingers are identified. From the edges, the gaps between fingers are located in each column and the mid points of all the finger gaps are computed. Continuing inwards along the valley of the fingers, the column encountered next to the last column should contain hand pixels (Fig.~\ref{fig:hand_kpc}). The search is terminated when four such valley closings are recorded. This column search succeeds when the hand rotation is within $\pm90^{\circ}$ in the image plane. It is also required that there is some minimal amount of separation between consecutive fingers so that the mid points can be recorded. This is was explicitly achieved by instructing the users on the use of the system. It should also be noted that the out of plane movement of the hand does not affect the localization to a certain extent, as long as the gaps between fingers is maintained.

The mid points corresponding to the four valleys are clustered and the one corresponding to the index-thumb valley is discarded. Due to the natural contour of the fingers, it can be observed that the mid points do not follow a linear trend towards the valley as shown in Fig.~\ref{fig:hand_mid}. Therefore, a second order polynomial is fitted to the mid points of each valley excluding the last $p_e$ points which tend to deviate from the path towards the landmark. The estimated polynomial is then used to extend the mid points in the last $p_e$ points towards the valley as shown in Fig.~\ref{fig:hand_kpf}. We set $p_e$ as $10\%$ of the total number of mid points $p_t$ of a valley. The final location of a landmark is the last encountered background pixel in the sequence of extrapolated points.

\subsection{ROI Extraction}

\begin{figure}[t]
\subfloat[]{\label{fig:roi_scale}\includegraphics[trim = 80pt 10pt 20pt 0pt, clip, width=0.49\linewidth]{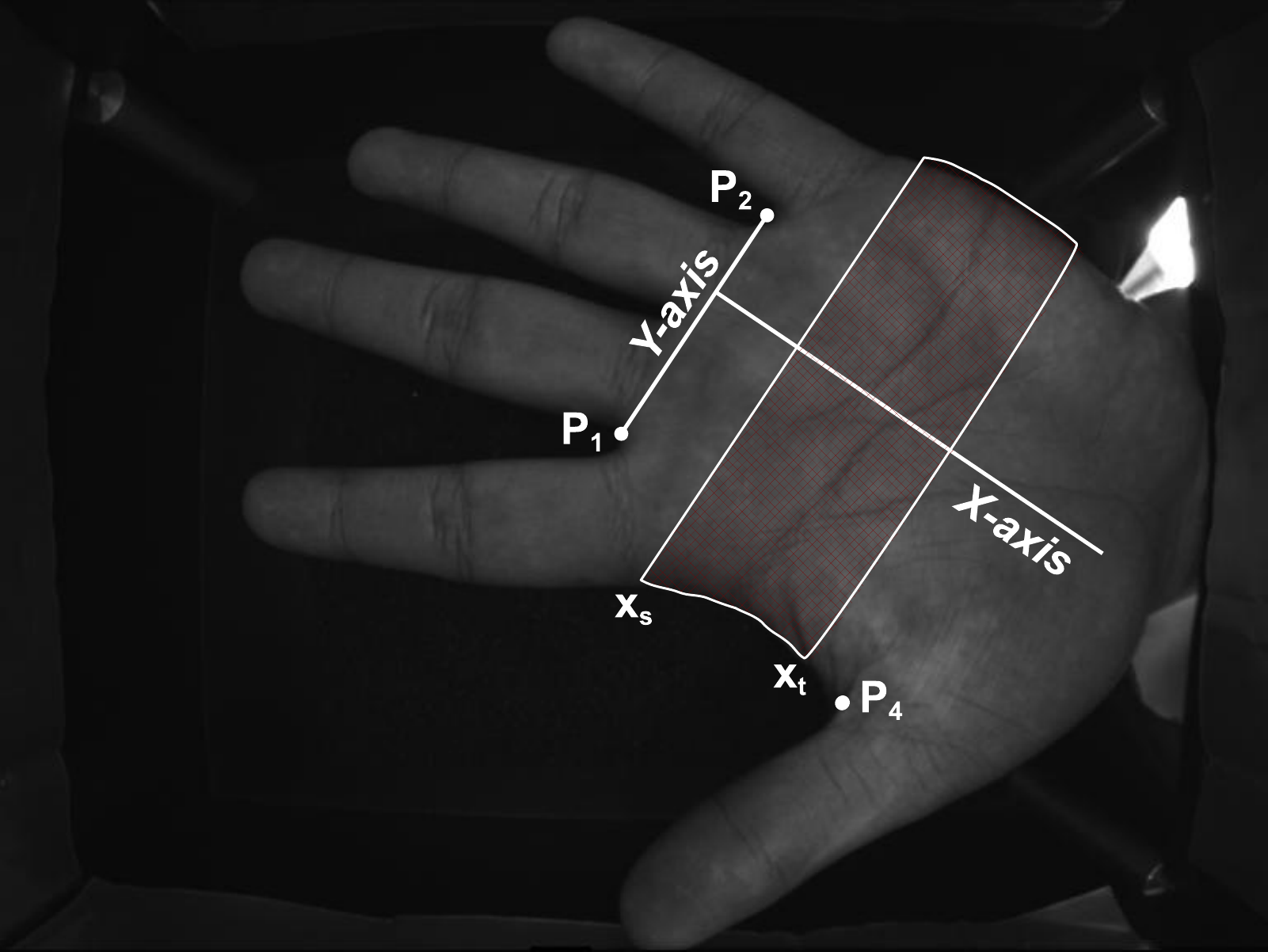}}\hspace{1pt}
\subfloat[]{\label{fig:roi_extract}\includegraphics[trim = 80pt 10pt 20pt 0pt, clip, width=0.49\linewidth]{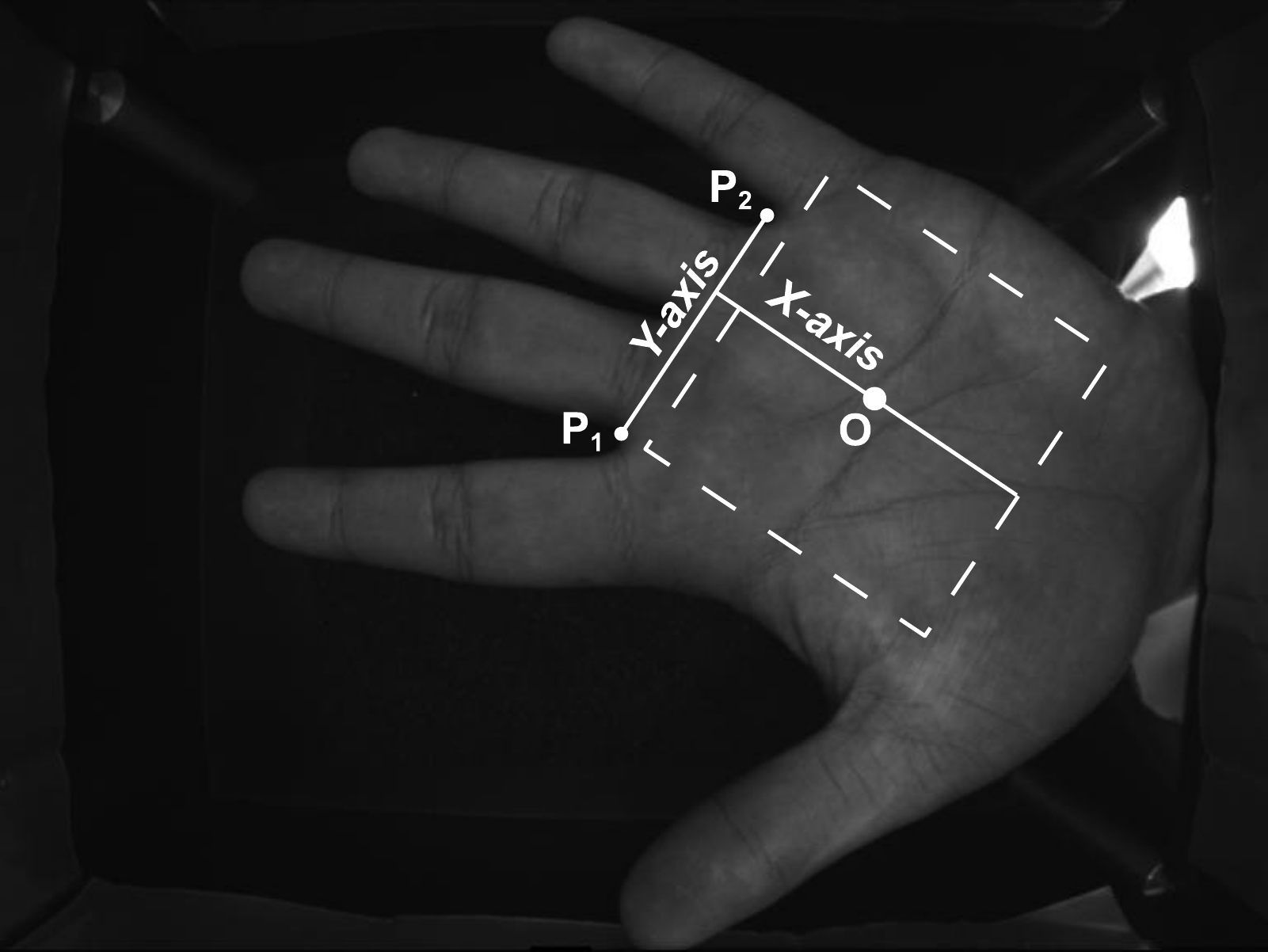}}
\caption{ROI extraction based on detected landmarks. The points $P_{1},P_{2}$ define the \emph{Y-axis}. The \emph{X-axis} is orthogonal to the \emph{Y-axis} at 2/3 distance from $P_{1}$ to $P_{2}$. (a) Average palm width $\bar{w}$ is computed in the range $(x_{s},x_{t})$ (b) The distance of origin O of the ROI is proportional to the average palm width found in (a).}
\label{fig:roi_all}
\end{figure}

Using the three landmarks, an RST invariant ROI can be extracted (see Fig.~\ref{fig:roi_all}). The landmarks ($P_{1}$,$P_{2}$) form a reference \emph{Y-axis}. We fix the \emph{X-axis} at two-thirds of the distance from $P_{1}$ to $P_{2}$ so that the ROI is centered over the palm. The automatically estimated palm width $\bar{w}$ serves as the scale identifier to extract a scale invariant ROI. To compute $\bar{w}$, we find the average width of the palm from point $x_{s}$ to $x_{t}$. To keep $x_{s}$ from being very close to the fingers and affected by their movement we set a safe value of $x_{s}=P_{1}(x)+(P_{2}(y)-P_{1}(y))/3$. Moreover, we set $x_{t}=P_{4}(x)-(P_{2}(y)-P_{1}(y))/12$. Note that the scale computation is not sensitive to the values of $x_{s}$ and $x_{t}$ as the palmwidth is averaged over the range $(x_{s},x_{t})$.

In our experiments, the ROI side length is scaled to $70\%$ of $\bar{w}$ and extracted from the input image using an affine transformation. The same region is extracted from the remaining bands of the multispectral palm.

\subsection{Inter-band Registration}
Since the bands of the multispectral images were sequentially acquired, minor hand movement can not be ruled out. Therefore, an inter-band registration of the ROIs based on the maximization of \emph{Mutual Information} is carried out. The approach has shown to be effective for registering multispectral palmprints~\cite{hao2008multispectral}. Since, there is negligible rotational and scale variation within the consecutive bands, we limit the registration search space to only $\pm2$ pixels translations along both dimensions. The registered ROI of each band is then downsampled to $32\times32$ pixels using bicubic interpolation. This resampling step has several advantages. First, it suppresses the inconsistent lines and noisy regions. Second, it reduces the storage requirement for the final feature. Third, it significantly reduces the time required for the extraction of features and matching.

\section{Contour Code}
\label{sec:cnt_code}

\subsection{Multidirectional Feature Encoding}

The nonsubsampled contourlet transform (NSCT)~\cite{da2006nonsubsampled} is a multi-directional expansion with improved directional frequency localization properties and efficient implementation compared to its predecessor, the contourlet transform~\cite{do2005contourlet}. It has been effectively used in basic image processing operations such as image denoising and enhancement. Here, we exploit the directional frequency localization characteristics of the NSCT for multidirectional feature extraction from palmprint images.

\begin{figure}[t]
\centering
\includegraphics[width=1\linewidth]{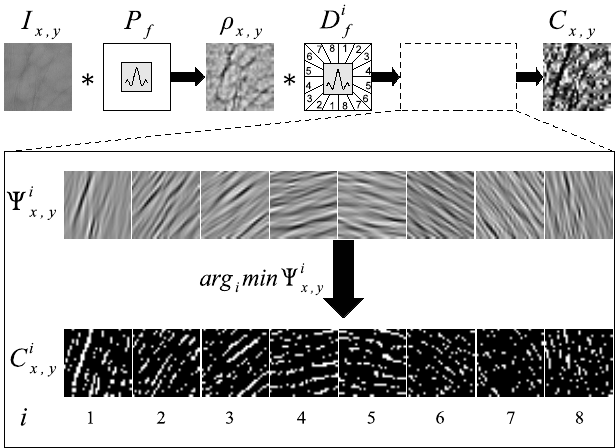}
\caption{Extraction of the Contour Code representation. $\rho$ is the pyramidal bandpass subband and $\Psi^{i}$ are the bandpass directional subbands. The images $C^{i}$ represent the dominant points existing in the $i^{th}$ directional subband. The intensities in $C$ correspond to $i=1,2,\ldots,8$ (from dark to bright).}
\label{fig:ContourCode}
\end{figure}

An ROI $I\in\mathbb{R}^{m\times n}$ of a band is first convolved with a nonsubsampled bandpass pyramidal filter $(P_{f})$ which captures the details in the palm at a single scale as shown in Fig.~\ref{fig:ContourCode}. This filtering operation allows only the robust information in a palm to be passed on to the subsequent directional decomposition stage.
\begin{equation}
\rho = I\ast P_{f}
\end{equation}
The band pass filtered component $(\rho)$ of the input image is subsequently processed by a nonsubsampled directional filter bank $(D_{f})$, comprising $2^{k}$ directional filters.
\begin{equation}
\Psi^{i} = \rho\ast D_{f}^{i}~,
\end{equation}
where $\Psi^{i},i=1,2,\ldots,2^{k}$ is the set of directional subbands. For example, a third order directional filter bank ($k=3$) will result in $8$ directional filtered subbands. Therefore, each directional subband covers an angular region of $\pi/2^{k}$ radians. The combination of pyramidal and directional filter decomposition stages determine their capability to capture line like features. We perform a detailed experimental analysis of pyramidal-directional filter pairs in Section~\ref{sec:filter}.

\begin{figure*}[b]
\centering
\includegraphics[width=0.65\linewidth]{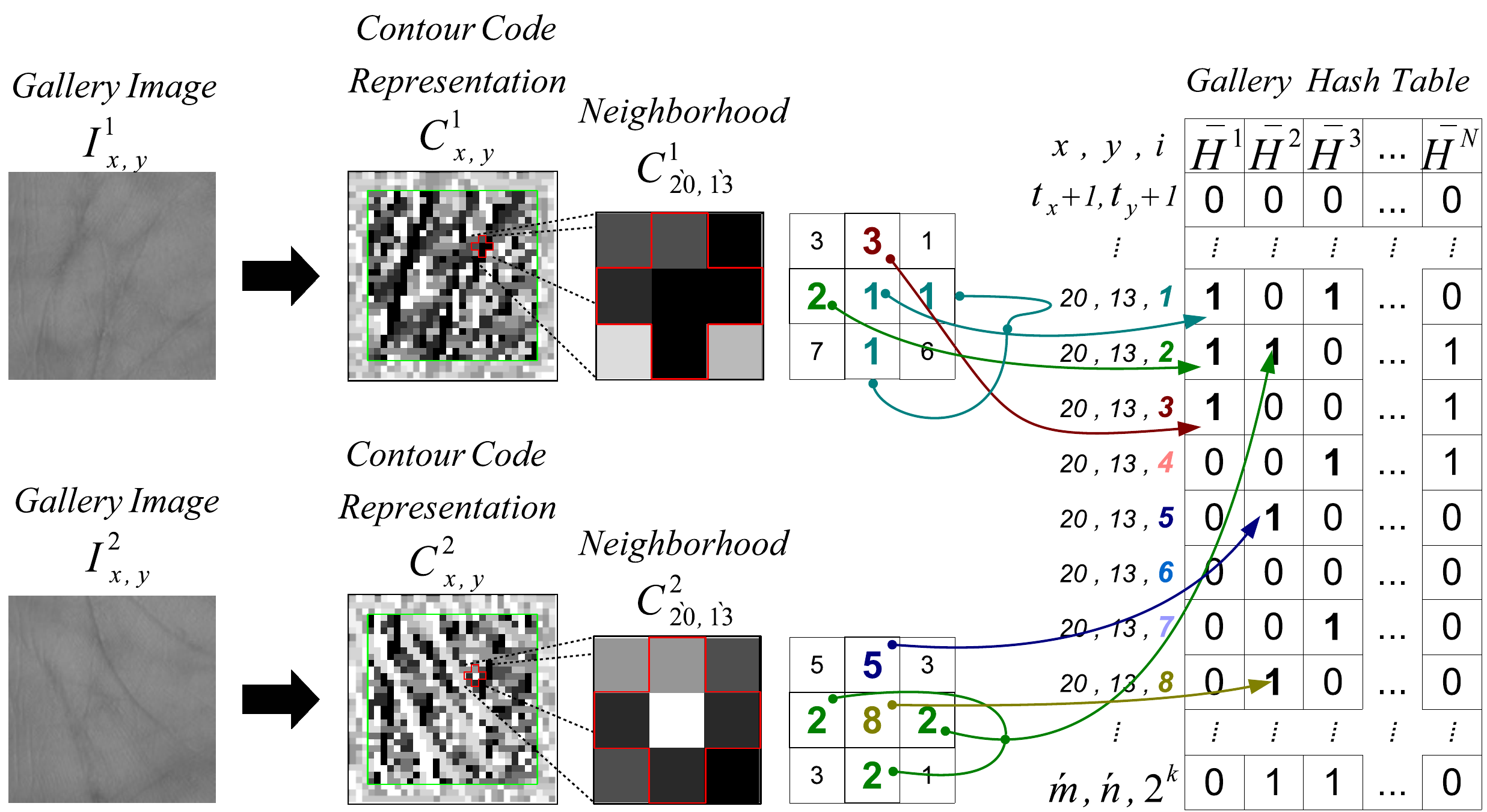}
\caption{Illustration of binary hash table encoding. A galley image $I$ is first converted to its Contour Code representation $C$. We show how a single point $C_{\acute{20},\acute{13}}$ is encoded into the binary hash table based on its \emph{z-connected} neighborhood (z=4 in this example). For $\bar{H}^{1}$, bins 1,2 and 3 are encoded as binary 1 and the rest as 0. Whereas for $\bar{H}^{2}$, bins 2,8 and 5 are encoded as binary 1 and the remaining are set as 0.}
\label{fig:hash_pop}
\end{figure*}

Generally, both the line and vein patterns appear as dark intensities in a palm and correspond to a negative filter response. The orientation of a feature is determined by the coefficient corresponding to the minimum peak response among all directional subbands at a specific point. Let $\Psi^{i}$ denote the coefficient at point $(x,y)$ in the $i^{th}$ directional subband where $i=1,2,3,\ldots,2^{k}$. We define a rule similar to the competitive rule~\cite{kong2004competitive}, to encode the dominant orientation at each $(x,y)$.
\begin{equation}
C=\operatorname*{arg\,min}_i\Psi^{i}~,
\end{equation}
where $C$ is the Contour Code representation of $I$. Similarly, $C$ is computed for all bands of the multispectral image of a palm. An example procedure for a single band of a multispectral palm image is shown in Fig.~\ref{fig:ContourCode}.

\subsection{Binary Hash Table Encoding}

A code with $2^{k}$ orientations requires only $k$ bits for encoding. However, we binarize the Contour Code using $2^{k}$ bits to take advantage of a fast binary code matching scheme. Unlike other methods which use the Angular distance~\cite{kong2004competitive} or the Hamming distance~\cite{hao2008multispectral}, we propose an efficient binary hash table based Contour Code matching. Each column of the hash table refers to a palm's binary hash vector derived from its Contour Code representation. Within a column, each hash location $(x,y)$ has $2^{k}$ bins. We define a hash function so that each Contour Code can be mapped to the corresponding location and bin in the hash table. For an orientation at $(x,y)$, the hash function assigns it to the $i^{th}$ bin according to
\begin{equation}
H_{(xyi)} =
\begin{dcases}
1, & i=C\\
0, & otherwise
\end{dcases}
\label{eq:hash}
\end{equation}
where $H_{(xyi)}$ is the binarized form\footnote{$(xyi)$ represents a single subscript that takes on a distinct value for each combination of an element from the index sets $x,y$ and $i$.} of a Contour Code representation $C$.

Since palmprint features are non rigid, it is not possible to achieve a perfect 1-1 correspondences between all the orientations of two palm images taken at different instances of time. It is, therefore, intuitive to assign multiple orientations to a hash location $(x,y)$ based on its neighborhood. Therefore, the hash function is blurred so that it assigns for each hash bin in $\bar{H}_{(xyi)}$ with all the orientation indices in $(x,y)$ and its neighbors.
\begin{equation}
\bar{H}_{(xyi)} =
\begin{dcases}
1, & \forall~~i=C_{\acute{x},\acute{y}}\\
0, & otherwise
\end{dcases}
\end{equation}
where $\acute{x},\acute{y}$ is the set of $(x,y)$ and its \emph{z-connected} neighbors. The extent of the blur neighborhood $(z)$, determines the robustness of matching. Less blurring will result in a small number of points matched, however, with high confidence. On the other hand, too much blur will result in a large number of points matched but with low confidence.

\begin{figure}[t]
\centering
\subfloat[]{\label{fig:blur-no}\includegraphics[width=0.49\linewidth]{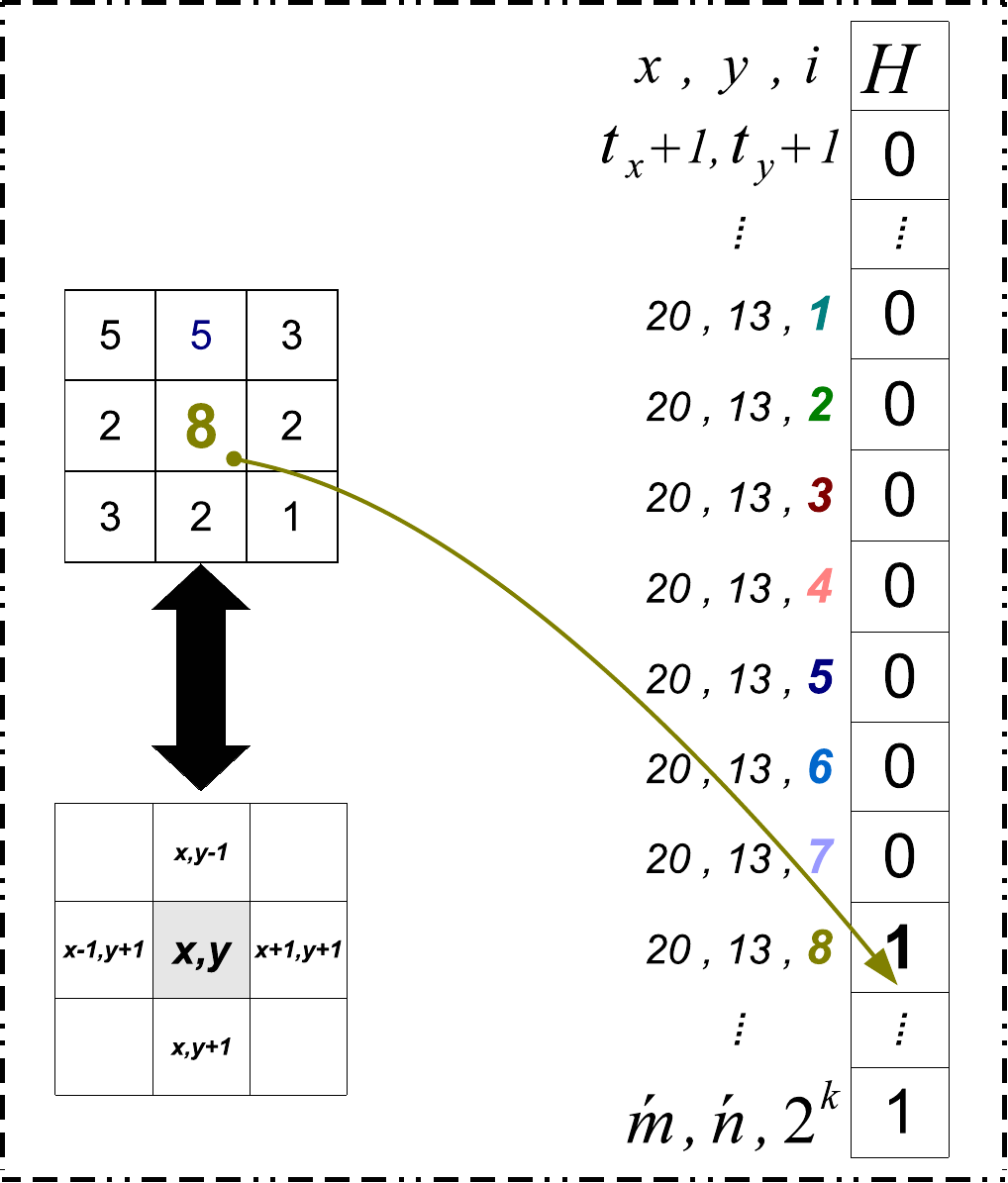}}\hspace{1pt}
\subfloat[]{\label{fig:blur-plus}\includegraphics[width=0.49\linewidth]{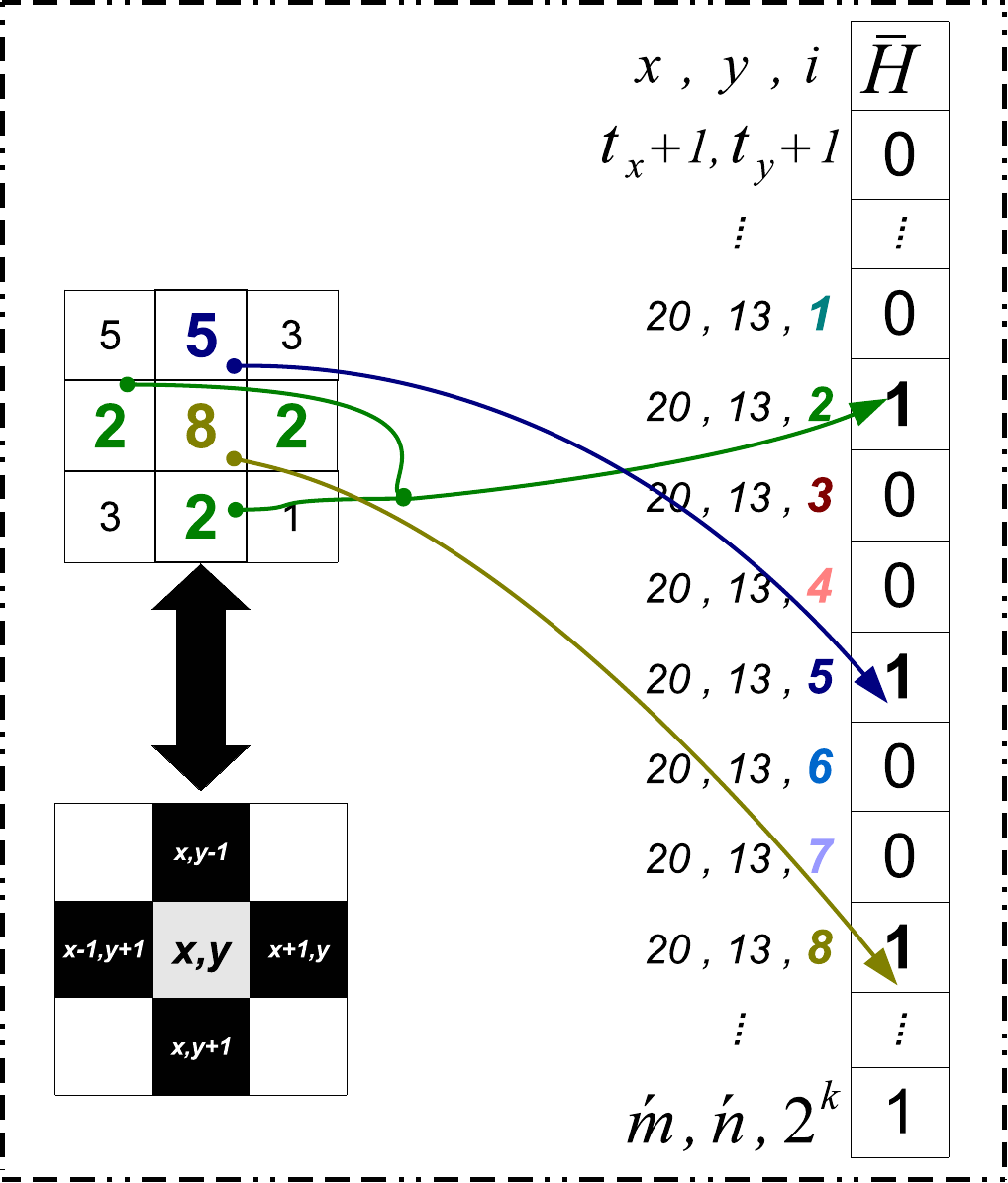}}
\caption{Binary hash table encoding with (a) no blurring, (b) a \emph{4-connected} neighborhood blurring.}
\label{fig:blur-eg}
\end{figure}

Since, the hash table blurring depends on a certain neighborhood as opposed to a single pixel, it robustly captures crossover line orientations. A detailed example of Contour Code binarization using the blurred hash function is given in Fig.~\ref{fig:hash_pop}. Fig.~\ref{fig:blur-eg} illustrates hash table encoding without blurring and with a {\em 4-connected} neighborhood blurring. The discussion of an appropriate blur neighborhood is presented later in Section~\ref{sec:params}.

\subsection{Matching}

\begin{figure}
\centering
\includegraphics[width=1\linewidth]{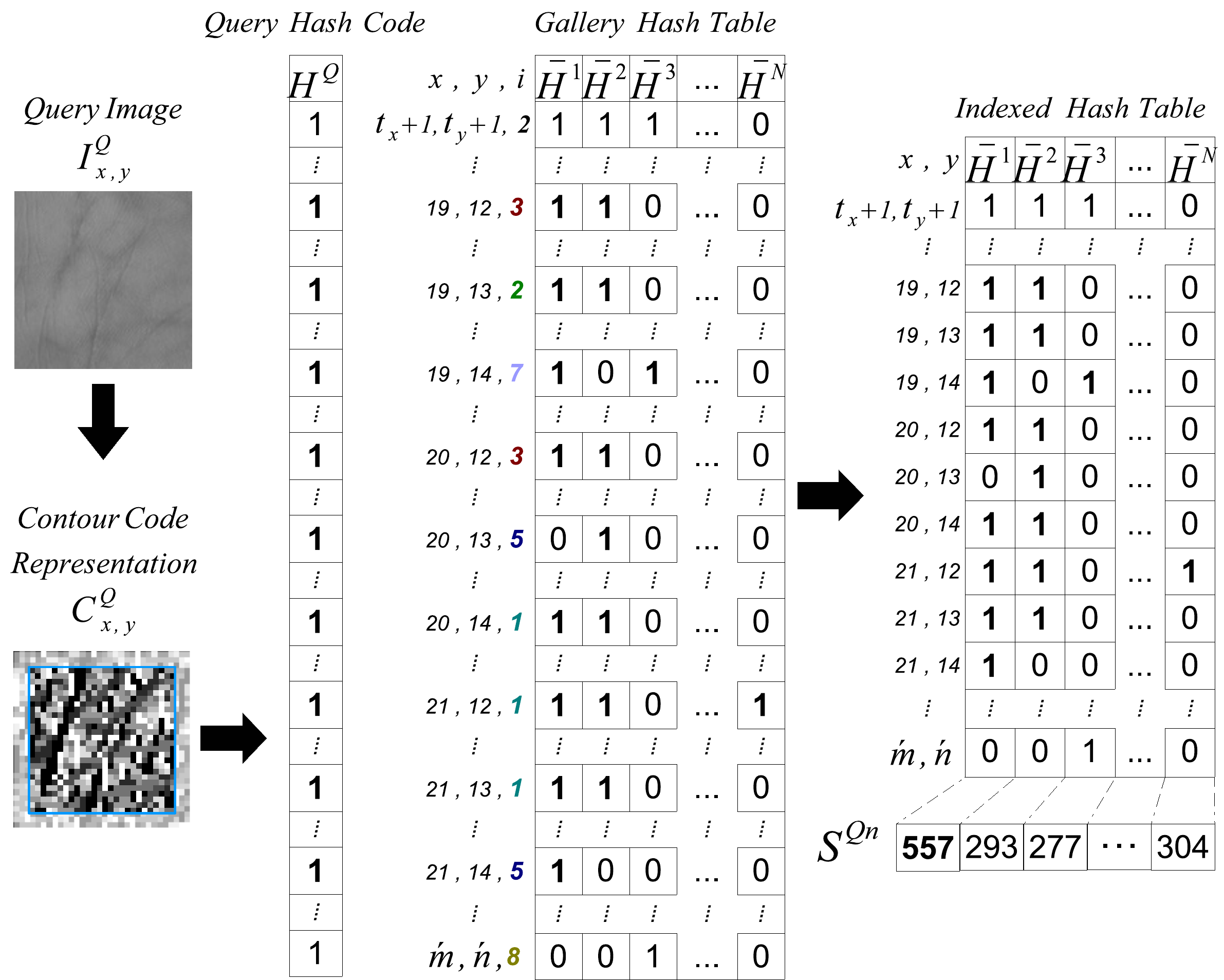}
\caption{Illustration of the Contour Code matching. A query image $I^{Q}$ is first transformed into its Contour Code representation, $C^{Q}$. Unlike a gallery image, a query image is encoded without blurring to get $H^{Q}$, which is subsequently used to index the rows of the gallery hash table. Finally, the columns of the indexed hash table are summed up to obtain the match scores of the query image with the respective gallery images of the hash table. In the above example, $I^{Q}$ is best matched to $I^{1}$ as $\bar{H}^{1}$ has more $1$s in the hash table indexed with $H^{Q}$ resulting in $S^{Q_{1}}$ to be the maximum summation score.}
\label{fig:hash_mtch}
\end{figure}

The binary hash table facilitates simultaneous one-to-many matching for palmprint identification. Fig.~\ref{fig:hash_mtch} illustrates the process. The Contour Code representation $C^{Q}$ of a query image $I^{Q}$ is first converted to the binary form $H_{(xyi)}^{Q}$ using equation \eqref{eq:hash}. No blurring is required now, since it has already been performed offline on all the gallery images. The match score $(S^{Qn})$ between $H_{(xyi)}^{Q}$ and $\bar{H}_{(xyi)}^{n}$, $n=1,2,3\ldots,N$ is computed as
\begin{equation}
S^{Qn}=\|\bar{H}_{(xyi)}^{n}~~~\forall~~~\operatorname*{arg\,}_{(xyi)} H_{(xyi)}^{Q}=1\|_{0}
\label{eq:mtch}
\end{equation}
where $\|.\|_{0}$ is the $\ell_{0}$ norm which is the number of non-zero hash entries in $\bar{H}_{(xyi)}^{n}$ for all $(xyi)$ indices where $H_{(xyi)}^{Q}=1$. The hash table, after indexing, produces a relatively sparse binary matrix which can be efficiently summed. Since, the $\ell_{0}$ norm of a binary vector is equivalent to the summation of all the vector elements, equation \eqref{eq:mtch} can be rewritten as
\begin{equation}
S^{Qn}=\sum \left\{\bar{H}_{(xyi)}^{n}~~~\forall~~~\operatorname*{arg\,}_{(xyi)} H_{(xyi)}^{Q}=1\right\}~.
\end{equation}

\subsubsection{Translated Matches}
Apart from the blurring, which caters for minor orientation location errors within an ROI, during matching, the query image is shifted $\pm t_{x}$ pixels horizontally and $\pm t_{y}$ pixels vertically to cater for the misalignment between different ROIs. The highest score among all translations is considered as the final match. The class of a query palm is determined by the gallery image $n$ corresponding to the best match.
\begin{equation}
\emph{Class}=\operatorname*{arg\,max}_{n} (S^{Qn}_{t_{x},t_{y}})
\end{equation}

We present two variants of matching and report results for both in Section~\ref{sec:exp}. When the bands are translated in synchronization, it is referred to as \emph{Synchronous Translation Matching} (denoted by {ContCode-STM}) and when the bands are independently translated to find the best possible match, it is referred to as \emph{Asynchronous Translation Matching} (denoted by {ContCode-ATM}).

Due to translated matching, the overlapping region of two matched images reduces to ($\acute{m}=m-2t_{x}$,$\acute{n}=n-2t_{y}$) pixels. Therefore, only the central $\acute{m}\times\acute{n}$ region of a Contour Code is required to encoded in the hash table, further reducing the storage requirement. Consequently during matching, a shifting window of $\acute{m}\times\acute{n}$ of the query image is matched with the hash table. We selected $t_{x},t_{y}=3$, since no improvement in match score was observed for translation beyond $\pm3$ pixels. The indexing part of the matching is independent of the database size and only depends on the matched ROI size $(\acute{m}\times\acute{n})$. Additionally, we vertically stacked the hash table entries of all the bands to compute the aggregate match score in a single step for {ContCode-STM}. Thus the score level fusion of bands is embedded within the hash table matching. In the case of {ContCode-ATM}, each band is matched separately and the resulting scores are summed. Since, a match score is always an integer, no floating point comparisons are required for the final decision.

In a verification (1-1 matching) scenario, the query palm may be matched with the palm samples of the claimed identity only. Thus the effective width of the hash table becomes equal to the number of palm samples of the claimed identity but its height remains the same $(2^{k}\acute{m}\acute{n})$.

\section{Experimental Results}
\label{sec:exp}

\subsection{Databases}
\label{sec:data}

The PolyU \cite{polyu} and CASIA \cite{casia} multispectral palmprint databases are two publicly available datasets used in our experiments. Both databases contain low resolution ($<$150 \emph{dpi}) multispectral palm images stored as 8-bit images per band. Each database was acquired in two different sessions. The ROIs, in the PolyU database, are already extracted according to~\cite{zhang2003online}. However, the ROIs are not extracted in the case of the CASIA database and were extracted according to the technique proposed in Section~\ref{sec:roi_ext}.
Note that the CASIA database contains large RST variations since it was acquired using a non-contact sensor. Detailed specifications of the databases are given in Table~\ref{tab:databases}.

\begin{table}[h]
\caption{Overview of the PolyU and CASIA databases.}
\label{tab:databases}
\footnotesize
\begin{center}
\begin{tabular}{l|l|l} \hline
Database                        &   PolyU       &   CASIA           \\ \hline
Sensor                          &  contact 	    & non-contact       \\
Identities                      &   500         &   200             \\
Samples per identity            &   12          &   6               \\
Total MSIs                      &   6000        &   1200            \\
Bands per sample                &   4           &   6               \\
\multirow{2}{*}{Wavelength(nm)} &   470, 525,   &   460, 630, 700,  \\
                                &   660, 880    &   850, 940, White \\ \hline
\end{tabular}
\end{center}
\end{table}

We quantitatively evaluate the ROI extraction accuracy in Section~\ref{sec:roi_acc}. In Section~\ref{sec:params}, we analyze the effect of different parameters and pyramidal-directional filter combinations. Verification experiments are detailed in Section~\ref{sec:roc} and identification experiments are given in Section~\ref{sec:cmc}.

\subsection{ROI Extraction Accuracy}
\label{sec:roi_acc}

To the best of our knowledge, we are the first to perform a quantitative analysis of ROI extraction from palm images. Our evaluation methodology and results can serve as a benchmark for future techniques. The accuracy of automatic landmark localization was determined by comparison with a manual identification of landmarks for 200 multispectral images in the CASIA database. Manual selections were averaged over the six bands to minimize human error. The spatial deviation of the automatically detected landmarks from the manually marked ground truth was evaluated. The rotation and scale deviation, as a consequence of the deviation in localization, was also computed and analyzed.

\begin{figure}[t]
\begin{center}
\subfloat[]{\label{fig:cum_trans}\includegraphics[trim = 0pt 5pt 50pt 20pt, clip, width=0.33\linewidth]{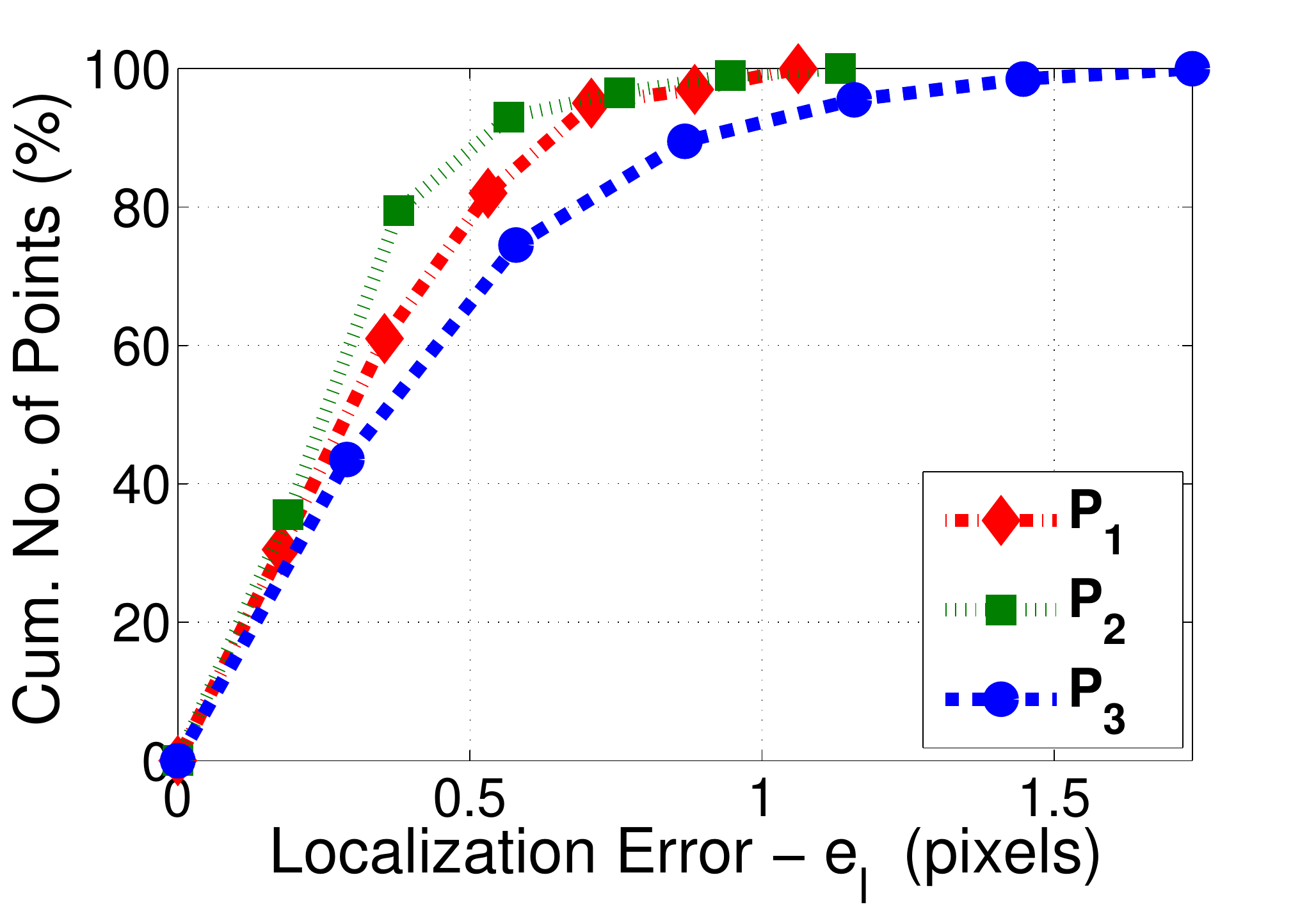}}
\subfloat[]{\label{fig:cum_rot}\includegraphics[trim = 0pt 5pt 50pt 20pt, clip, width=0.33\linewidth]{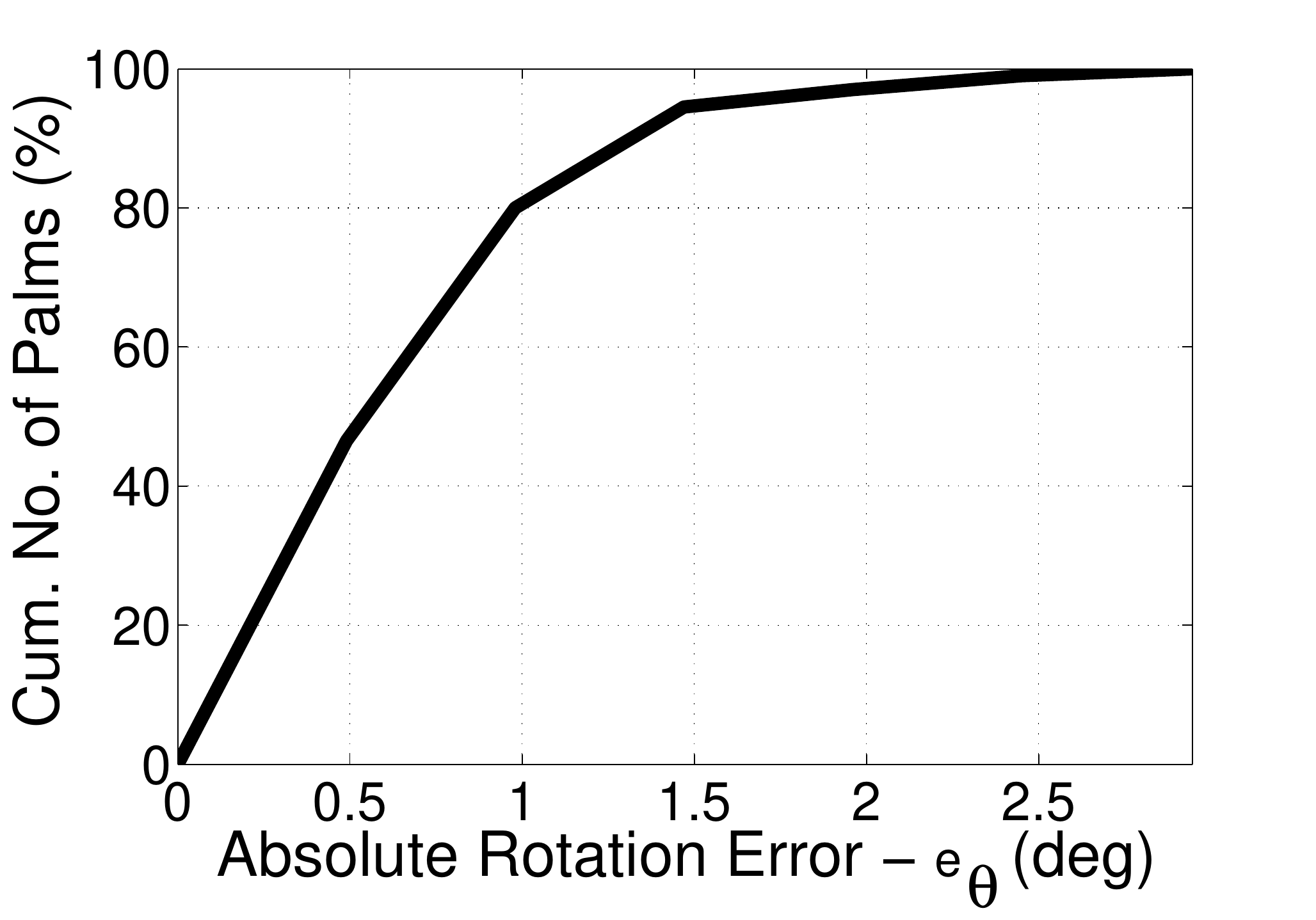}}
\subfloat[]{\label{fig:cum_scale}\includegraphics[trim = 0pt 5pt 50pt 20pt, clip, width=0.33\linewidth]{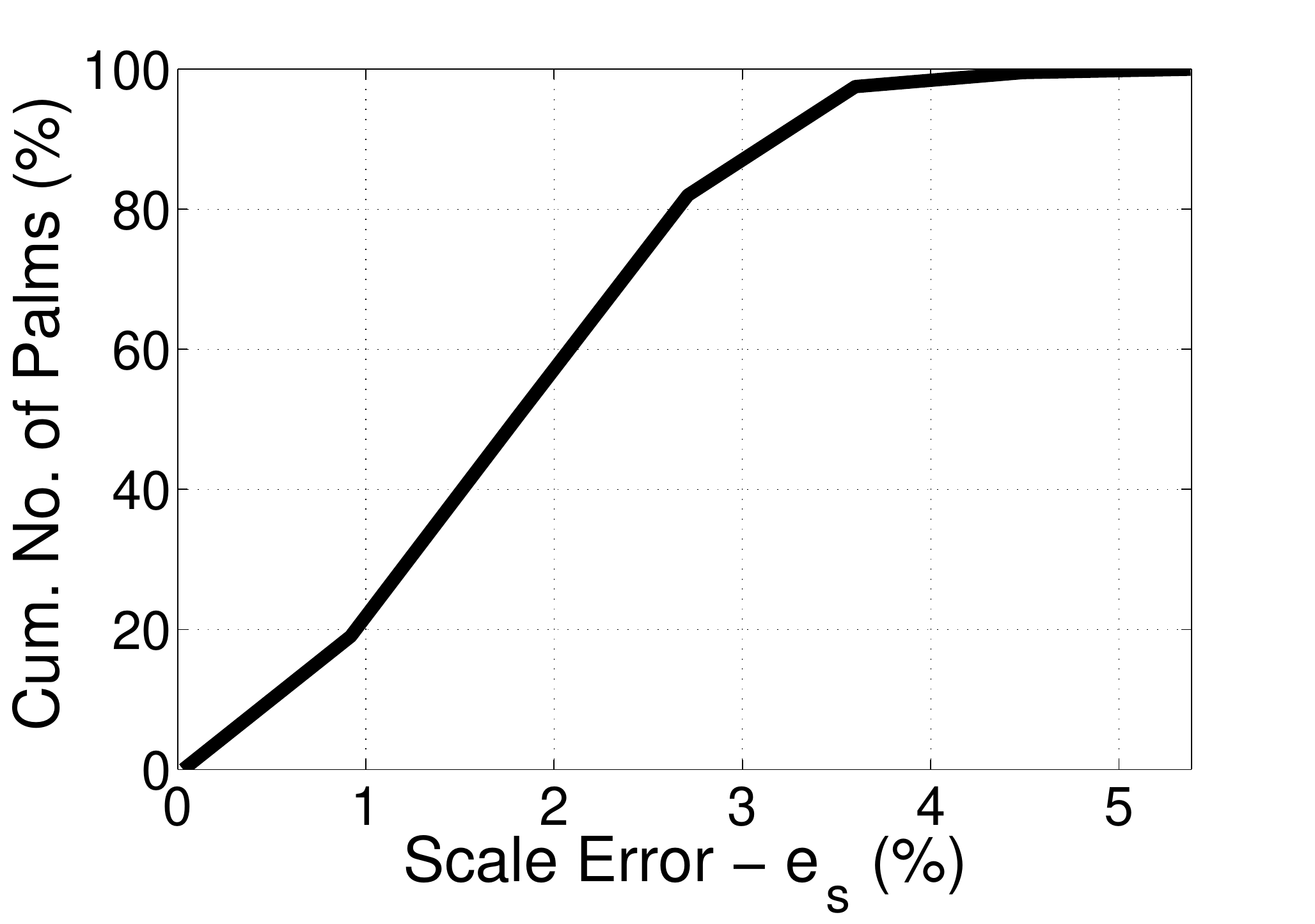}}
\end{center}
\caption{Evaluation of ROI extraction accuracy. (a) Cumulative percentage of landmarks within a localization error $(e_{\ell})$. (b) Cumulative percentage of palm samples within an absolute rotation error $(e_{\theta})$. (c) Cumulative percentage of palm samples within a scale error $(e_{s})$.}
\label{fig:cum_tsr}
\end{figure}

We define the localization error as the Chebyshev distance between the manually selected and the automatically extracted landmarks. The localization error $(e_{\ell})$, between two landmarks is computed as
\begin{equation}
e_{\ell}=\max(|x-\bar{x}|,|y-\bar{y}|)\times \hat{S}~,
\end{equation}
where $(x,y)$, $(\bar{x},\bar{y})$ correspond to the manually and automatically identified landmark coordinates respectively. The $e_{\ell}$ is 1 if the located landmark falls within the first 8 neighboring pixels, 2 if it falls in the first 24 neighboring pixels and so on. Due to scale variation, the $e_{\ell}$ between different palm images could not be directly compared. For example, a localization error of 5 pixels in a close-up image may correspond to only a 1 pixel error in a distant image. To avoid this, we normalize $e_{\ell}$ by determining it at the final size of the ROI. The normalization factor $\hat{S}$, is the side length of ROI at the original scale, $0.7\times \bar{w}$ divided by the final side length of the ROI ($m=32$).

The absolute rotation error $e_{\theta}$ between two palm samples is defined as
\begin{equation}
e_{\theta}=|\theta-\bar{\theta}|~,
\end{equation}
where, $\theta$ and $\bar{\theta}$ correspond to the angle of rotation of a palm computed from manual and automatic landmarks respectively. Finally, the percentage scale error $e_{s}$ is defined as
\begin{equation}
e_{s}=\left(\max\left(\frac{w}{\bar{w}},\frac{\bar{w}}{w}\right)-1\right)\times100 \%~,
\end{equation}
where $w$ is the manually identified palm width averaged over three measurements, while $\bar{w}$ is automatically calculated.

Fig.~\ref{fig:cum_trans} shows the error, in pixels, of landmarks $P_{1},P_{2},P_{3}$. It can be observed that the three landmarks are correctly located within an $e_{\ell}\le2$ for all of the samples. It is important to emphasize that $P_{1},P_{2}$ which are actually used in the ROI extraction are more accurately localized ($e_{\ell}\simeq$ 1 pixel) compared to $P_{3}$. Thus, our ROI extraction is based on relatively reliable landmarks.

Fig.~\ref{fig:cum_rot} shows the absolute rotation error of the automatically extracted ROIs. All the ROIs were extracted within a $e_{\theta} \le 3^{\circ}$. The proposed ROI extraction is robust to rotational variations given the overall absolute rotational variation, $(\mu_{\theta},\sigma_{\theta})\!=\!(20.7^{\circ},7.24^{\circ})$ of the CASIA database. The proposed technique is able to estimate the palm scale within an error of $\pm5.4\%$ (Fig.~\ref{fig:cum_scale}).

\begin{figure}[h]
\begin{center}
\begin{minipage}[b]{0.3\linewidth}
\includegraphics[trim = 64pt 0pt 0pt 0pt, clip, width=1\linewidth]{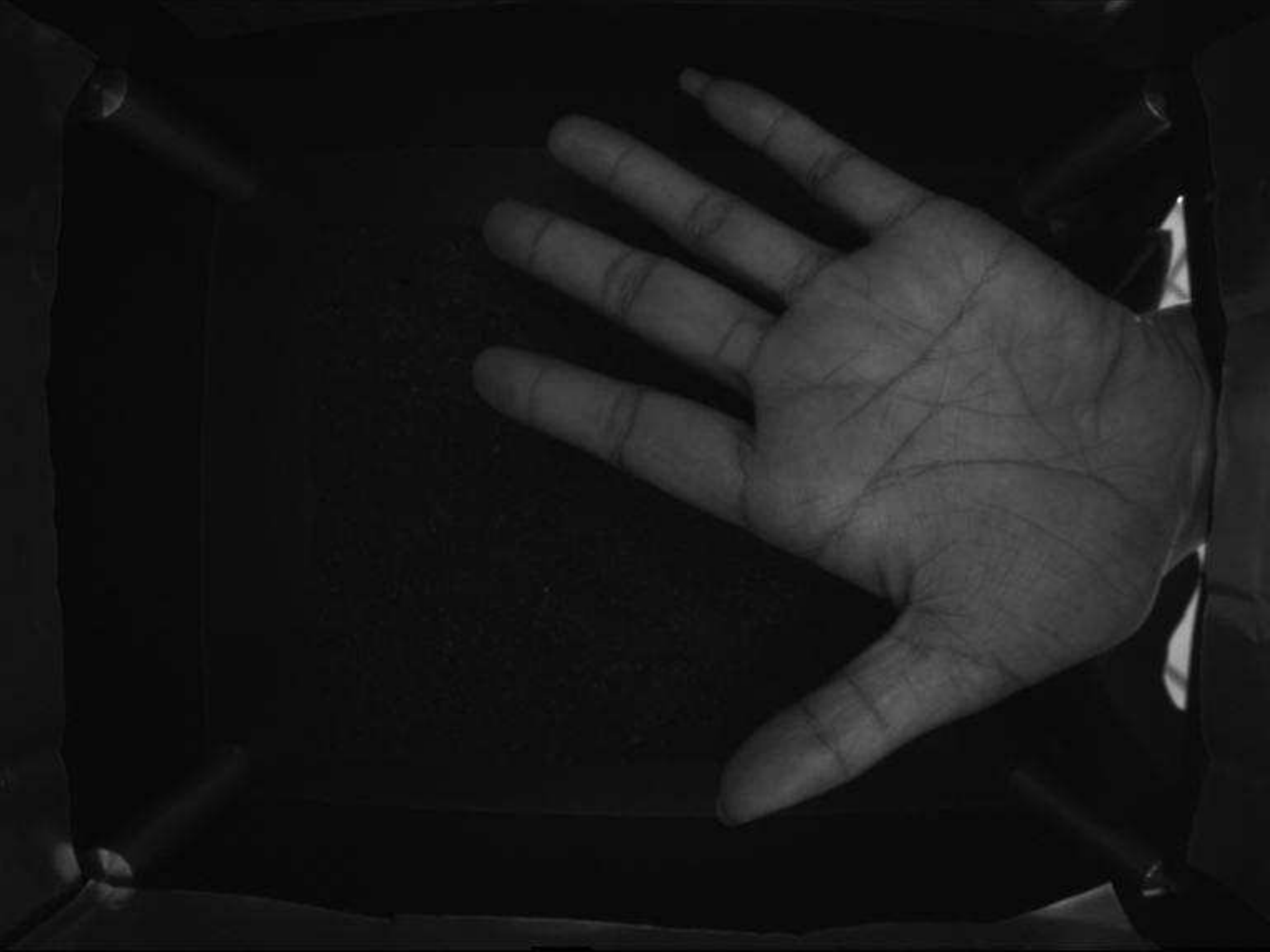}\\
\subfloat[]{\includegraphics[trim = 64pt 0pt 0pt 0pt, clip, width=1\linewidth]{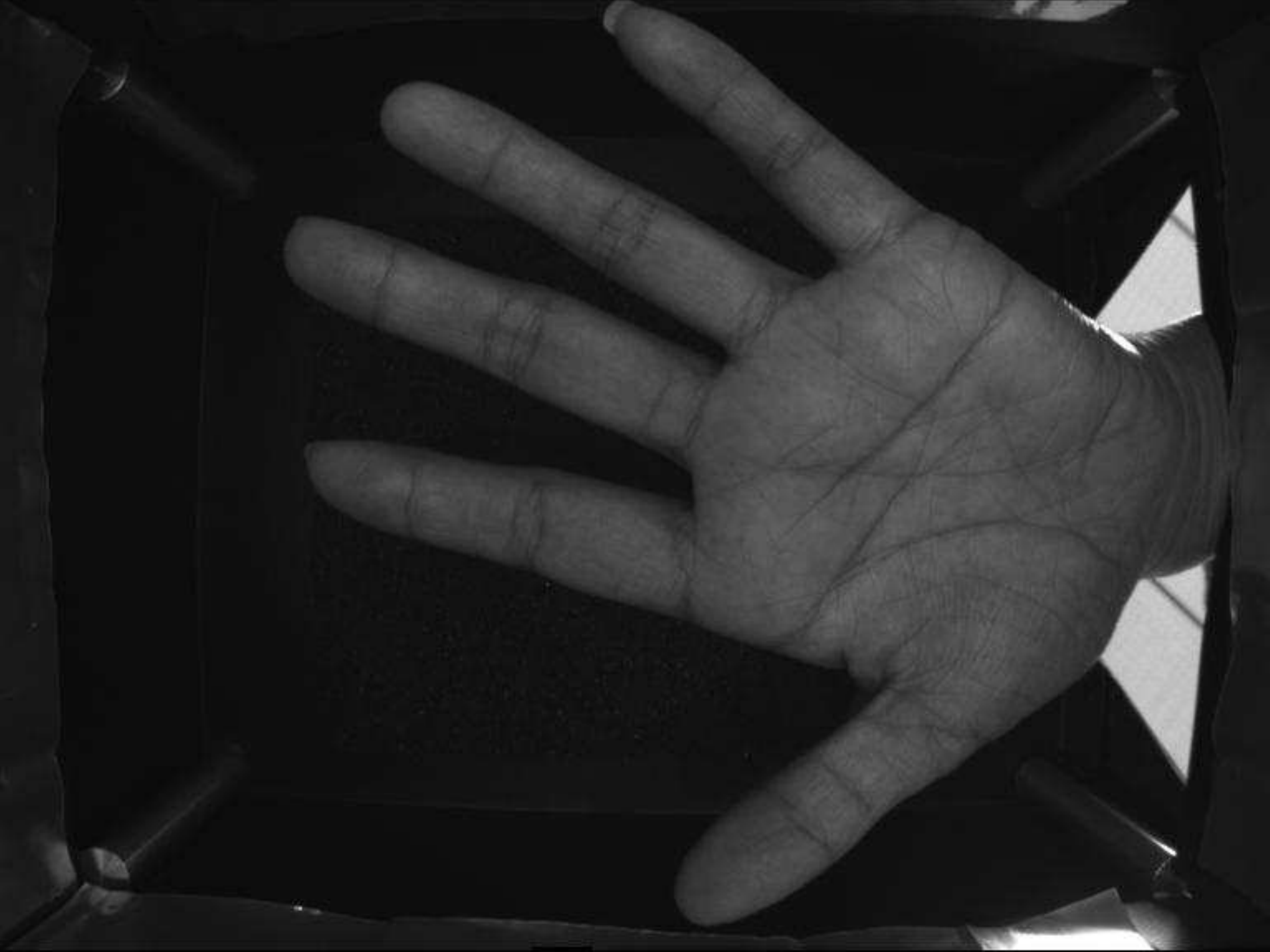}}
\end{minipage}\hspace{0.5pt}
\begin{minipage}[b]{0.3\linewidth}
\includegraphics[width=1\linewidth]{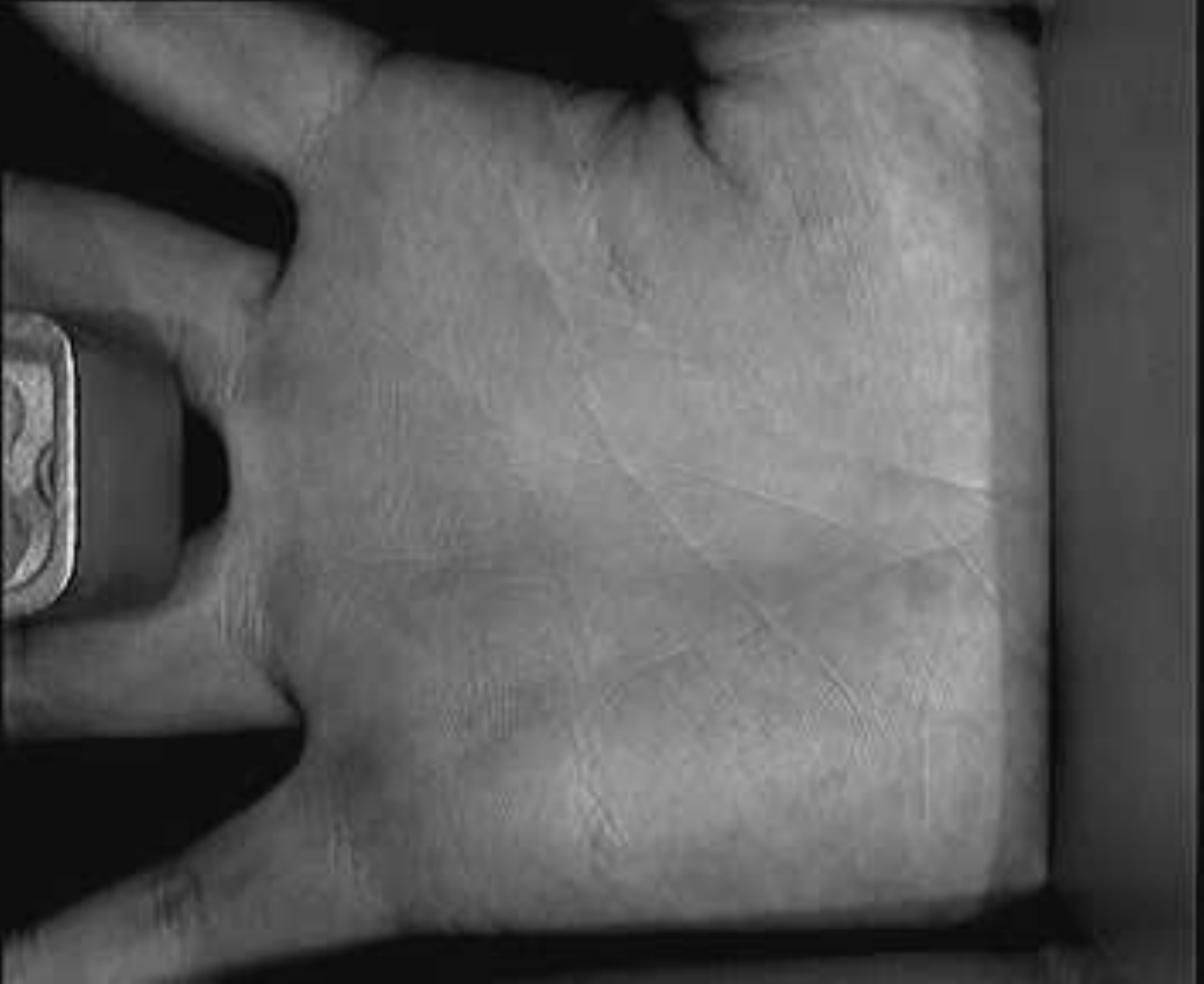}\\
\subfloat[]{\includegraphics[width=1\linewidth]{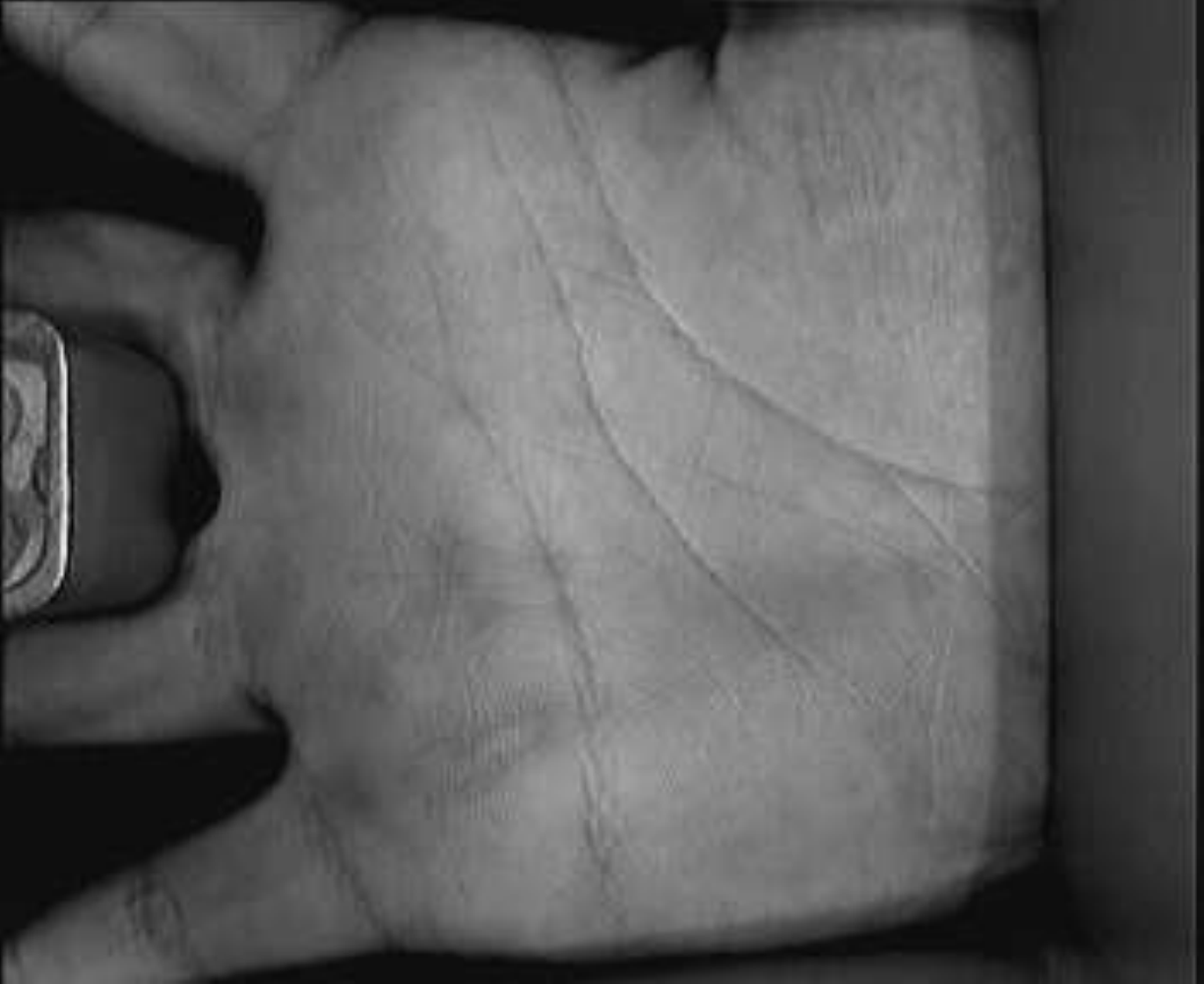}}
\end{minipage}
\end{center}
\caption{Examples of improper hand presentation. (a) A palm correctly presented to a non-contact sensor (top), and out of plane rotated (bottom), resulting in incorrect estimatation of the palm scale. (b) A palm correctly presented to a contact sensor (top), and deformed due to excessive pressure (bottom). This may result in a different ROI and with inconsistent spacing between the palm lines.}
\label{fig:fail}
\end{figure}

ROI extraction from palm images acquired with non-contact sensors must overcome the challenges of RST variation. The proposed technique addresses these challenges except for excessive out-of-plane rotations or palm deformations. Fig.~\ref{fig:fail} shows examples of images that resulted in a low match score due to out-of-plane rotation and excessive deformation of the second (probe) palm. We noted that these two anomalies are a major source of error in matching. In such cases, additional information is required to correct the out-of-plane rotation error and remove the non-rigid deformations. One solution is to use 3D information~\cite{li2010efficient,zhang2009palmprint}. However, the discussion of such techniques is out of the scope of this paper.

\subsection{Parameter Analysis}
\label{sec:params}
In this section, we examine the effects of various parameters including, ROI size, number of Contour Code orientations, hash table blurring neighborhood and pyram-idal-directional filter pair combination. All other parameters are kept fixed during the analysis of a particular parameter. These experiments are performed on a sample subset (50\%) of the PolyU multispectral palmprint database comprising equal proportion of images from the $1^{st}$ and $2^{nd}$ session. The same optimal parameters were used for the CASIA database. For the purpose of analysis, we use the {ContCode-ATM} technique.

\subsubsection{ROI Dimension}
\begin{figure}[t]
\centering
\subfloat[]{\label{fig:params-dim}\includegraphics[trim = 0pt 0pt 10pt 5pt, clip, width=0.33\linewidth]{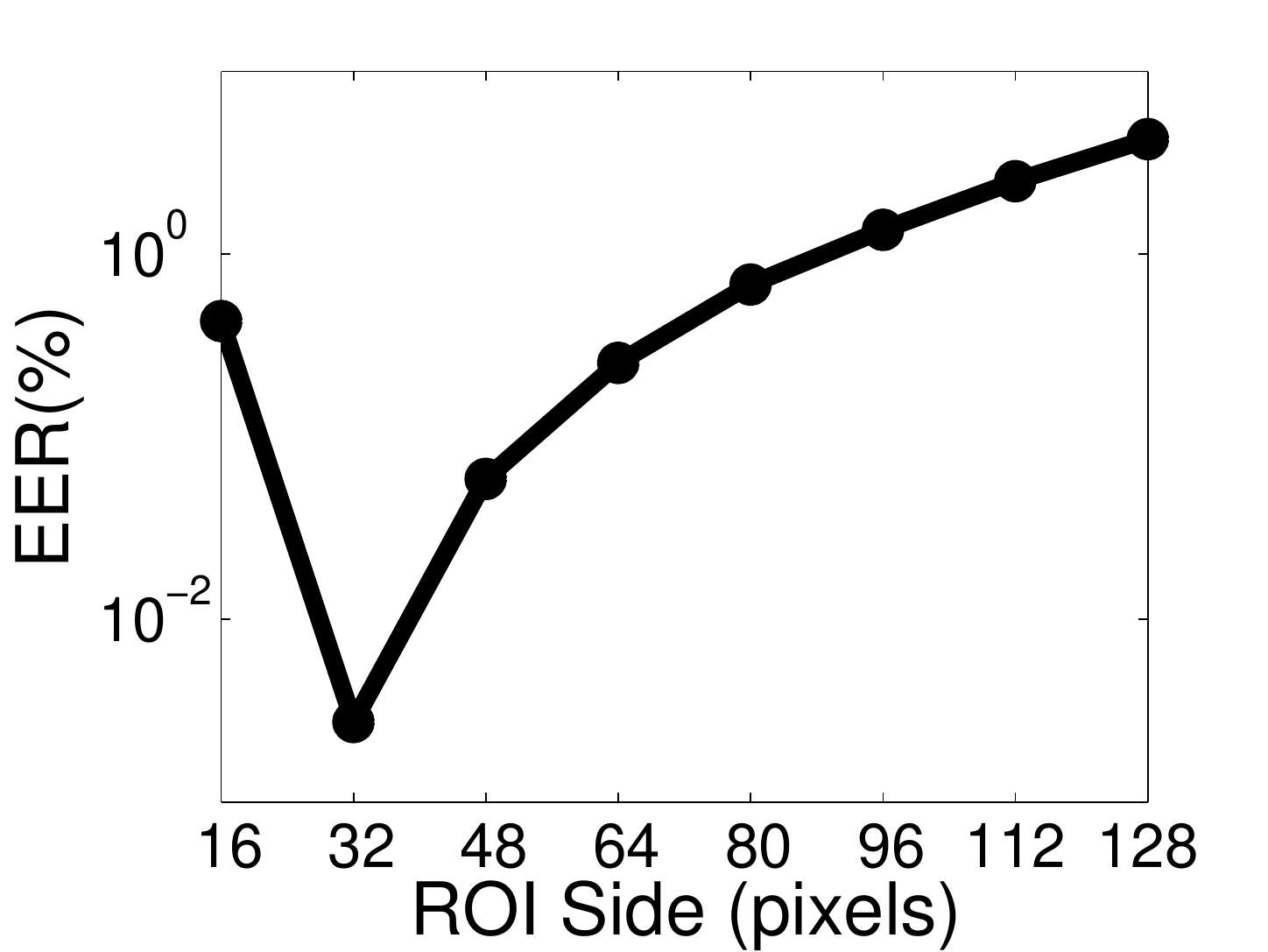}}
\subfloat[]{\label{fig:params-orient}\includegraphics[trim = 0pt 0pt 10pt 5pt, clip, width=0.33\linewidth]{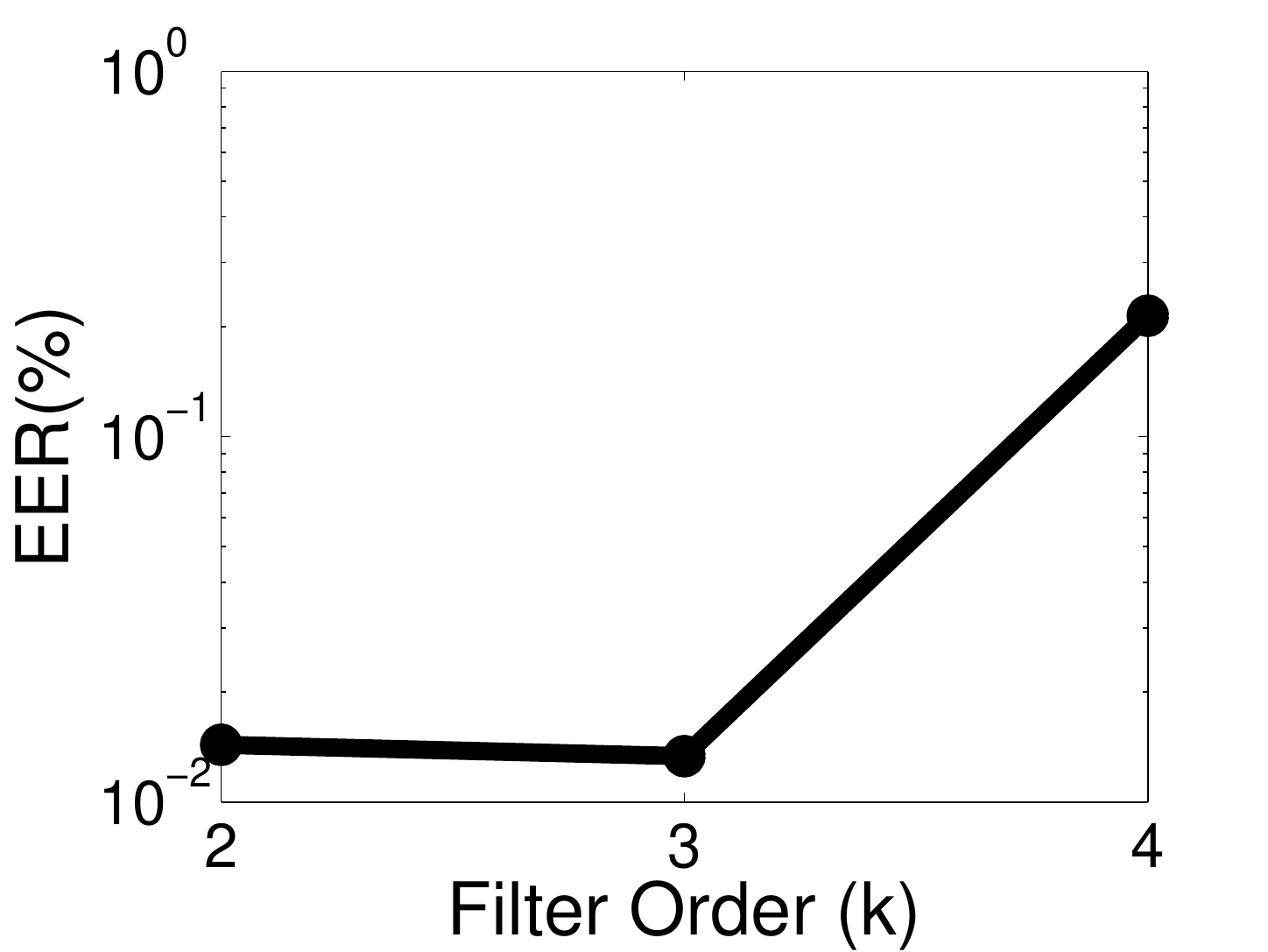}}
\subfloat[]{\label{fig:params-blur}\includegraphics[trim = 0pt 0pt 10pt 5pt, clip, width=0.33\linewidth]{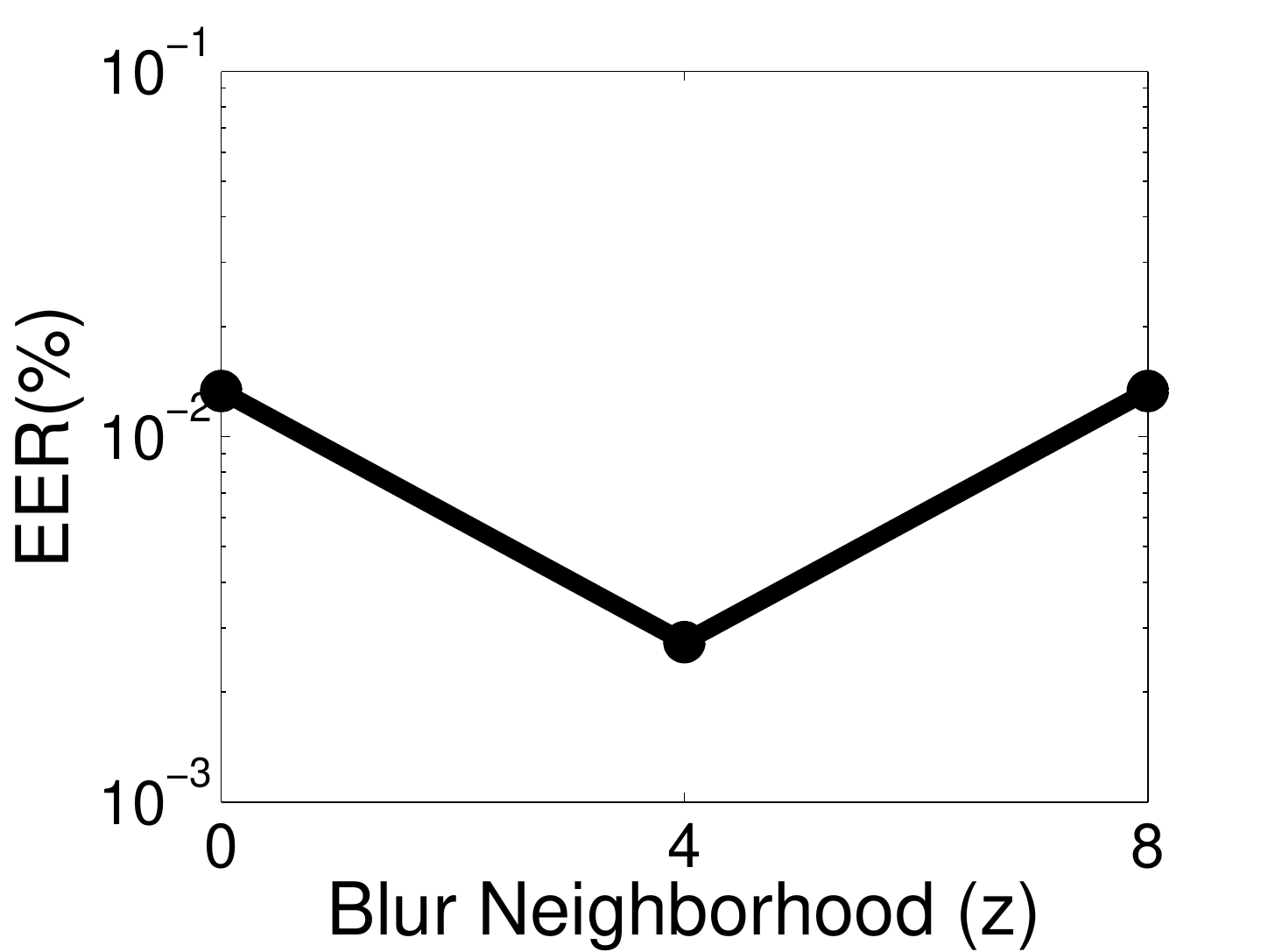}}
\caption{Analysis of parameters (a) ROI dimensions $(m,n)$ (b) Filter Order $(k)$ (c) Blur Neighbourhood $(z)$}
\label{fig:params}
\end{figure}

Palmprint features have varying length and thickness. A larger ROI size may include unnecessary detail and unreliable features. A smaller ROI, on the other hand, may leave out important discriminative information. We empirically find the best ROI size by testing square $(m=n)$ ROI regions of side ${16,32,48,\ldots,128}$. The results in Fig.\ref{fig:params-dim} show that a minimum EER is achieved at 32 after which the EER increases. We use an ROI of $(m,n)=(32,32)$ in all our remaining experiments. Note that a peak performance at such a small ROI is in fact favorable for the efficiency of the Contour Code representation and subsequent matching.

\subsubsection{Orientation Fidelity}
The orientations of the dominant feature directions at points are quantized into a certain number of bins in the Contour Code. A greater number of bins offers better fidelity but also increases the size of the Contour Code representation and its sensitivity to noise. The minimum number of orientation bins that can achieve maximum accuracy is naturally a preferred choice. The orientations are quantized into $2^{k}$ bins, where $k$, the order of the directional filter determines the number of orientations. Fig.~\ref{fig:params-orient} shows the results of our experiments for $k={2,3,4}$. We observe that $k=3$ (i.e.~$2^3=8$ directional bins) minimizes the EER. Although, there is a small improvement from $k=2$ to $k=3$, we prefer the latter as it provides more orientation fidelity which supports the process of hash table blurring. Therefore, we set $k=3$ in all our remaining experiments.

\subsubsection{Hash Table Blur Neighborhood}
Hash table blurring is performed to cater for small misalignments given that the palm is not a rigid object. However, too much blur can result in incorrect matches. We analyzed three different neighborhood types for blurring. The results are reported in Fig.~\ref{fig:params-blur} which show that the lowest error rate is achieved with the \emph{4-connected} blur neighborhood. This neighborhood is, therefore, used in all of the following experiments.

\subsubsection{Pyramidal-Directional Filter Pair}
\label{sec:filter}

An appropriate pyramidal and directional filter pair combination is critical to robust feature extraction. It is important to emphasize that the ability of a filter to capture robust line like features in a palm should consequently be reflected in the final recognition accuracy. Hence, we can regard the pyramidal-directional filter combination with the lowest EER as the most appropriate.

\begin{figure}[h]
\centering
\subfloat{\includegraphics[trim = 60pt 10pt 60pt 5pt, clip,width=0.2\linewidth]{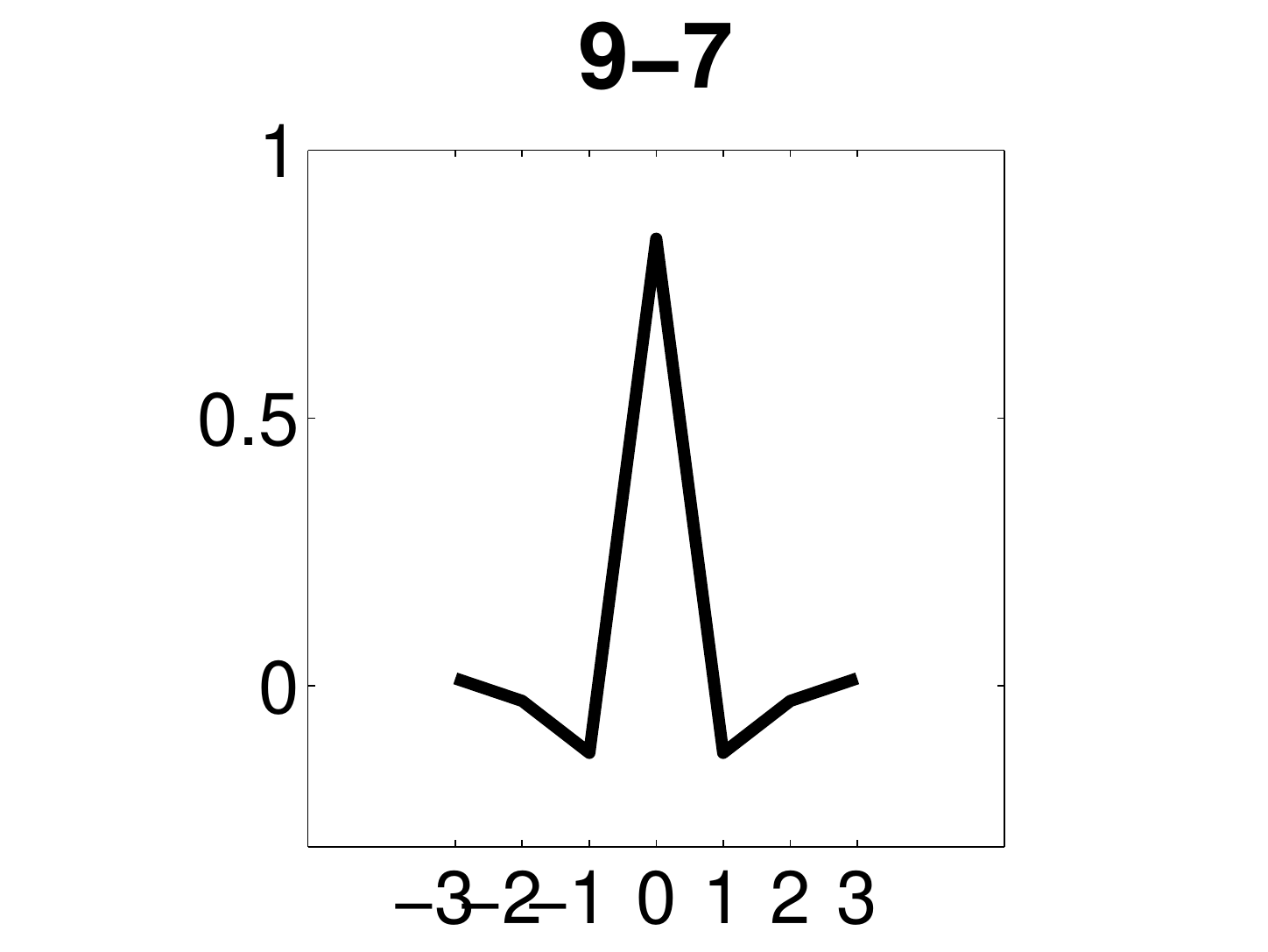}}
\subfloat{\includegraphics[trim = 60pt 10pt 60pt 5pt, clip,width=0.2\linewidth]{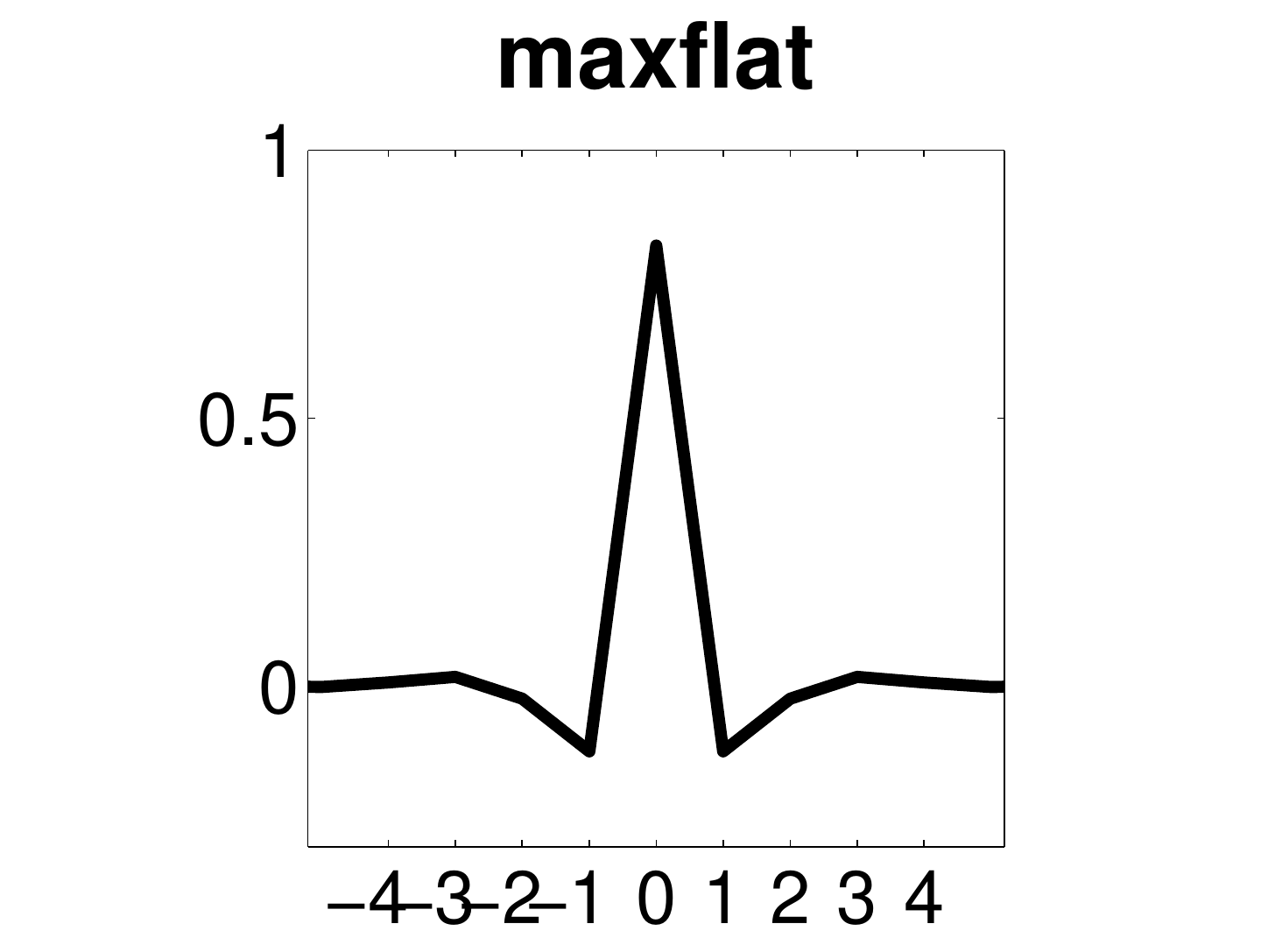}}
\subfloat{\includegraphics[trim = 60pt 10pt 60pt 5pt, clip,width=0.2\linewidth]{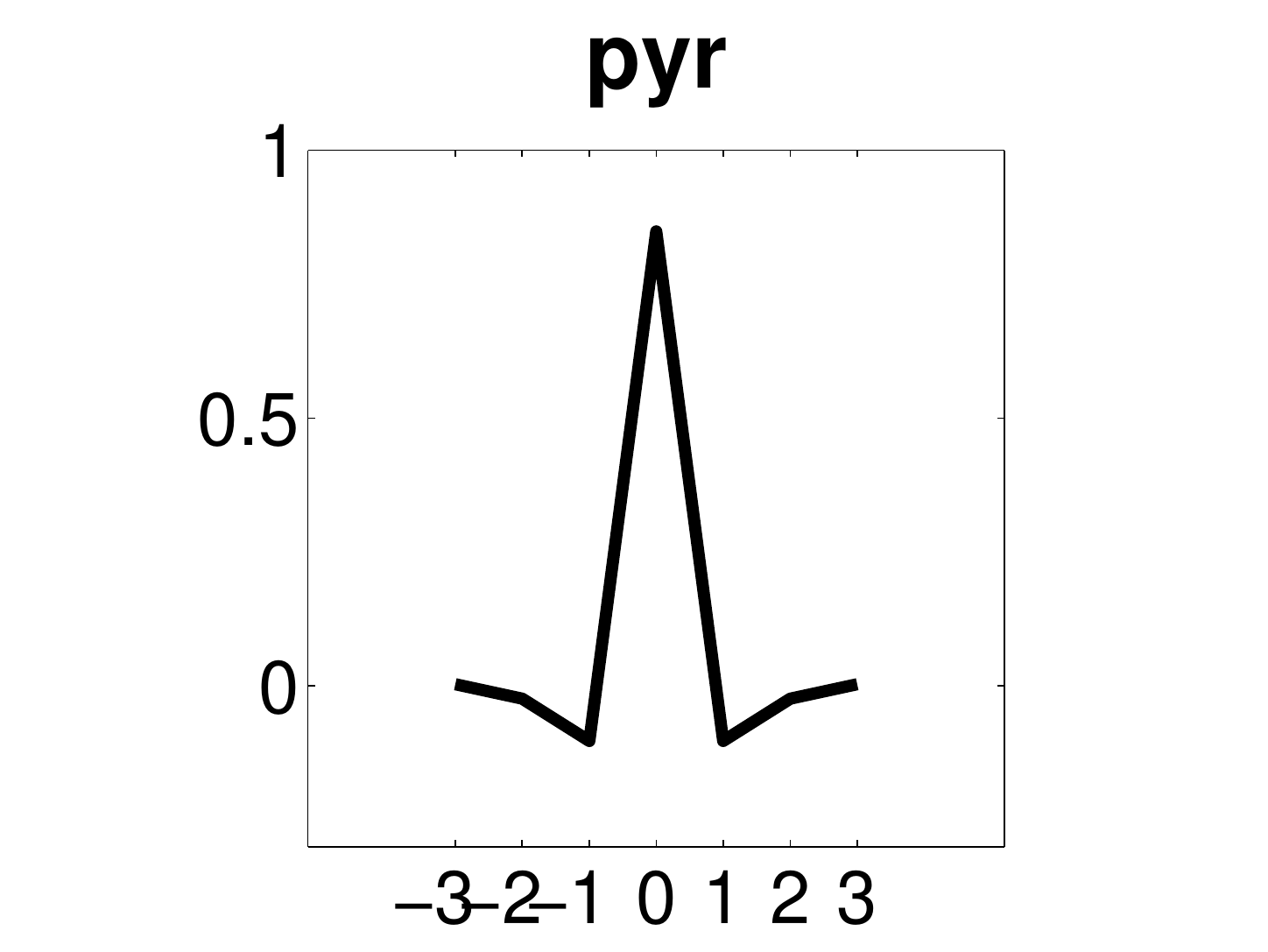}}
\subfloat{\includegraphics[trim = 60pt 10pt 60pt 5pt, clip,width=0.2\linewidth]{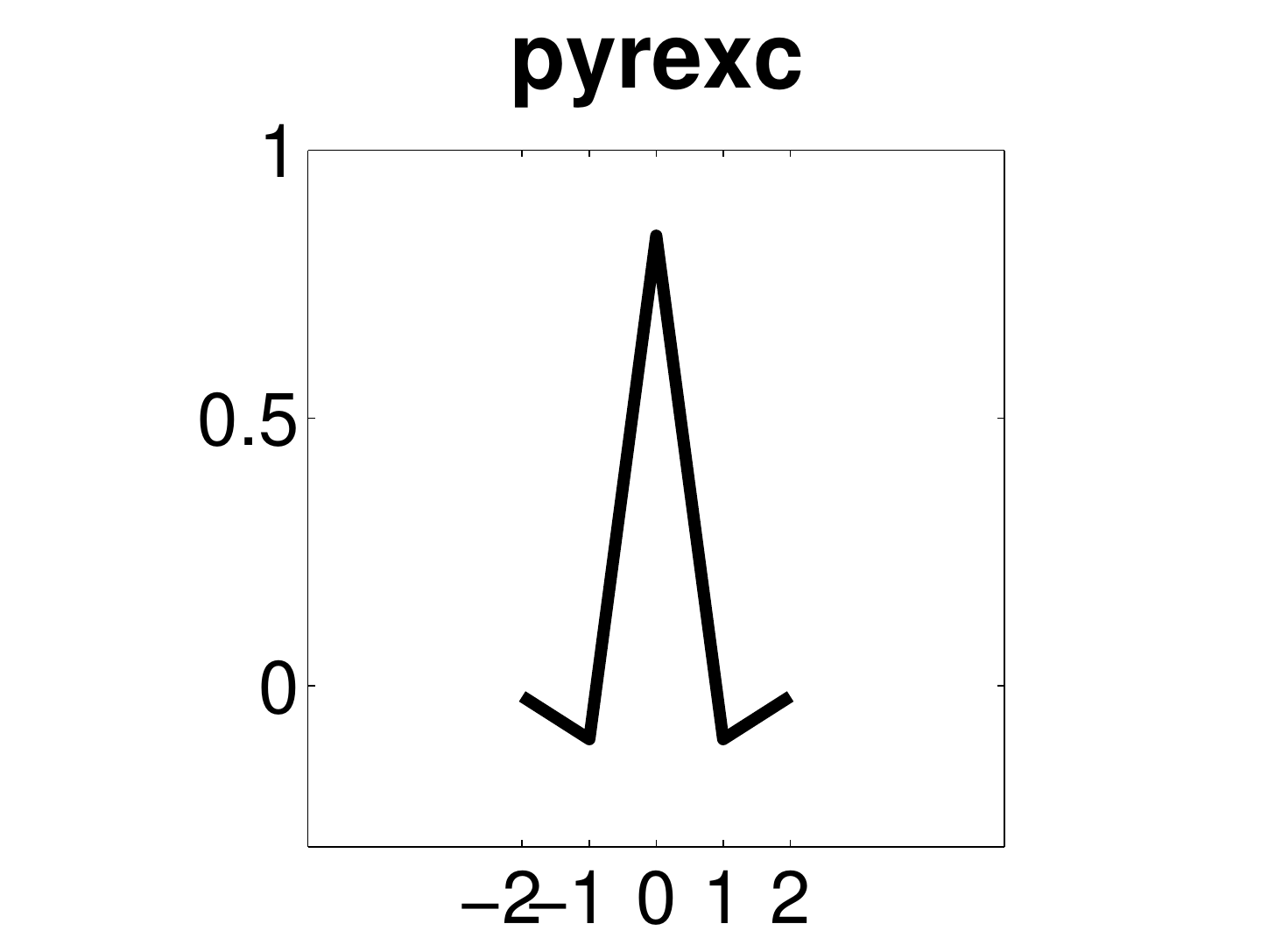}}
\caption{NSCT pyramidal highpass filters. Planar profile of the 2D filters are shown for better visual comparison. (a) Filters from 9-7 1-D prototypes. (b) Filters derived from 1-D using maximally flat mapping function with 4 vanishing moments. (c) Filters derived from 1-D using maximally flat mapping function with 2 vanishing moments. (d) Similar to pyr but exchanging two highpass filters.}
\label{fig:pfilters}
\end{figure}

\begin{figure}[t]
\centering
\subfloat{\includegraphics[trim = 60pt 10pt 60pt 5pt, clip,width=0.20\linewidth]{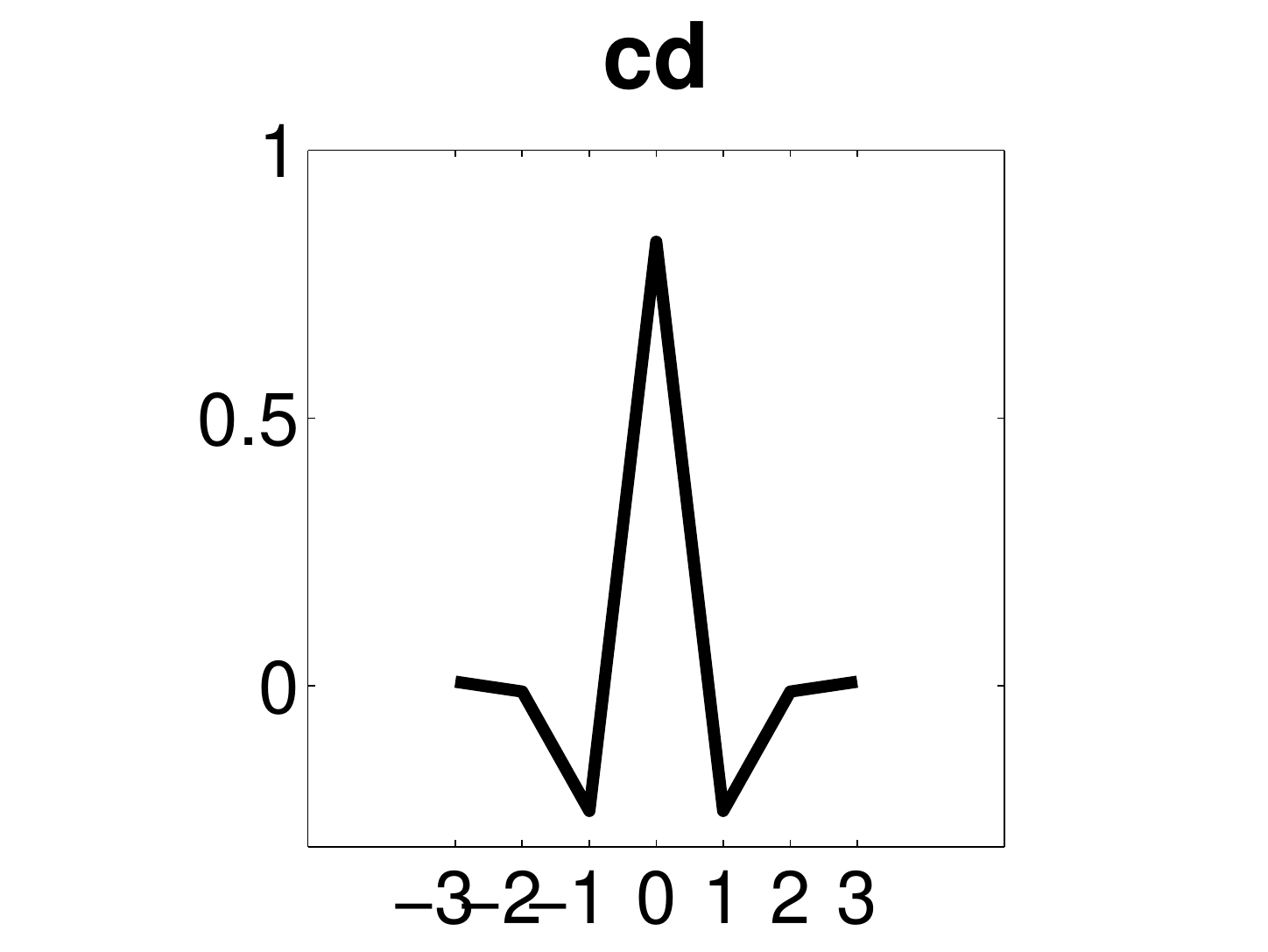}}
\subfloat{\includegraphics[trim = 60pt 10pt 60pt 5pt, clip,width=0.20\linewidth]{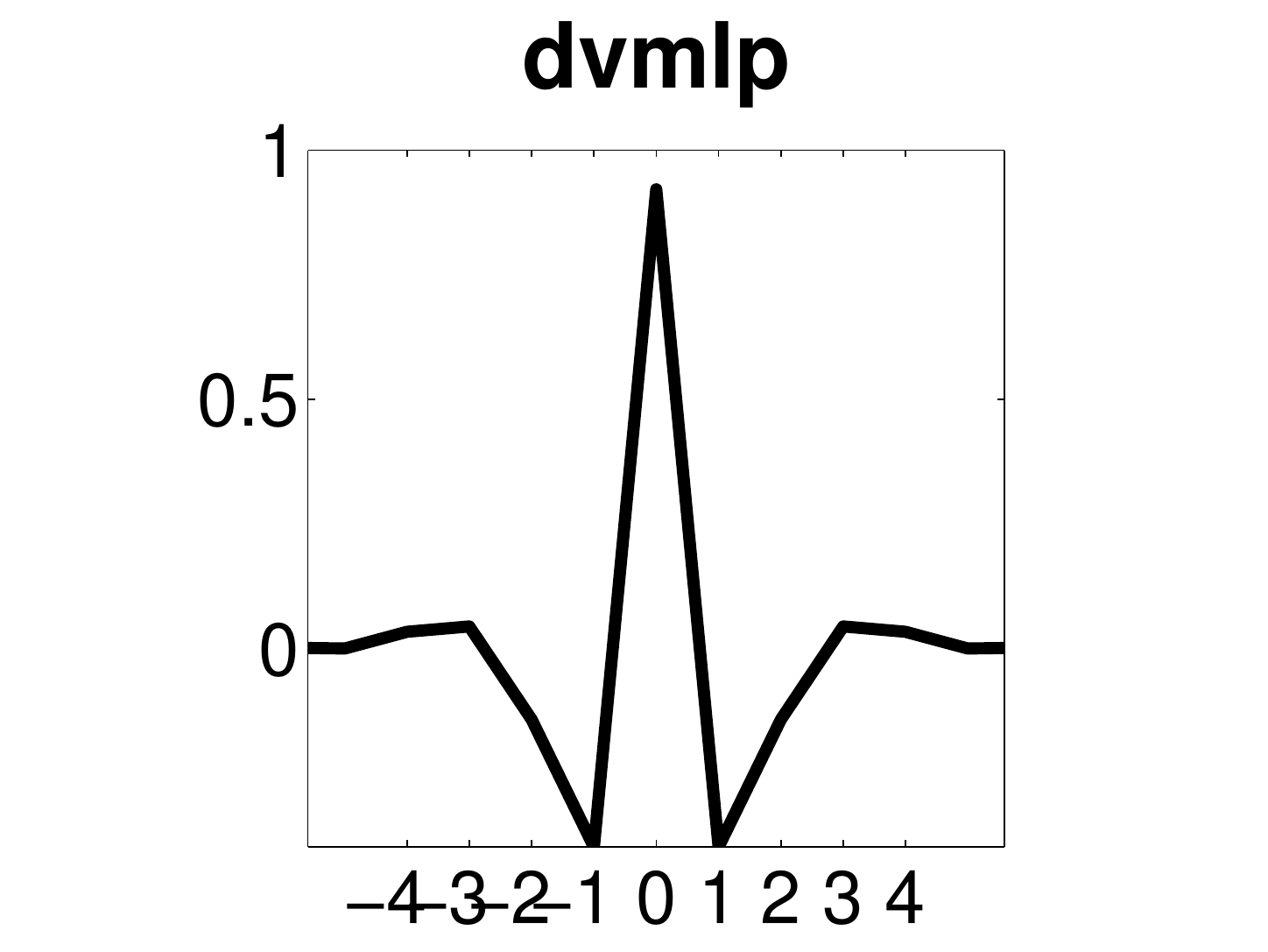}}
\subfloat{\includegraphics[trim = 60pt 10pt 60pt 5pt, clip,width=0.20\linewidth]{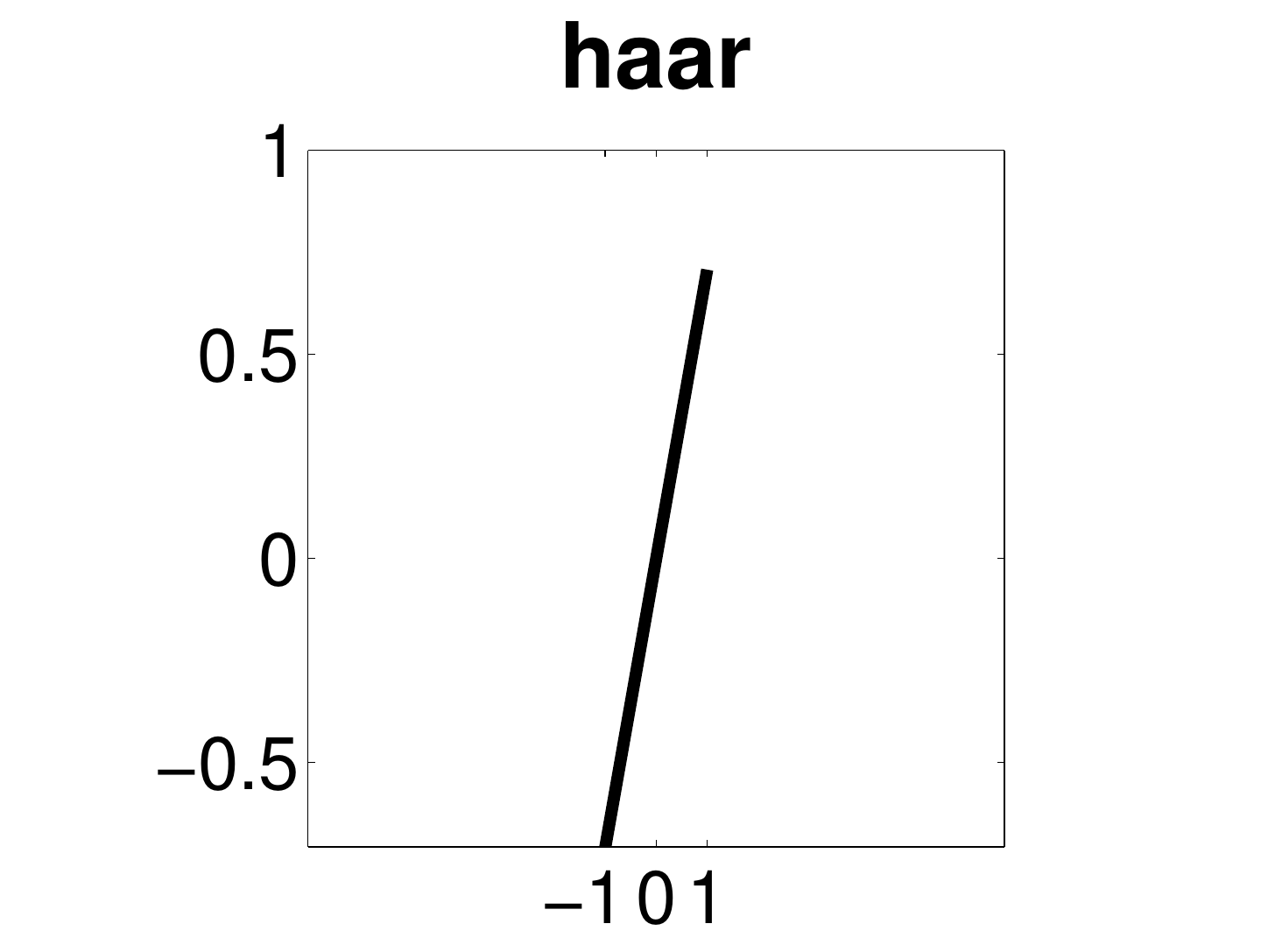}}
\subfloat{\includegraphics[trim = 60pt 10pt 60pt 5pt, clip,width=0.20\linewidth]{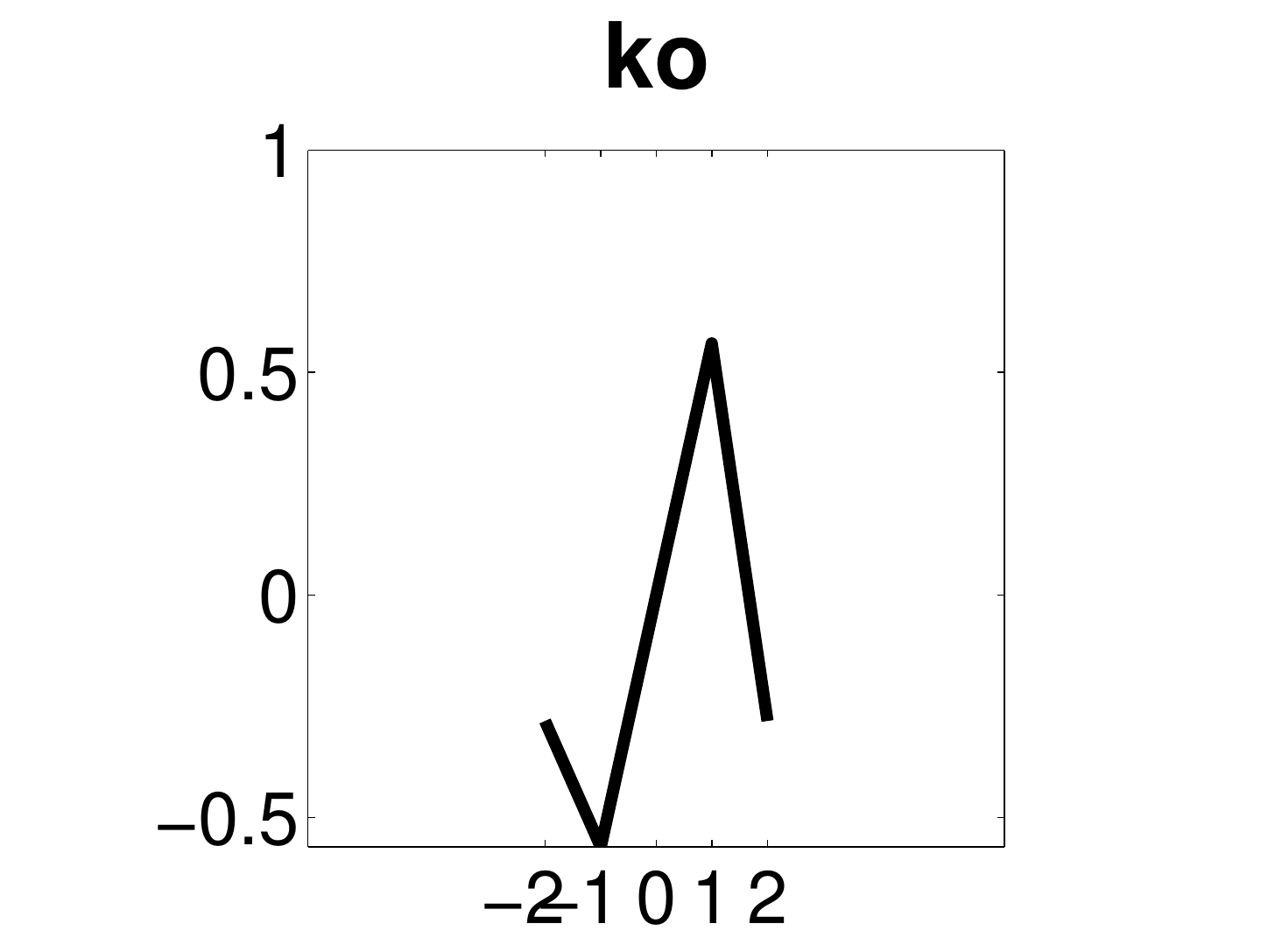}}
\subfloat{\includegraphics[trim = 60pt 10pt 60pt 5pt, clip,width=0.20\linewidth]{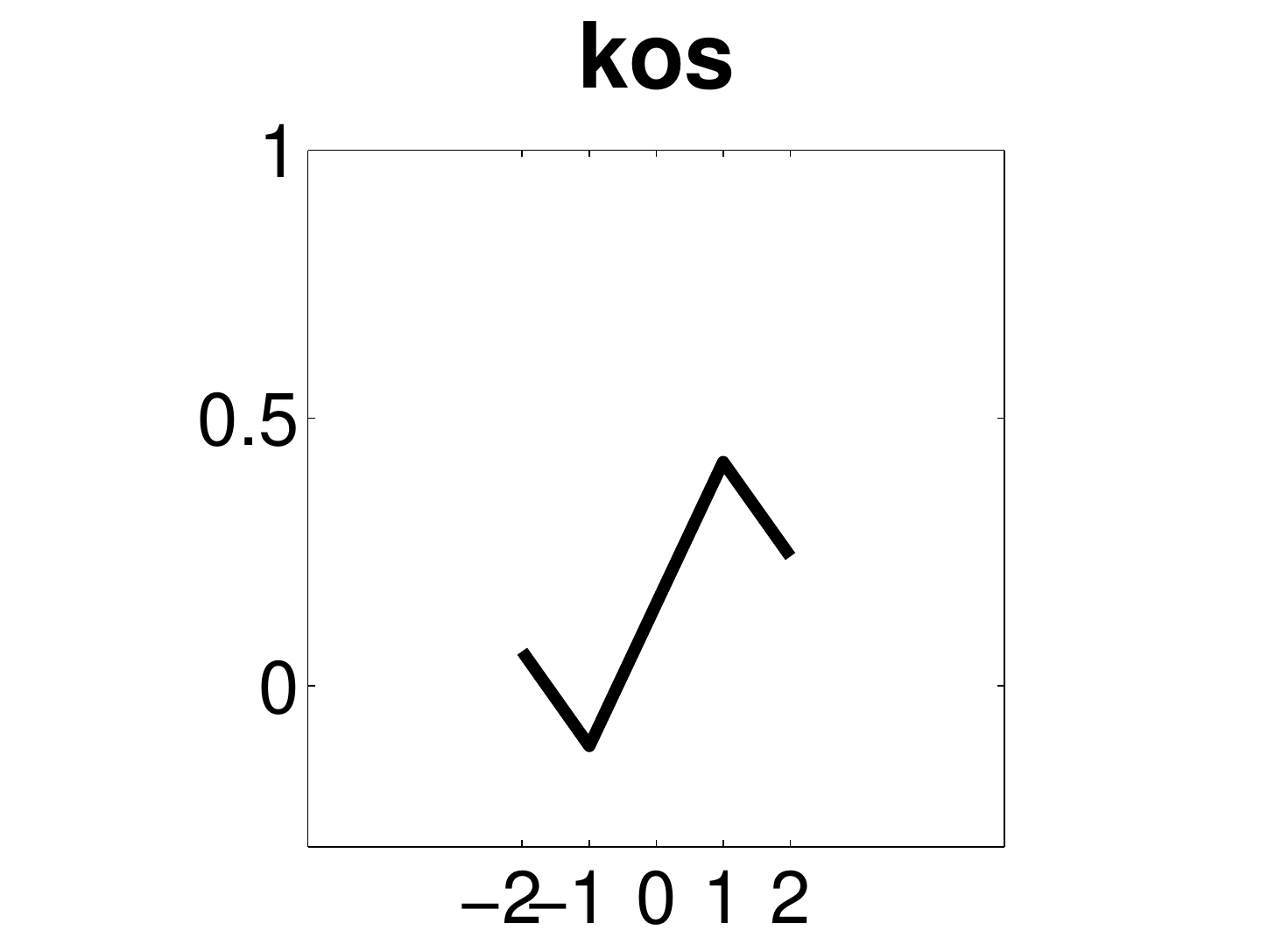}}\\
\subfloat{\includegraphics[trim = 60pt 10pt 60pt 5pt, clip,width=0.20\linewidth]{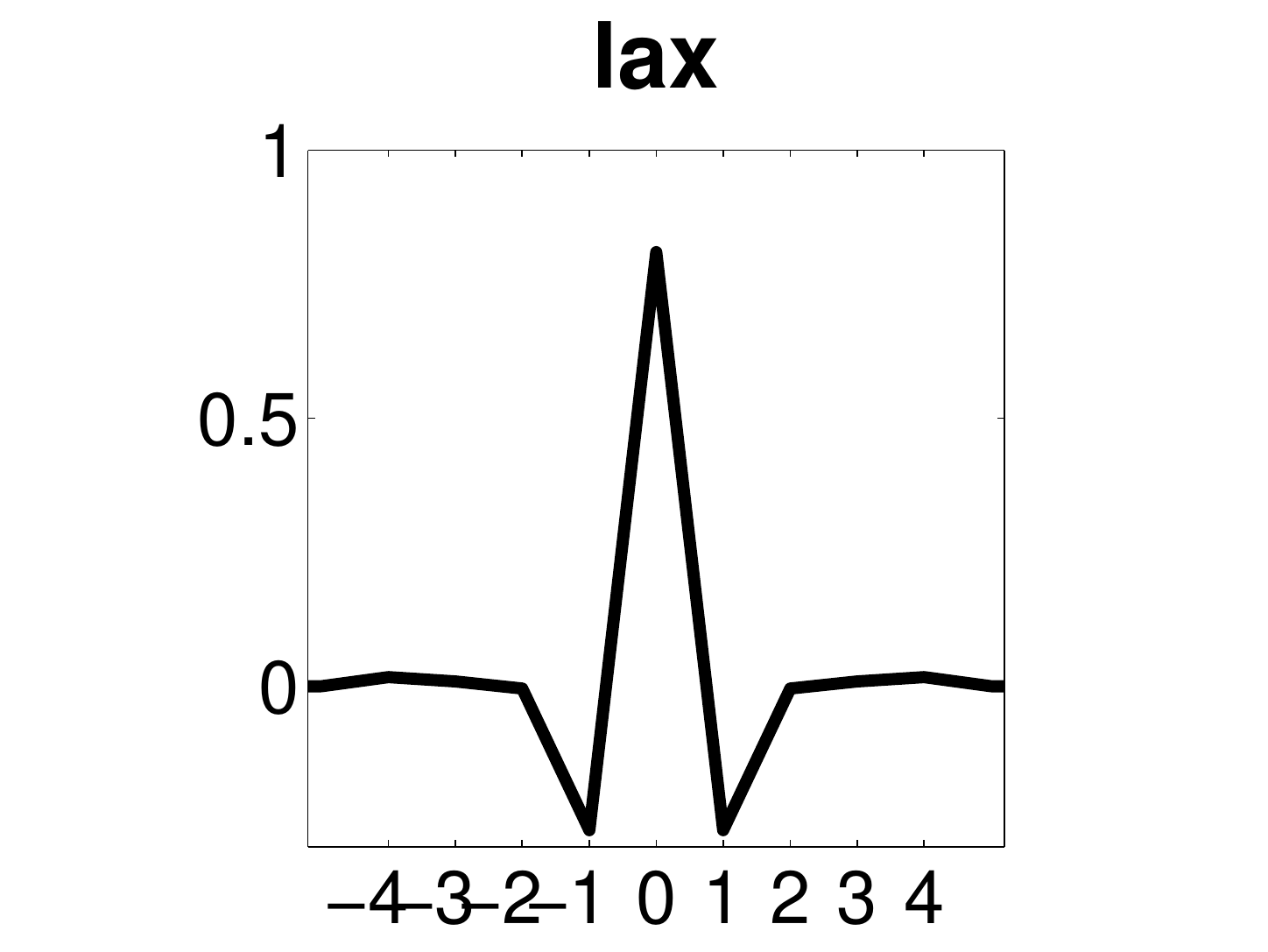}}
\subfloat{\includegraphics[trim = 60pt 10pt 60pt 5pt, clip,width=0.20\linewidth]{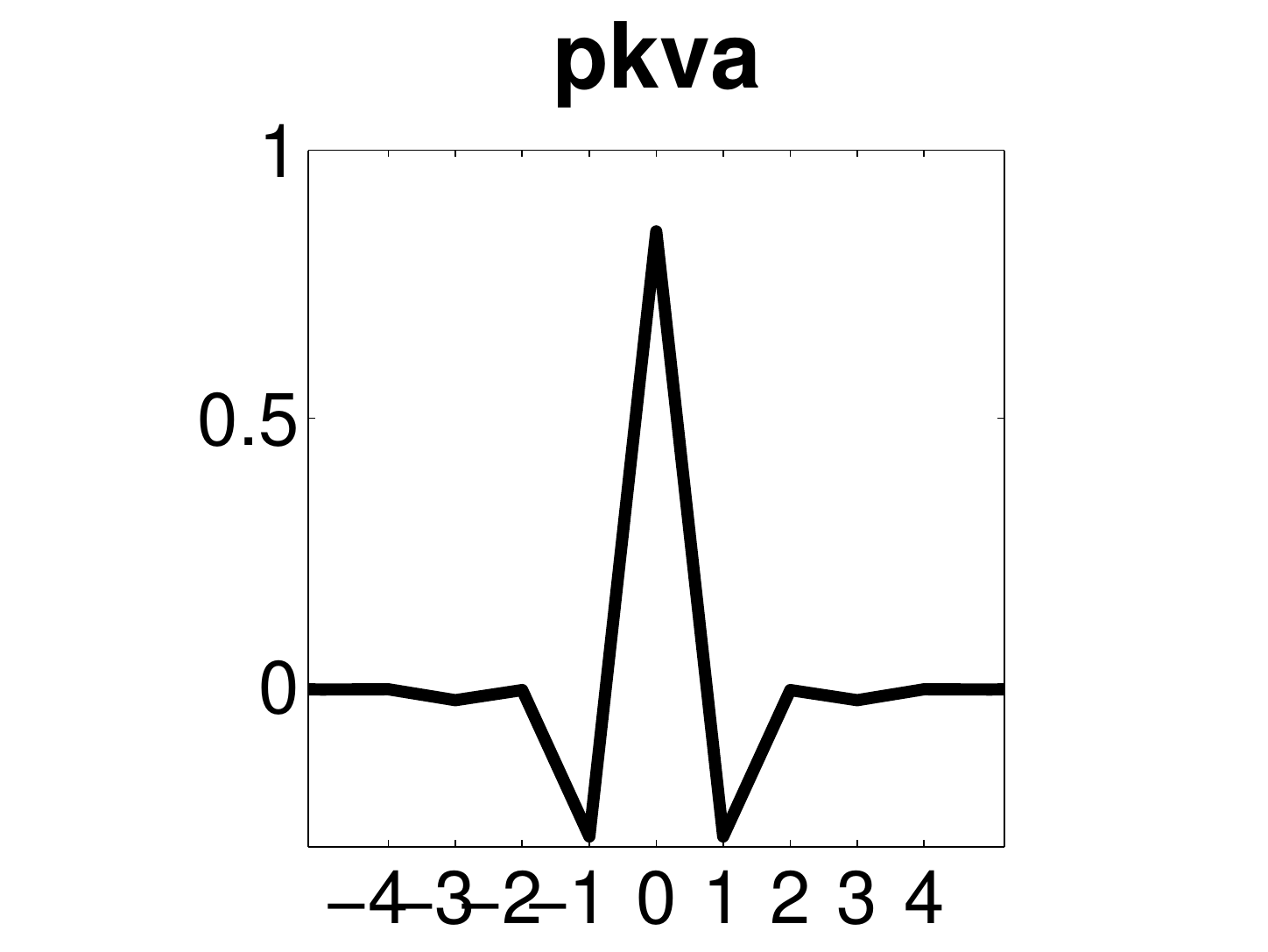}}
\subfloat{\includegraphics[trim = 60pt 10pt 60pt 5pt, clip,width=0.20\linewidth]{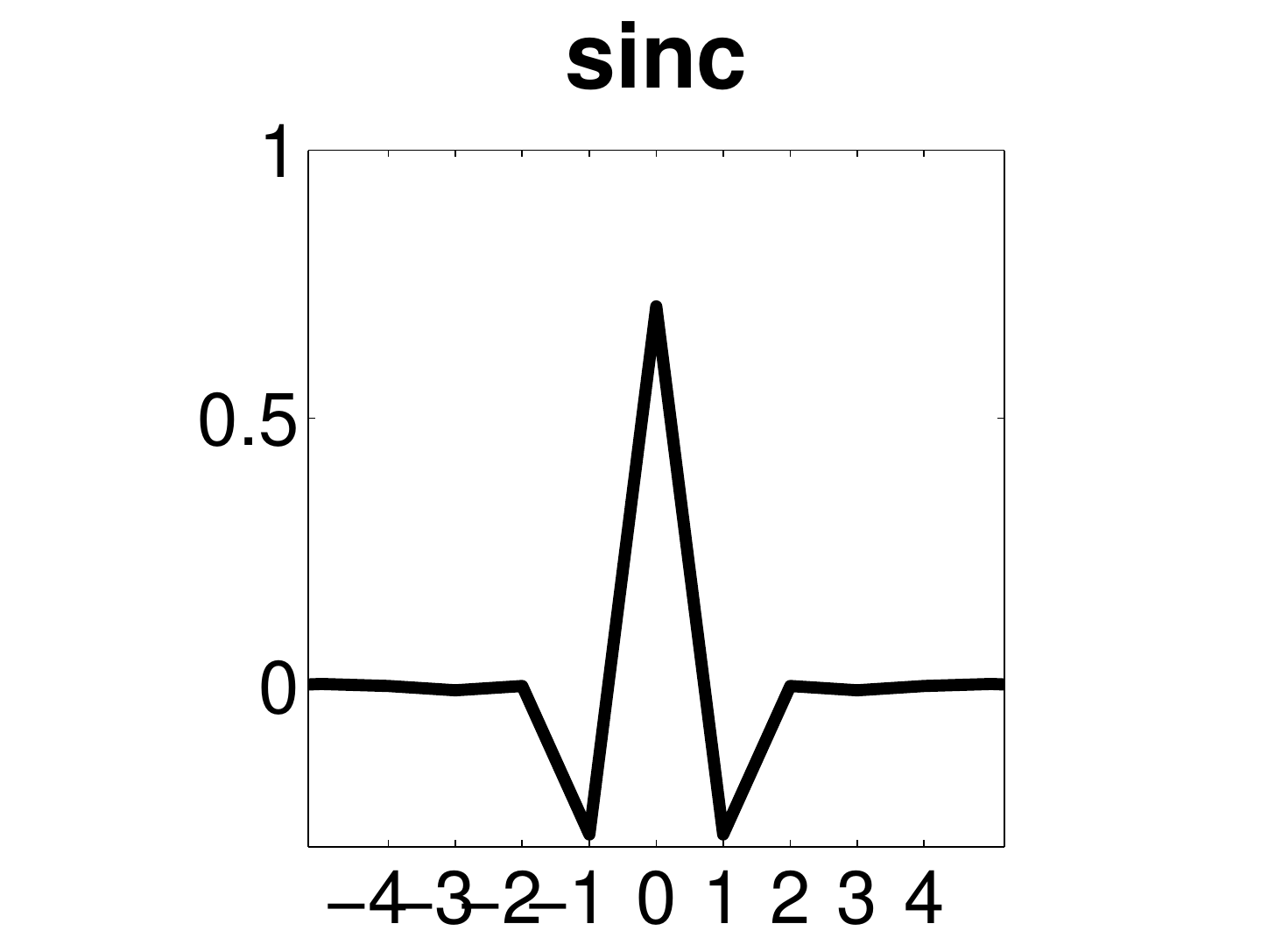}}
\subfloat{\includegraphics[trim = 60pt 10pt 60pt 5pt, clip,width=0.20\linewidth]{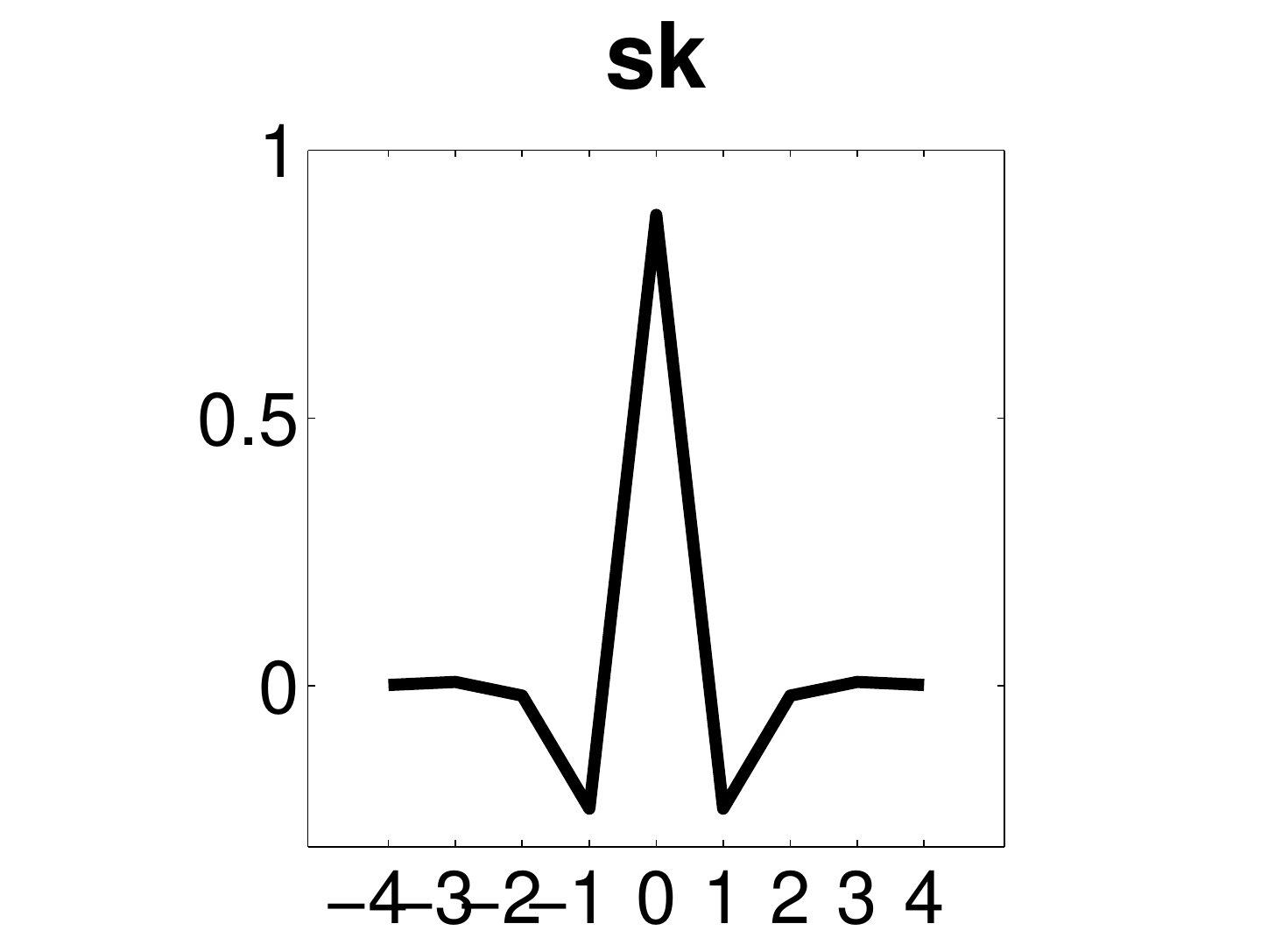}}
\subfloat{\includegraphics[trim = 60pt 10pt 60pt 5pt, clip,width=0.20\linewidth]{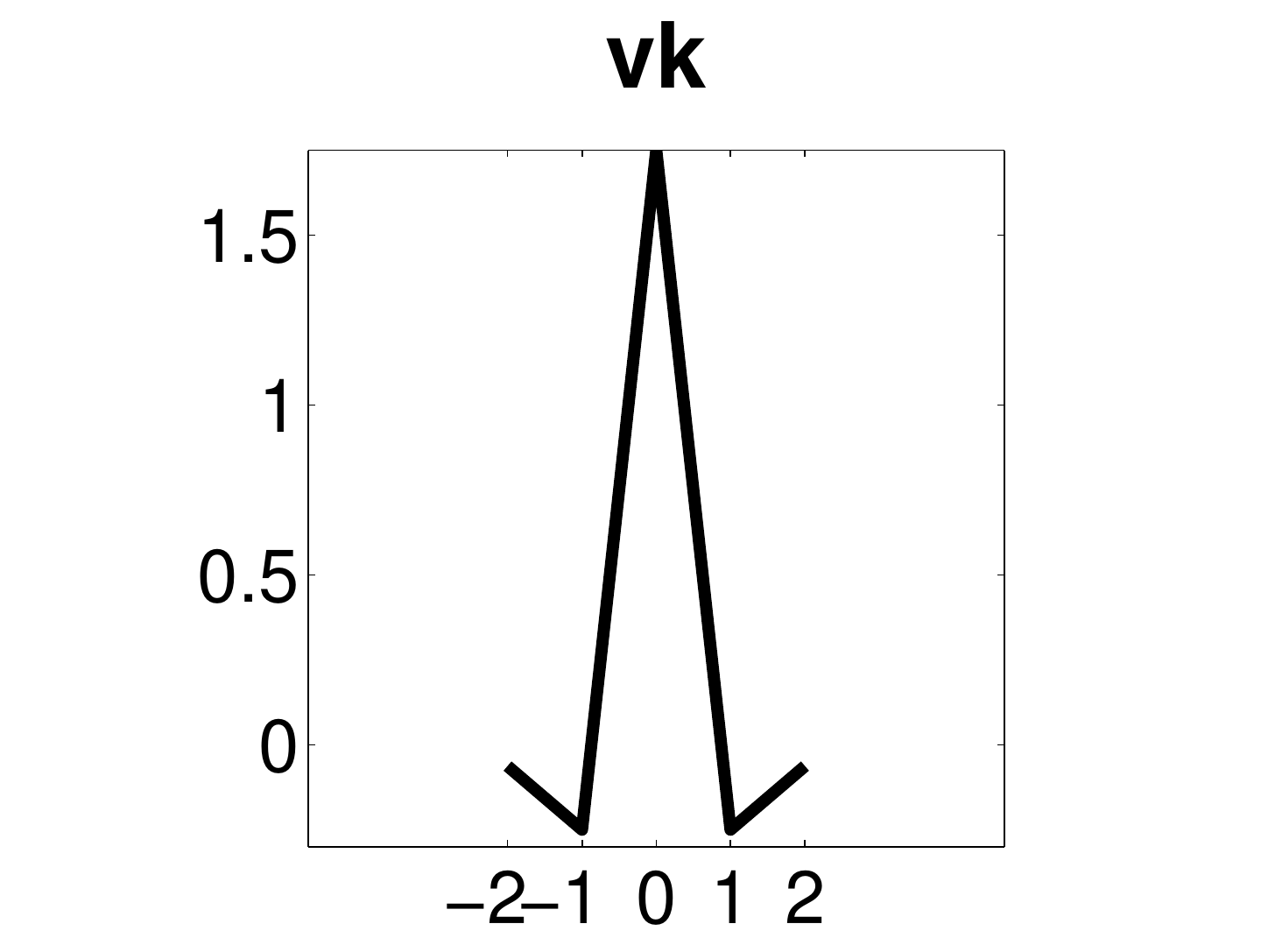}}
\caption{NSCT directional highpass filters. (a) cd: 7 and 9 McClellan transformed by Cohen and Daubechies \cite{cd}. (b) dvmlp: regular linear phase biorthogonal filter with 3 dvm \cite{dvmlp}. (c) haar: the `Haar' filter. (d) ko: orthogonal filter from Kovacevic (e) kos: smooth `ko' filter. (f) lax: $17\times17$ by Lu, Antoniou and Xu \cite{lax}. (g) pkva: ladder filters by Phong et al.~\cite{pkva}. (h) sinc: ideal filter. (i) sk: $9\times9$ by Shah and Kalker \cite{sk}. (j) vk: McClellan transform of filter from the VK book \cite{vk}.}
\label{fig:dfilters}
\end{figure}

\begin{figure}[t]
\centering
\includegraphics[trim = 20pt 100pt 38pt 80pt, clip,width=1\linewidth]{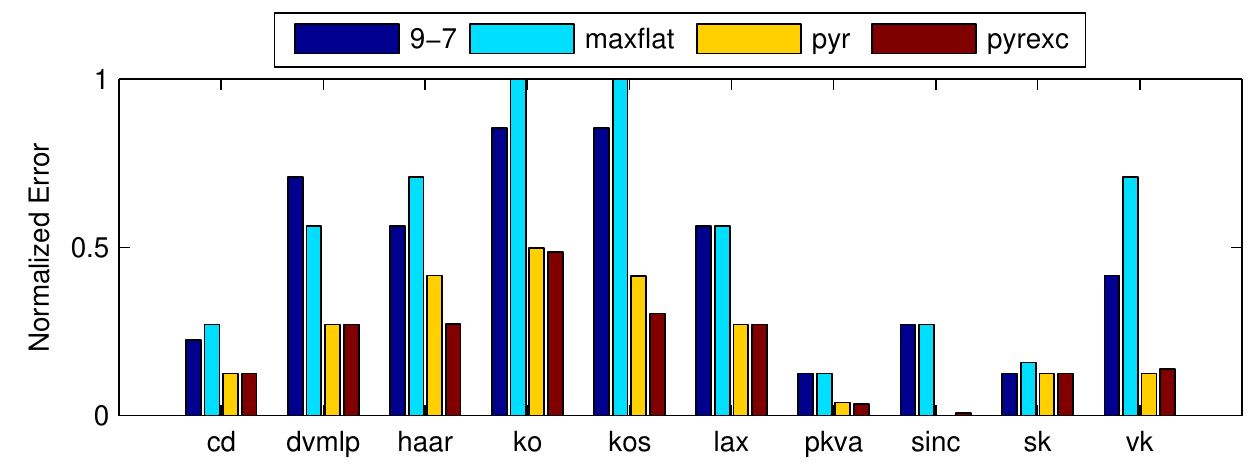}
\caption{Normalized EERs for various combinations of pyramidal-directional filter pairs. The \emph{9-7} and \emph{maxflat} pyramidal filters can be safely ruled out due to their relatively poor performance overall. The \emph{pyrexc} filter performs equal or better than the \emph{pyr} filter for atleast 8 out of the 10 combinations with the directional filters. Thus, the \emph{pyrexc} filter is an obvious choice for the pyramidal filter stage. For the directional filter stage, the \emph{sinc} filter clearly outperforms all other directional filters, particularly in combination with the \emph{pyrexc} pyramidal filter. From this analysis, the \emph{sinc-pyrexc} filter pair is used in the Contour Code representation.}
\label{fig:Filt_PolyU_MS}
\end{figure}

We tested all possible pairs of $4$ pyramidal and $10$ directional filters available in the Nonsubsampled Contourlet Toolbox~\cite{nsct}. These filters exhibit characteristic response to line-like features. The pyramidal filters used at the first stage are shown in Fig.~\ref{fig:pfilters} and the directional filters used at the second stage are shown in Fig.~\ref{fig:dfilters}. The EER is normalized by dividing with the maximum EER in all filter combinations. As shown in Fig.~\ref{fig:Filt_PolyU_MS}, the \emph{pyrexc} filter consistently achieves lower EER compared to the other three pyramidal decomposition filters for most combinations of the directional filters. Moreover, the \emph{sinc} filter exhibits the lowest EER which shows its best directional feature capturing characteristics over a broad spectral range. Based on these results, we select the \emph{pyrexc-sinc} filter combination for Contour Code representation in all experiments. Close inspection of the pyramidal filters shows that {\em pyr} and {\em pyrexc} are quite similar to each other. Similarly, {\em sinc}, {\em pkva} and {\em sk} directional filters have approximately the same shape. Therefore, it is not surprising to see that all combinations of these two pyramidal filters with the three directional filters give significantly better performance compared to the remaining combinations.

\subsection{Verification Experiments}
\label{sec:roc}

Verification experiments are performed on both PolyU and CASIA databases. In both cases, half of the samples per identity were acquired in one session and the other half were acquired in a different session. We follow the protocol of~\cite{zhang2012comparative}, where session based experiments are structured to observe the recognition performance. We design five experiments to test the proposed technique in real world scenarios. The experiments proceed by matching

\vspace{1mm}
\noindent
\emph{Exp.1:} individual bands of palm irrespective of the session (all vs.~all).\\
\emph{Exp.2:} multispectral palmprints acquired in the $1^{\textrm{st}}$ session.\\
\emph{Exp.3:} multispectral palmprints acquired in the $2^{\textrm{nd}}$ session.\\
\emph{Exp.4:} multispectral palmprints of the $1^{\textrm{st}}$ session to the $2^{\textrm{nd}}$ session.\\
\emph{Exp.5:} multispectral palmprints irrespective of the session (all vs.~all).
\vspace{1mm}

\begin{table}[t]
\caption{The number of genuine and imposter matches for each experiment.}
\label{tab:expStats}
\begin{center}
\begin{tabular}{|l|l|ccc|} \hline
\multirow{2}{*}{}               & \multirow{2}{*}{} & \multicolumn{3}{c|}{Experiments}\\ \hline
                        &       &   Exp.2 \& 3                      &   Exp.~4                      &   Exp.1 \& 5              \\ \hline
\multirow{2}{*}{PolyU}  &   Gen.&   \multicolumn{1}{r}{7,500}       &   \multicolumn{1}{r}{18,000}  &   \multicolumn{1}{r|}{33,000}     \\
                        &   Imp.&   \multicolumn{1}{r}{4,491,000}   &   \multicolumn{1}{r}{8,982,000}&\multicolumn{1}{r|}{17,964,000}   \\ \hline
\multirow{2}{*}{CASIA}  &   Gen.&   \multicolumn{1}{r}{600}         &   \multicolumn{1}{r}{1,800}   &   \multicolumn{1}{r|}{3,000}      \\
                        &   Imp.&   \multicolumn{1}{r}{179,100}     &   \multicolumn{1}{r}{358,200} &   \multicolumn{1}{r|}{716,400}    \\ \hline
\end{tabular}
\end{center}
\end{table}

In all cases, we report the ROC curves, which depict False Rejection Rate (FRR) versus the False Acceptance Rate (FAR). We also summarize the Equal Error Rate (EER), and the Genuine Acceptance Rate (GAR) at 0.1\% FAR and compare performance of the proposed Contour Code with three state-of-the-art techniques namely, the CompCode, OrdCode and DoGCode. Note that we used our implementation of these methods as their code is not publicly available. The source code of these algorithms including the techniques proposed in this work will soon be made publicly available for research. The number of genuine and imposter matches for each experiment are given in Table~\ref{tab:expStats}. Unless otherwise stated, we use a \emph{4-connected} blur neighborhood for gallery hash table encoding and the matching is performed in ATM mode followed by score-level fusion of bands. A list of frequently used abbreviations is given in Table~\ref{tab:abbrev}.

\begin{table}[h]
\caption{List of abbreviations.}
\footnotesize
\label{tab:abbrev}
\begin{center}
\begin{tabular}{|l|l|} \hline
Abbreviation                        & Meaning \\ \hline
ATM                                 & Asynchronous Translation Matching     \\
STM                                 & Synchronous Translation Matching      \\
ContCode                            & Proposed Contour Code representation           \\
CompCode~\cite{kong2004competitive} & Competitive Code representation       \\
OrdCode~\cite{sun2005ordinal}       & Ordinal Code representation           \\
DoGCode~\cite{wu2006palmprint}      & Derivative of Gaussian Code representation \\ \hline
\end{tabular}
\end{center}
\end{table}

\subsubsection{Experiment 1}

\begin{figure}[t]
\centering
\subfloat{\label{fig:roc_PolyU_MS_Exp1}\includegraphics[trim = 0pt 0pt 10pt 0pt, clip, width=0.5\linewidth]{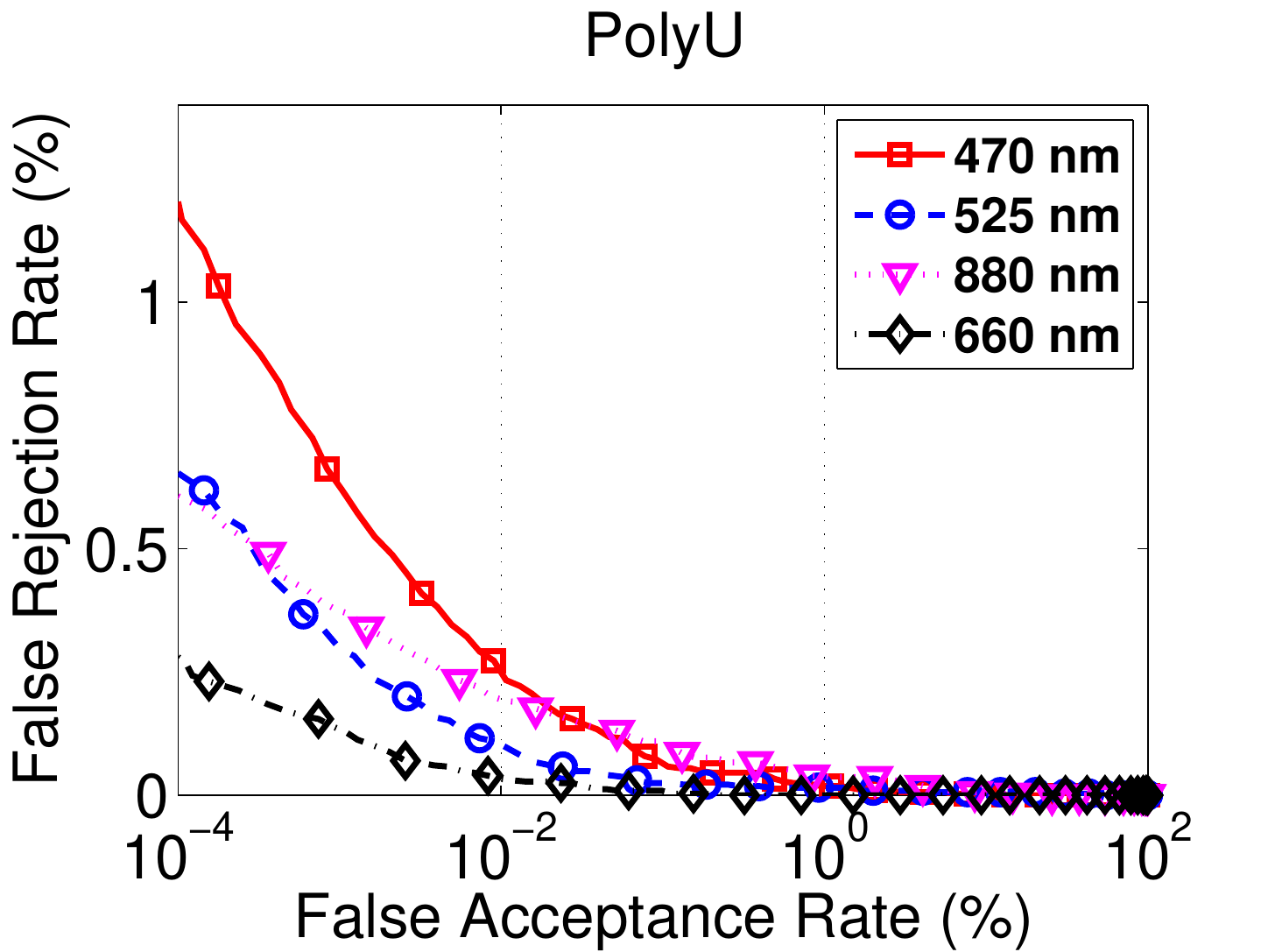}}
\subfloat{\label{fig:roc_CASIA_MS_Exp1}\includegraphics[trim = 0pt 0pt 10pt 0pt, clip, width=0.5\linewidth]{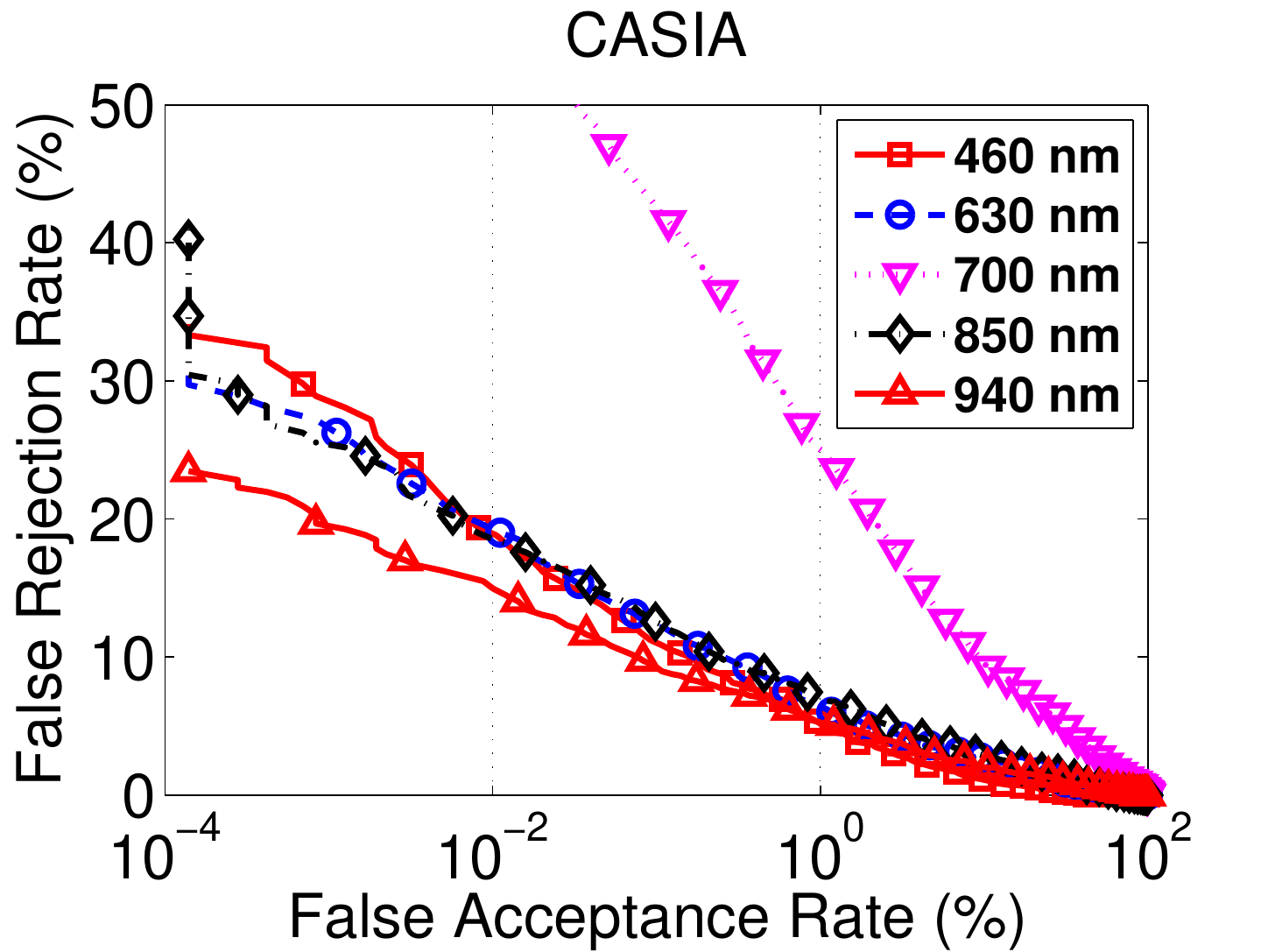}}
\caption{Exp.1: ROC curves of {ContCode-ATM} on individual bands}
\label{fig:roc_Exp1}
\end{figure}

\emph{Exp.1} compares the relative discriminant capability of individual bands. We compare the performance of individual bands of the PolyU and CASIA database using {ContCode-ATM}. Fig.~\ref{fig:roc_Exp1} shows the ROC curves of the individual bands and Table~\ref{tab:res_Exp1} lists their EERs. For the PolyU database, the 660nm band gives the best performance indicating the presence of more discriminatory features. A logical explanation could be that the 660nm wavelength partially captures both the line and vein features making this band relatively more discriminative. For CASIA database, the most discriminant information is present in the 460nm, 630nm and 940nm bands which are close competitors.

\begin{table}[h]
\caption{Individual band performance of {ContCode-ATM}}
\label{tab:res_Exp1}
\centering
\scriptsize
\begin{tabular}[t]{|l|cc|}
\multicolumn{3}{c}{PolyU} \\ \noalign{\smallskip}\hline
Band    & GAR(\%)   &  EER(\%)  \\ \hline
470 nm  & 99.94     &   0.0784  \\
525 nm  & 99.98     &   0.0420  \\
660 nm  & 99.99     &   0.0242  \\
880 nm  & 99.90     &   0.1030  \\ \hline
\end{tabular}
\begin{tabular}[t]{|l|cc|}
\multicolumn{3}{c}{CASIA} \\ \noalign{\smallskip}\hline
Band    & GAR(\%)   &  EER(\%)  \\ \hline
460 nm  & 88.95     &   2.9246  \\
630 nm  & 87.79     &   3.9065  \\
700 nm  & 57.35     &   9.7318  \\
850 nm  & 87.45     &   4.1398  \\
940 nm  & 90.73     &   3.4769  \\ \hline
\end{tabular}
\end{table}

\begin{figure}[t]
\centering
\subfloat{\label{fig:roc_PolyU_MS_Exp2}\includegraphics[trim = 0pt 0pt 10pt 0pt, clip, width=0.5\linewidth]{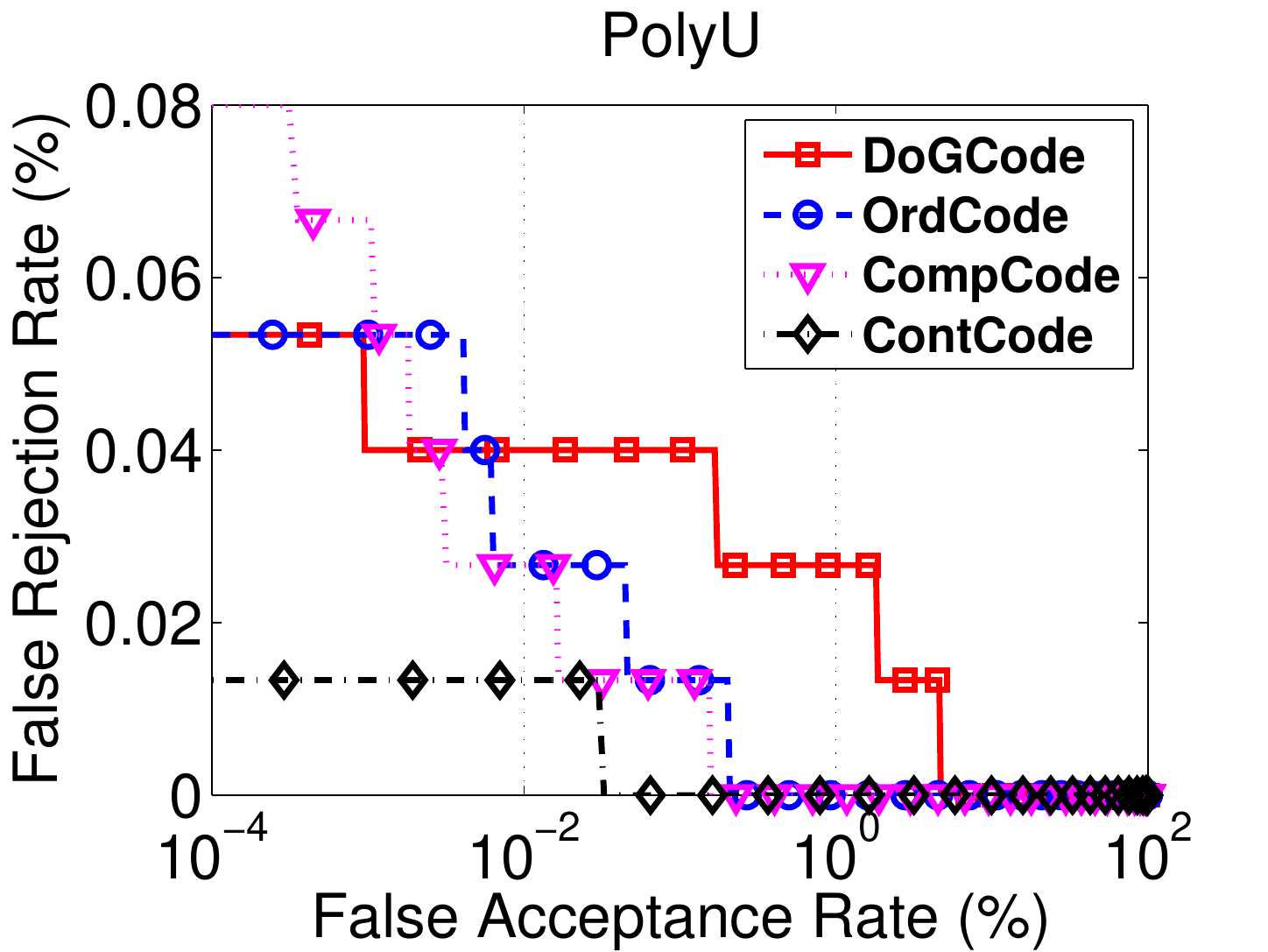}}
\subfloat{\label{fig:roc_CASIA_MS_Exp2}\includegraphics[trim = 0pt 0pt 10pt 0pt, clip, width=0.5\linewidth]{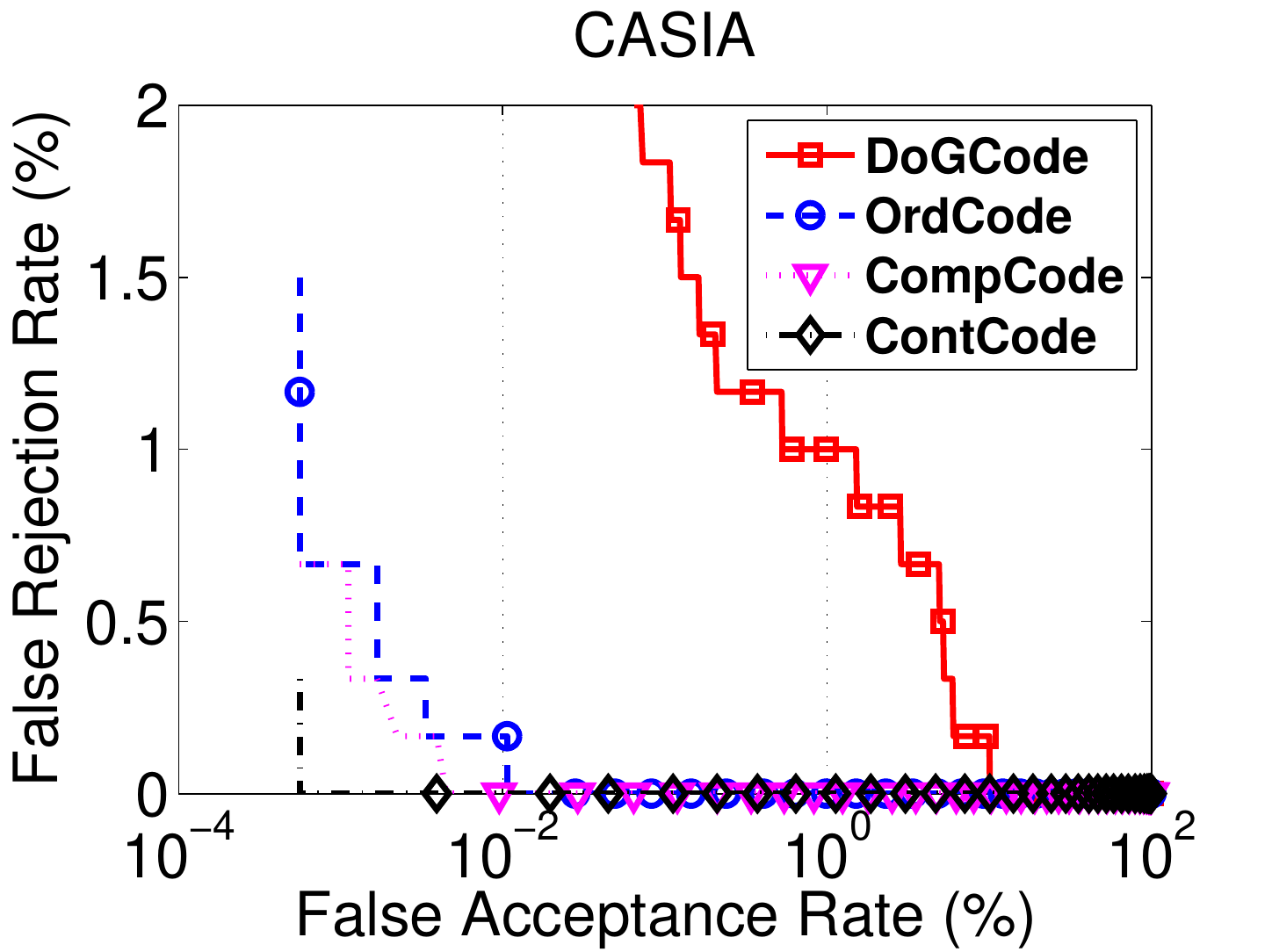}}\\
\caption{Exp.2: Matching palmprints of $1^{st}$ session. The CompCode and ContCode-ATM perform better on both databases and the latter performs the best overall.}
\label{fig:roc_Exp2}
\end{figure}

\subsubsection{Experiment 2}
\emph{Exp.2} analyzes the variability in the palmprint data acquired in the $1^{st}$ session. Fig.~\ref{fig:roc_Exp2} compares the ROC curves of the {ContCode-ATM} with three other techniques on the PolyU and CASIA databases.It is observable that, the CompCode and the OrdCode show intermediate performance close to {ContCode-ATM}. The DoGCode exhibits a drastic degradation of accuracy implying its inability to sufficiently cope with the variations of CASIA data.

\begin{table*}[t]
\caption{Summary of verification results for \emph{Exp.2} to \emph{Exp.5}}
\scriptsize
\label{tab:res_Exp}
\centering
\begin{tabular}{|l|c|c|c|c|c||c|c|c|c|} \hline
\multicolumn{2}{|c|}{} & \multicolumn{4}{c||}{PolyU} & \multicolumn{4}{c|}{CASIA} \\ \cline{3-10}
\multicolumn{2}{|c|}{} & {DoGCode} & {OrdCode}  & {CompCode}  & {ContCode-ATM} & DoGCode & \multicolumn{1}{c}{OrdCode}  & \multicolumn{1}{c}{CompCode}  & \multicolumn{1}{c|}{{ContCode-ATM}}  \\ \hline
\multirow{2}{*}{Exp.2}  & EER(\%)  & 0.0400 & 0.0267      &   0.0165  &   0.0133   & 1.000   & 0.1667     &   0.0140  &   0       \\
                        & GAR(\%)  & 99.96  & 99.99       &   99.99   &   100.00   & 98.00   & 99.67      &   100.00  &   100.00  \\ \hline
\multirow{2}{*}{Exp.3}  & EER(\%)  & 0.0133 & 0           &   0.0098  &   0        & 0.6667  & 0.1667     &   0.1667  &   0.0011  \\
                        & GAR(\%)  & 99.99  & 100.00      &   100.00  &   100.00   & 98.50   & 99.83      &   99.83   &   100.00  \\ \hline
\multirow{2}{*}{Exp.4}  & EER(\%)  & 0.0528 & 0.0247      &   0.0333  &   0.0029   & 3.8669  & 1.2778     &   0.6667  &   0.2778  \\
                        & GAR(\%)  & 99.96  & 99.98       &   99.99   &   100.00   & 87.70   & 97.39      &   97.72   &   99.61   \\ \hline
\multirow{2}{*}{Exp.5}  & EER(\%)  & 0.0455 & 0.0212      &   0.0263  &   0.0030   & 2.8873  & 0.8667     &   0.4993  &   0.2000  \\
                        & GAR(\%)  & 99.97  & 99.99       &   99.99   &   100.00   & 92.01   & 98.37      &   98.60   &   99.76   \\ \hline
\end{tabular}
\end{table*}

\begin{figure}[t]
\centering
\subfloat{\label{fig:roc_PolyU_MS_Exp3}\includegraphics[trim = 0pt 0pt 10pt 0pt, clip, width=0.5\linewidth]{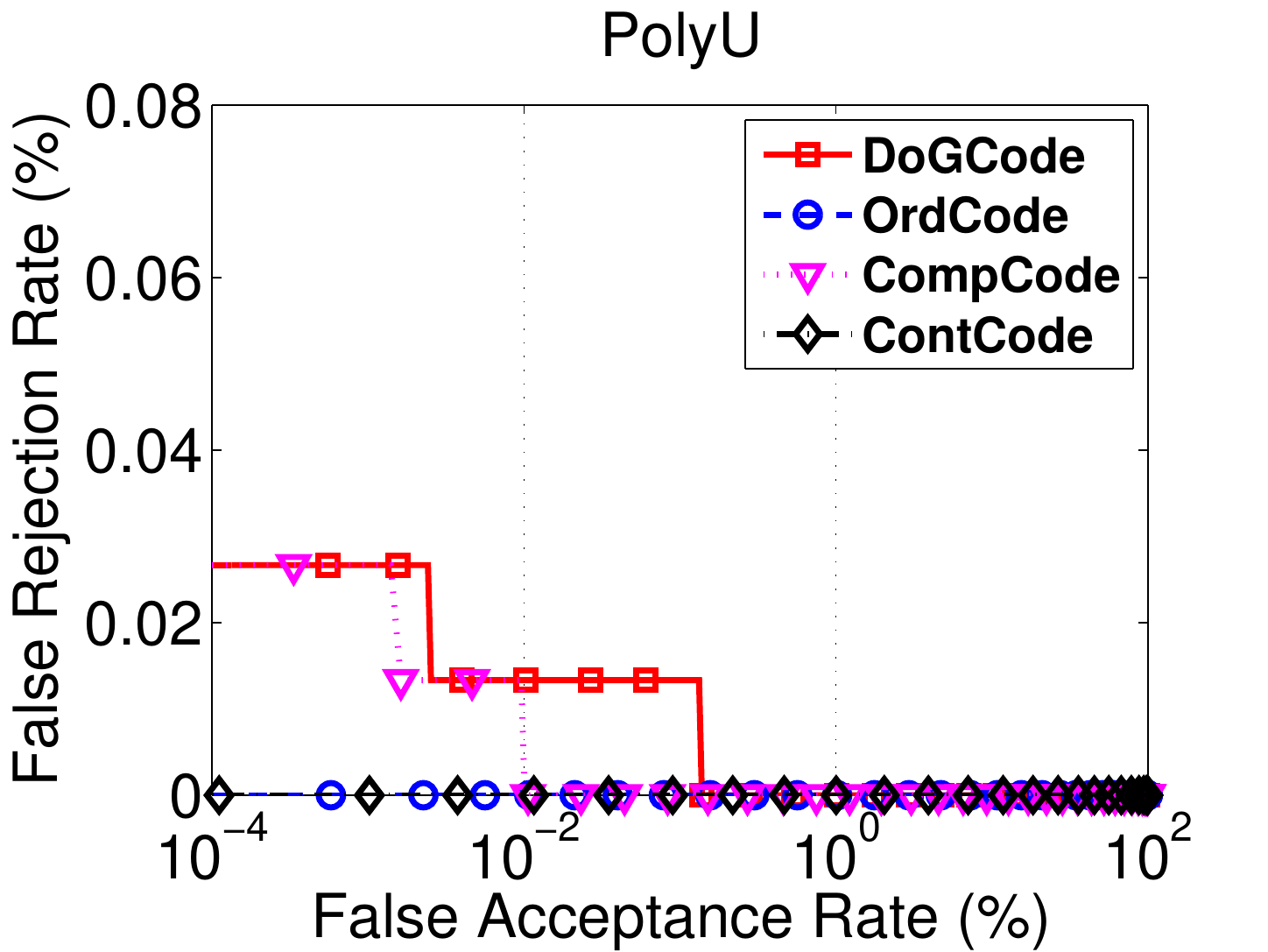}}
\subfloat{\label{fig:roc_CASIA_MS_Exp3}\includegraphics[trim = 0pt 0pt 10pt 0pt, clip, width=0.5\linewidth]{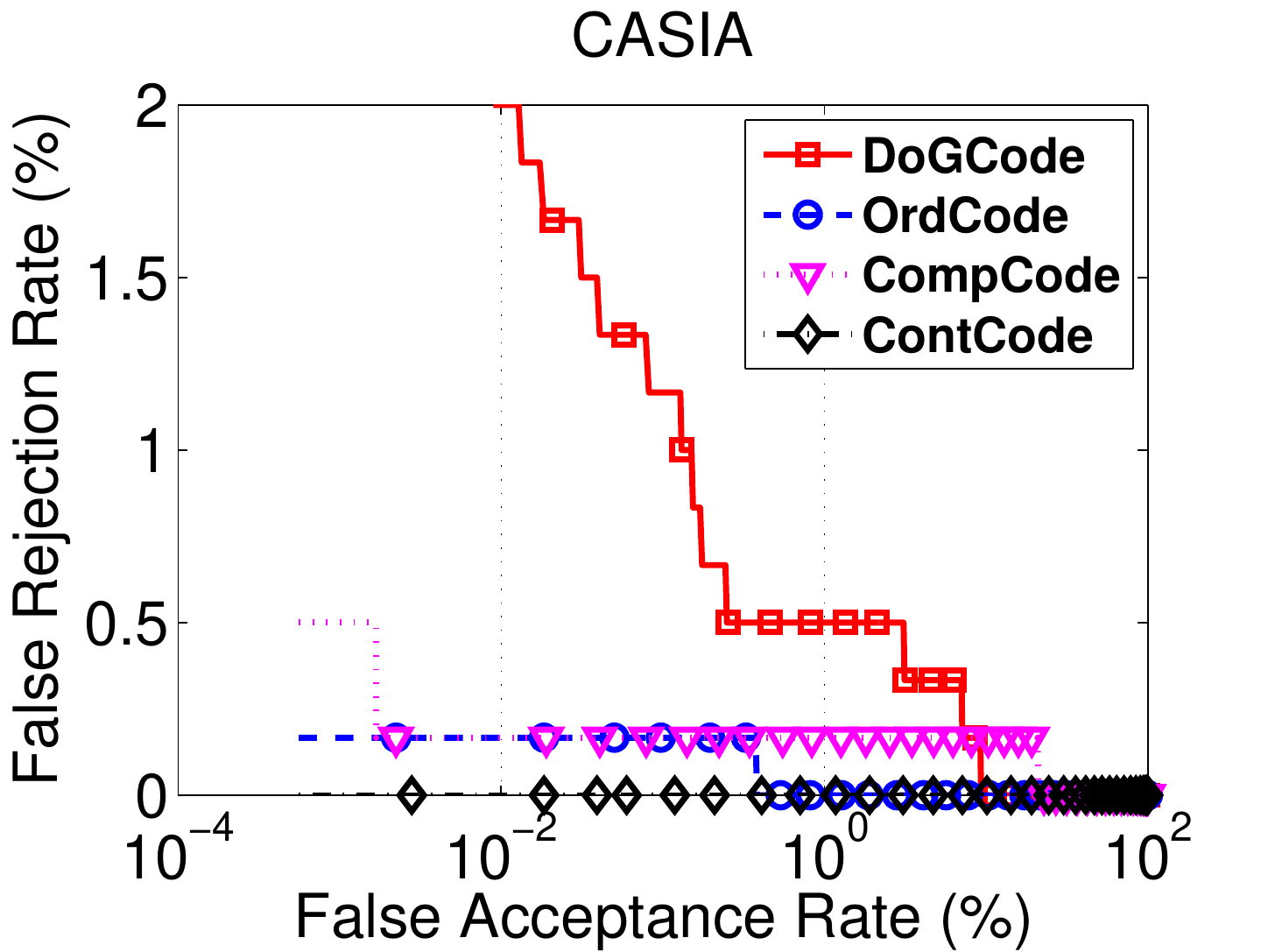}}\\
\caption{Exp.3: Matching palmprints of $2^{nd}$ session. The small improvement in verification performance on the images of $2^{nd}$ session on both databases can be attributed to the better quality of images. A similar trend is observed in verification performance of all techniques as in \emph{Exp.2}}
\label{fig:roc_Exp3}
\end{figure}

\subsubsection{Experiment 3}
\emph{Exp.3} analyzes the variability in the palmprint data acquired in the $2^{nd}$ session. This allows for a comparison with the results of \emph{Exp.2} to analyze the intra-session variability. Therefore, only the palmprints acquired in the $2^{nd}$ session are matched. Fig.~\ref{fig:roc_Exp3} compares the ROC curves of the {ContCode-ATM} with three existing techniques.

\subsubsection{Experiment 4}
\begin{figure}[t]
\centering
\subfloat{\label{fig:roc_PolyU_MS_Exp4}\includegraphics[trim = 0pt 0pt 10pt 0pt, clip, width=0.5\linewidth]{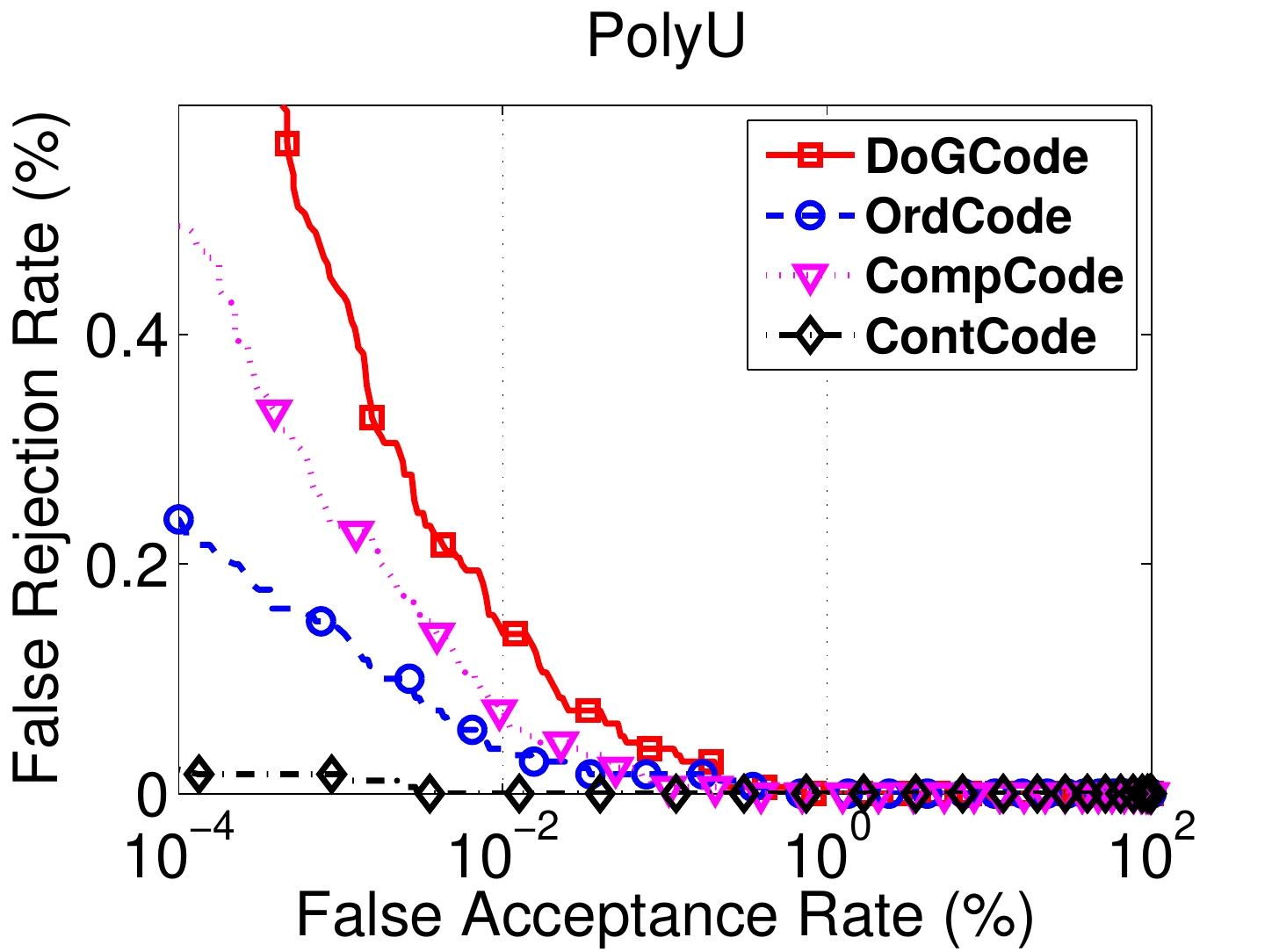}}
\subfloat{\label{fig:roc_CASIA_MS_Exp4}\includegraphics[trim = 0pt 0pt 10pt 0pt, clip, width=0.5\linewidth]{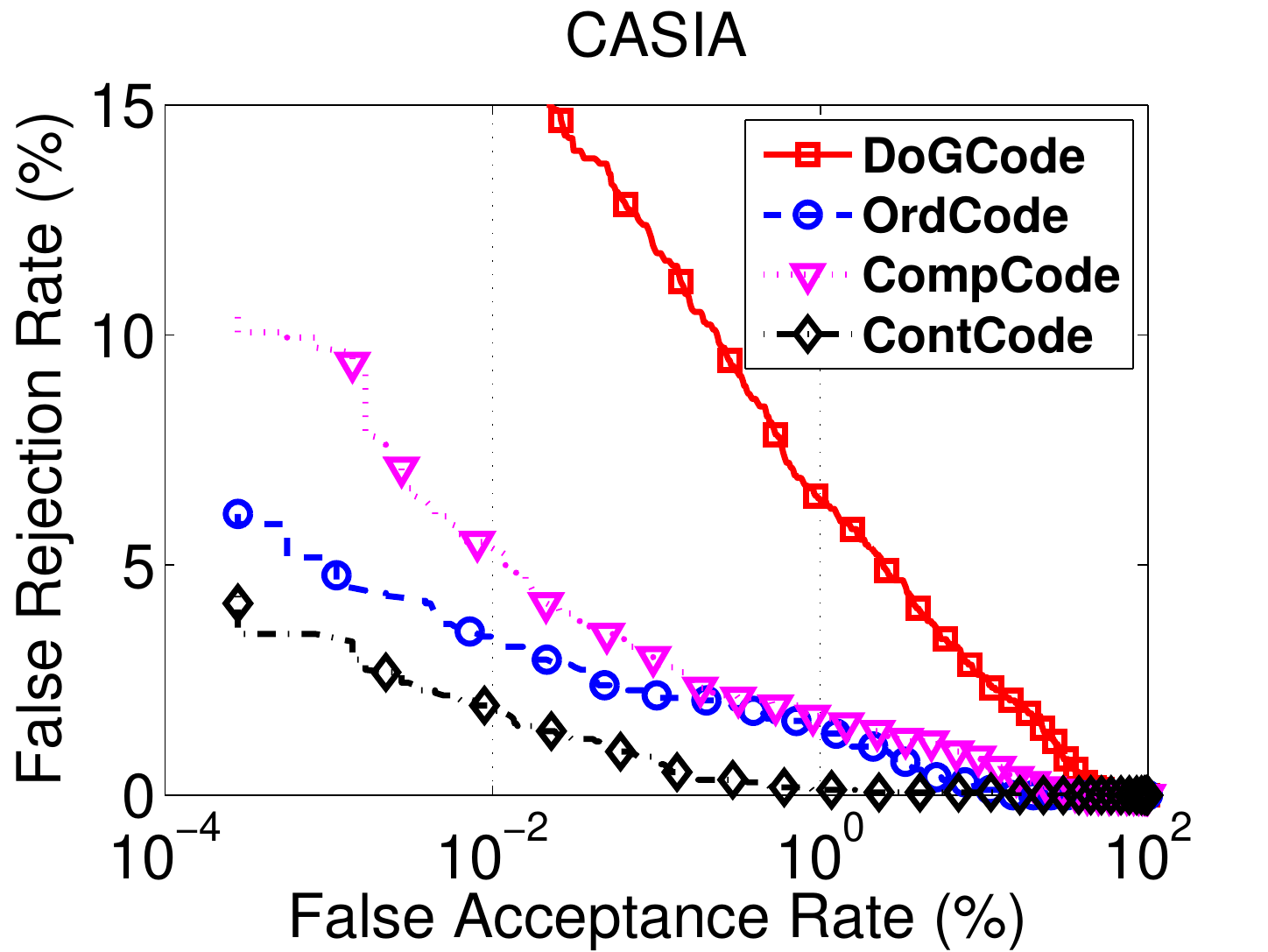}}\\
\caption{Exp.4: Matching palmprints of the $1^{st}$ session to the $2^{nd}$ session. The verification performance is low relative to \emph{Exp.2} and \emph{Exp.3}. However, the performance degradation of the proposed {ContCode-ATM} is much less than the other techniques on both databases, indicating its robustness to image variability.}
\label{fig:roc_Exp4}
\end{figure}

\emph{Exp.4} is designed to mimic a real life verification scenario where variation in image quality or subject behavior over time exists. This experiment actually analyzes the inter-session variability of multispectral palmprints. Therefore, all images from the $1^{st}$ session are matched to all images of the $2^{nd}$ session. Fig.~\ref{fig:roc_Exp4} compares the ROC curves of the {ContCode-ATM} with three other techniques.
Note that the performance of other techniques is relatively lower for this experiment compared to \emph{Exp.2} and \emph{Exp.3} because this is a difficult scenario due to the intrinsic variability in image acquisition protocol and the human behavior over time. However, the drop in performance of {ContCode-ATM} is the minimum. Therefore, it is fair to deduce that {ContCode-ATM} is relatively robust to the image variability over time.

\begin{figure}[h]
\begin{center}
\subfloat{\label{fig:roc_PolyU_MS_Exp5}\includegraphics[trim = 0pt 0pt 10pt 0pt, clip, width=0.5\linewidth]{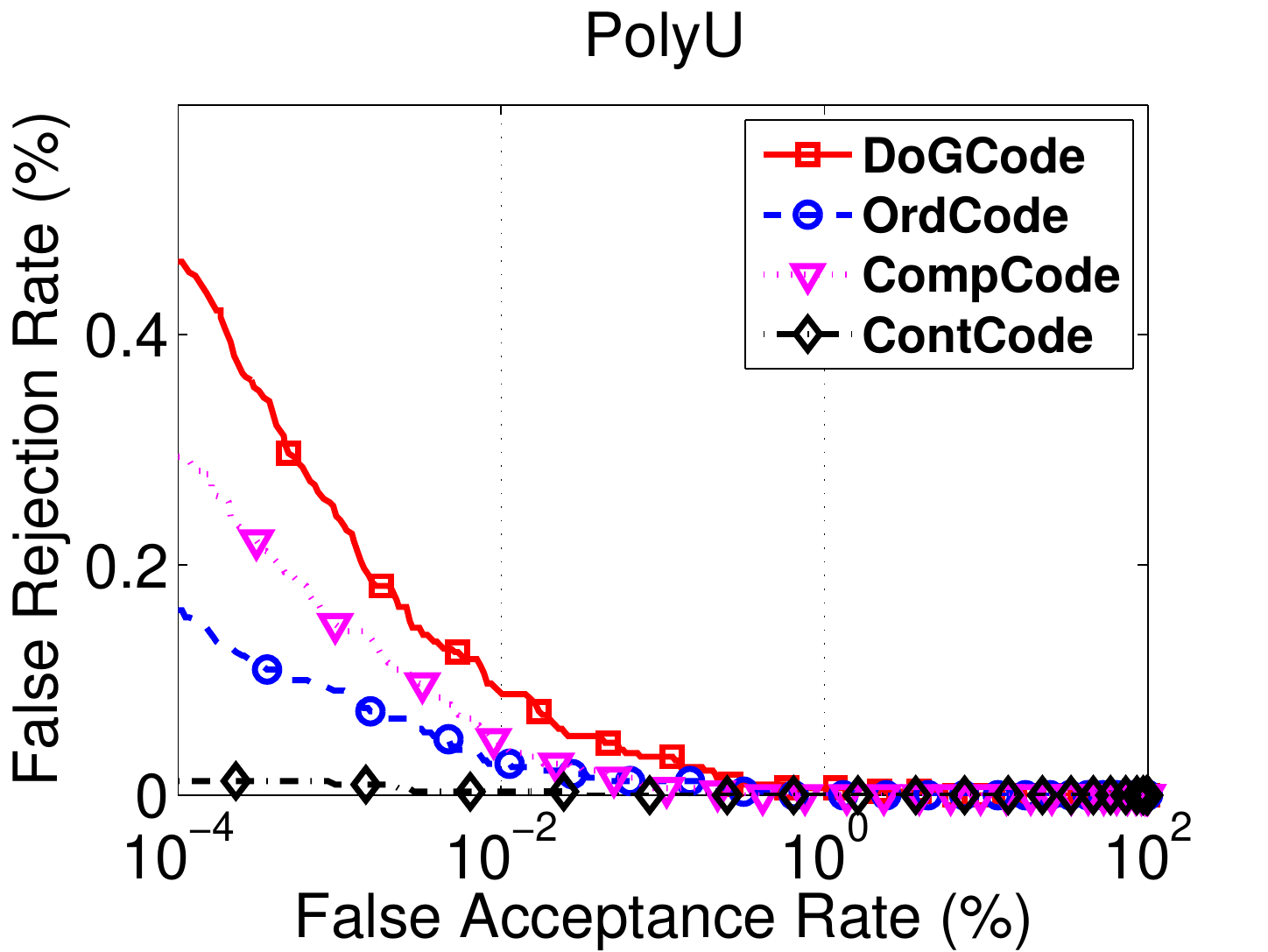}}
\subfloat{\label{fig:roc_CASIA_MS_Exp5}\includegraphics[trim = 0pt 0pt 10pt 0pt, clip, width=0.5\linewidth]{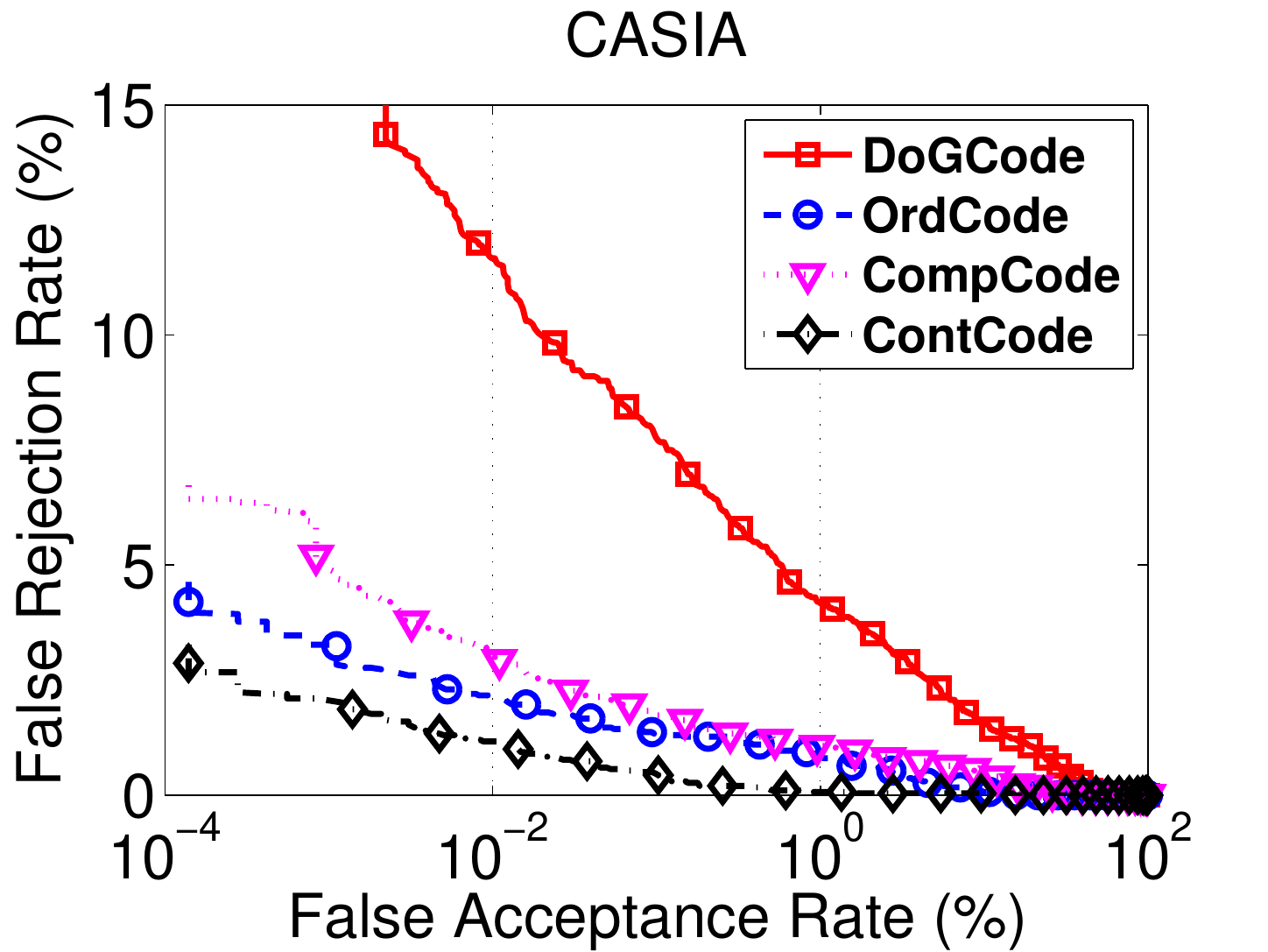}}\\
\end{center}
\caption{Exp.5: Matching palmprints irrespective of the acquisition session.}
\label{fig:roc_Exp5}
\end{figure}

\subsubsection{Experiment 5}
\emph{Exp.5} evaluates the overall verification performance by combining images from both sessions allowing a direct comparison to existing techniques whose results are reported on the same databases. All images in the database are matched to all other images, irrespective of the acquisition session which is commonly termed as an ``all versus all'' experiment in the literature. Fig.~\ref{fig:roc_Exp5} compares the ROC curves of the {ContCode-ATM} with three other techniques in the all versus all scenario. Similar to the previous experiments, the {ContCode-ATM} consistently outperforms all other techniques on both databases.

The results of \emph{Exp.2} to \emph{Exp.5} are summarized in Table~\ref{tab:res_Exp} for the PolyU and CASIA databases. The {ContCode-ATM} consistently outperforms the three methods in all experiments and on both datasets.
Moreover, CompCode is consistently the second best performer except for very low FAR values in \emph{Exp.4} and \emph{Exp.5} on the PolyU database (see Fig.~\ref{fig:roc_Exp4} and Fig.~\ref{fig:roc_Exp5}). It is interesting to note that the OrdCode performs better than the DoGCode on the both databases.

We also compare performance of the Contour Code with that of the other methods reported in the literature. The results can be directly compared because all methods have reported EERs for the ``all versus all'' scenario. Table~\ref{tab:res_PolyU_MS_all} compares the EERs of various methods on the PolyU database. The proposed {ContCode-STM} achieves an EER of $0.0061\%$ whereas in the case of {ContCode-ATM} the EER reduces to $0.0030\%$ on the PolyU database. The proposed {ContCode-ATM} thus achieves a $75\%$ reduction in EER compared to nearest competitor, the CompCode. The error rates are extremely low (0.06 in 1000 chance of error) and indicate the viability of using multispectral palmprints in a high security system.

\begin{table}[t]
\caption{Comparative performance on the PolyU database.}
\footnotesize
\label{tab:res_PolyU_MS_all}
\centering
\begin{tabular}{|l|c|c|} \hline
\multirow{2}{*}{Method}                                     &  \multicolumn{2}{c|}{EER (\%)}\\ \cline{2-3}
                                                            & \emph{No Blur} &   \emph{With Blur}   \\ \hline
*Palmprint and Palmvein~\cite{zhang2011online}              &   0.0158  &   -       \\
*CompCode-wavelet fusion~\cite{han2008multispectral}        &   0.0696  &   -       \\
*CompCode-feature-level fusion~\cite{luo2012multispectral}  &   0.0151  &   -       \\
*CompCode-score-level fusion~\cite{zhang2010online}         &   0.0121  &   -       \\
OrdCode~\cite{sun2005ordinal}                               &   0.0248  &   0.0212  \\
DoGCode~\cite{wu2006palmprint}                              &   0.0303  &   0.0455  \\
{ContCode-STM}                                              &   0.0182  &   0.0061  \\
{ContCode-ATM}                                              &   0.0182  &   \textbf{0.0030} \\ \hline
\end{tabular}
\\
{\scriptsize *Results taken from published papers using the ``all versus all'' protocol.}
\end{table}
Table~\ref{tab:res_CASIA_MS_All} compares the EERs of various methods reported on the CASIA database. Both variants of the Contour Code outperform other methods and achieve an EER reduction of 60\% compared to the nearest competitor, CompCode. The EER of Contour Code on the CASIA database is higher relative to the PolyU database because the former was acquired using a non-contact sensor. Interestingly, the performance of DoGCode and OrdCode deteriorates with blurring. While in the case of CompCode and ContCode, the performance improves with the introduction of blur. However, the improvement in the proposed ContCode is much larger. Another important observation is that the ATM mode of matching always performs better than the STM mode and is thus the preferable choice for multispectral palmprint matching.

\begin{table}[t]
\caption{Comparative performance on the CASIA database.}
\footnotesize
\label{tab:res_CASIA_MS_All}
\begin{center}
\begin{tabular}{|l|D{.}{.}{1.4}|D{.}{.}{1.4}|} \hline
\multirow{2}{*}{Method}                     &  \multicolumn{2}{c|}{EER (\%)}\\ \cline{2-3}
                                            &   \multicolumn{1}{c|}{\emph{No Blur}}      &   \multicolumn{1}{c|}{\emph{With Blur}}   \\ \hline
*Wavelet fusion with ACO~\cite{kisku2010multispectral}  &   3.125               &   \multicolumn{1}{c|}{-}          \\
*Curvelet fusion with OLOF~\cite{hao2008multispectral}  &   0.50$\dag$          &   \multicolumn{1}{c|}{-}          \\
OrdCode~\cite{sun2005ordinal}                           &   0.5667              &   0.8667                          \\
DoGCode~\cite{wu2006palmprint}                          &   1.9667              &   2.8873                          \\
CompCode~\cite{kong2004competitive}                     &   0.8667              &   0.4993                          \\
{ContCode-STM}                                          &   0.6279              &   0.2705                          \\
{ContCode-ATM}                                          &   0.4333              &\multicolumn{1}{c|}{\textbf{0.2000}}\\ \hline
\end{tabular}
\end{center}
\scriptsize{*Results taken from published papers.\\
$^{\dag}$This result was reported on a database of 330 hands whereas only a subset comprising 200 hands has been made public}
\end{table}

\subsection{Identification Experiments}
\label{sec:cmc}
We perform identification experiments using 5-fold cross validation and report Cumulative Match Characteristics (CMC) curves and rank-1 identification rates. In each fold, we randomly select one multispectral palmprint image per subject to form the gallery and treat all the remaining images as probes. So, identification is based on a single multispectral image in the gallery for any probe subject. The identification rates are then averaged over the five folds. This protocol is followed for both databases.

\begin{figure}[b]
\centering
\subfloat{\label{fig:cmc_PolyU_MS}\includegraphics[trim = 0pt 0pt 10pt 0pt, clip, width=0.5\linewidth]{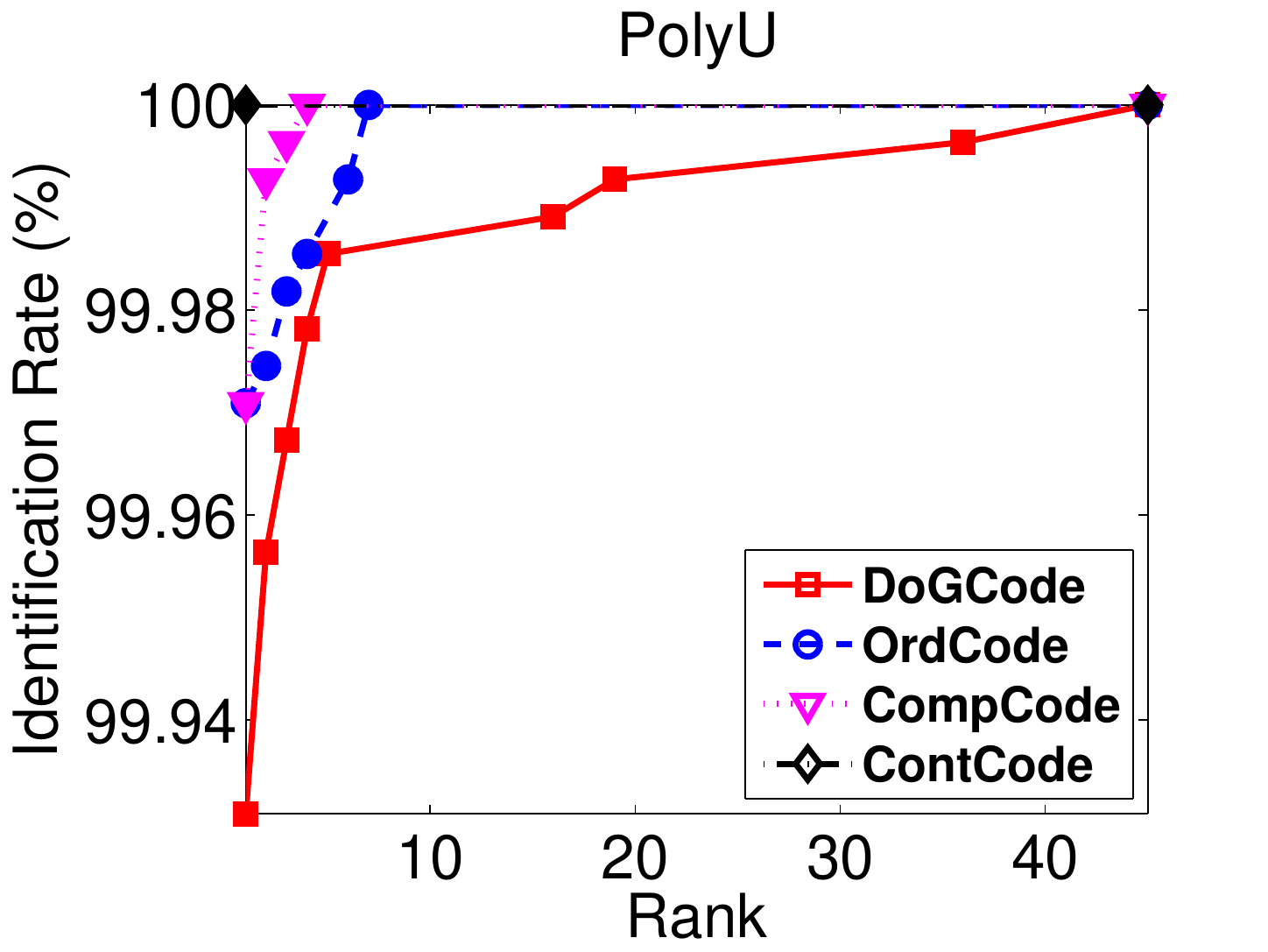}}
\subfloat{\label{fig:cmc_CASIA_MS}\includegraphics[trim = 0pt 0pt 10pt 0pt, clip, width=0.5\linewidth]{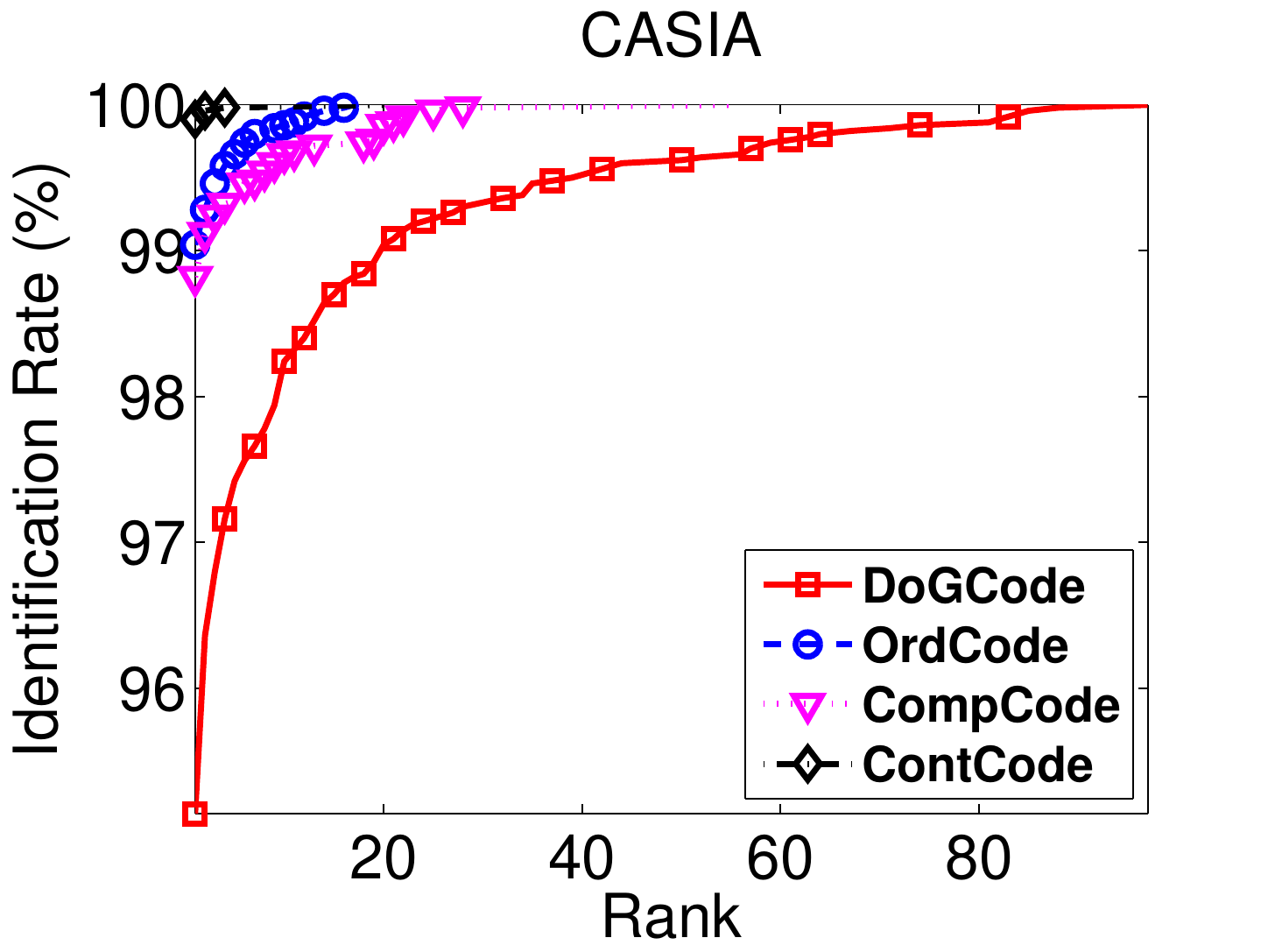}}
\caption{CMC curves for the identification experiment. Note that {ContCode-ATM} has an average rank-1 recognition rate of 100\% on the PolyU database.}
\label{fig:cmc}
\end{figure}

The CMC curves, for comparison with three other techniques, on both databases are given in Fig.~\ref{fig:cmc} and the identification results are summarized in Table~\ref{tab:res_ident}. The {ContCode-ATM} achieved an average of $99.88\%$ identification rate on the CASIA database and $100\%$ identification rate on the PolyU database. The proposed {ContCode-ATM} clearly demonstrates better identification performance in comparison to state-of-the-art techniques.

\begin{table}[h]
\caption{Comparison of rank-1 recognition rates and standard deviations on the CASIA and PolyU databases. Recognition rates are averaged over 5-folds.}
\footnotesize
\label{tab:res_ident}
\centering
\begin{tabular}{|l|c|c|} \hline
Method                              &  CASIA (\%)       &   PolyU (\%)  \\ \hline
OrdCode~\cite{sun2005ordinal}       &   99.02$\pm$0.11  &   99.93$\pm$0.05  \\
DoGCode~\cite{wu2006palmprint}      &   95.08$\pm$0.75  &   99.97$\pm$0.04  \\
CompCode~\cite{kong2004competitive} &   99.52$\pm$0.11  &   99.97$\pm$0.03  \\
{ContCode-ATM}                      &   99.88$\pm$0.08  &   100.0$\pm$0     \\ \hline
\end{tabular}
\end{table}

\subsection{Efficiency}

The computational complexity of matching is critical in practical identification scenarios because databases can be quite large. The binary Contour Code matching has been designed to carry out operations that are of low computational complexity. It comprises an indexing part whose complexity is independent of the database size. It depends on the Contour Code size and is, therefore, fixed for a given size. The indexing operation results in a relatively sparse binary matrix whose column wise summation can be efficiently performed. Summing a column of this matrix calculates the match score with an embedded score level fusion of the multispectral bands of an individual (in STM mode). A MATLAB implementation on a $2.67$ GHz machine with 8 GB RAM can perform over $70,000$ matches per second per band (using a single CPU core). The Contour Code extraction takes only $43$ms per band. In terms of memory requirement, the Contour Code takes only 676 bytes per palm per band.

\section{Conclusion}
\label{sec:conc}

We presented Contour Code, a novel multidirectional representation and binary hash table encoding for robust and efficient multispectral palmprint recognition. An automatic technique was designed for the extraction of a region of interest from palm images acquired with noncontact sensors. Unlike existing methods, we report quantitative results of ROI extraction by comparing the automatically extracted ROIs with manually extracted ground truth. The Contour Code exhibits robust multispectral feature capturing capability and consistently outperformed existing state-of-the-art techniques in various experimental setups using two standard databases i.e.~PolyU and CASIA. Binary encoding of the Contour Code in a hash table facilitates simultaneous matching to the database and score level fusion of the multispectral bands in a single step (in STM mode). Unlike existing techniques, score normalization is not required before fusion. The Contour Code is a generic orientation code for line-like features and can be extended to other biometric traits including fingerprints and finger-knuckle prints.


%

%
%

\ifCLASSOPTIONcompsoc
  \section*{Acknowledgments}
\else
  \section*{Acknowledgment}
\fi
\small{
This research was supported by ARC Grant DP0881813 and DP110102399. Authors acknowledge the Polytechnic University of Hong Kong for providing the PolyU-MS-Palmprint database and the Chinese Academy of Sciences' Institute of Automation for providing the CASIA-MS-PalmprintV1 database.}

\ifCLASSOPTIONcaptionsoff
  \newpage
\fi



\bibliographystyle{IEEEtran}
\IEEEtriggeratref{33}
\bibliography{IEEEabrv,ContCode_arXiv}
\end{document}